\documentclass{article}

\PassOptionsToPackage{numbers}{natbib}

 \usepackage[final,main]{neurips_2026}

\usepackage[utf8]{inputenc}
\usepackage[T1]{fontenc}
\usepackage{url}
\usepackage{booktabs}
\usepackage{enumitem}
\usepackage{amsfonts}
\usepackage{amstext}
\usepackage{amsmath}
\usepackage{amsthm}
\usepackage{mathtools}
\usepackage{nicefrac}
\usepackage{microtype}
\usepackage[table]{xcolor}
\usepackage{multirow}
\usepackage{algpseudocode}
\usepackage{wrapfig} 
\usepackage{marvosym}
\usepackage{longtable}
\usepackage{array}
\usepackage{threeparttable}
\usepackage[ruled]{algorithm}
\algtext*{EndFor}
\algtext*{EndIf}
\algtext*{EndWhile}
\usepackage{hyperref}
\usepackage{cleveref}
\usepackage{extdash}

\makeatletter
\newcommand*{\inlineequation}[2][]{%
  \begingroup
    \refstepcounter{equation}%
    \ifx\\#1\\%
    \else
      \label{#1}%
    \fi
    \relpenalty=10000 %
    \binoppenalty=10000 %
    \ensuremath{%
      #2%
    }%
    ~\@eqnnum
  \endgroup
}
\makeatother

\title{Data-Driven Soft Labeling Scales DNA Read Classification to Whole-Body Cell-Type Deconvolution}

\author{%
  Dmytro Rizdvanetskyi \quad Nathan Roos \quad Pavlo Lutsik  \\
  Department of Oncology\\
  KU Leuven\\
  Leuven, Belgium\\
  \texttt{\{dmytro.rizdvanetskyi, nathanericjeanbaptiste.roos, pavlo.lutsik\}@kuleuven.be}
}

\begin{document}

\maketitle

\begin{abstract}
  Cell-type deconvolution, the task of estimating the proportions of constituent cell types in a heterogeneous biological sample, is a core problem in computational biology.
  Methods that rely on epigenetic marks such as DNA methylation typically operate on aggregated methylation estimates, discarding the pattern-level information carried by individual DNA reads.
  Existing read-level approaches that exploit this information are scarce, and all remain restricted to few-class settings; scaling them further is an open problem because, at scale, non-discriminative reads dominate and hard labels conflict with the many-to-many mapping between methylation patterns and cell types, preventing classifier convergence.
  To overcome this, we propose data-driven soft labels that estimate the conditional cell-type distribution for each read, and integrate this scheme into \textit{Syto}, a new modular framework for read-level classification-based deconvolution.
  On a whole-body atlas of 39 human cell types, \textit{Syto} reduces MSE by 3.7$\times$ over the best examined baseline, with gains transferring to an out-of-distribution dataset spanning 16 tissues.
  \textit{Syto} lays the foundation for modeling increasingly large cell-type panels, with improved applications in biology and healthcare.
  The proposed soft-labeling scheme is further translatable to any setting with a many-to-many signal-to-label mapping.
\end{abstract}

\section{Introduction}

\begin{wrapfigure}{r}{0.41\textwidth}
  \vspace{-1.0em}
  \centering
  \includegraphics[width=0.4\textwidth]{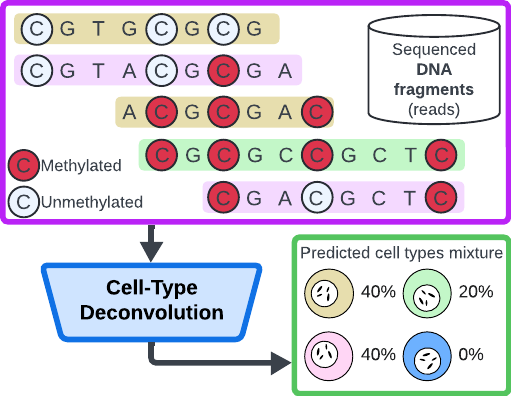}
  \caption{Principal scheme of the methylome-based cell-type deconvolution.}
  \label{fig:deconv-intro}
  \vspace{-0.5em}
\end{wrapfigure}

Epigenetic marks, such as histone modifications, chromatin accessibility and
DNA methylation, collectively define cell identity and can be used to
distinguish between different normal and malignant
cells~\cite{ehrlich1982amount,gama19835}. Among these, DNA methylation stands
out as the most stable and cost-effective to profile at scale. Advances in
methylome profiling have allowed researchers to exploit this biological signal
for crucial clinical tasks: tumor classification, tumor purity estimation, and
cell-type deconvolution. Traditionally, these tasks are treated separately.
Tumor classification scores a sample against predefined reference tumor cohorts
to confirm a diagnosis and guide treatment~\cite{capper2018dna,
  vermeulen2023ultra, yuan2025crossnn}. In contrast, tumor purity and fraction
estimation quantify the proportion of cancer cells in a tissue
sample~\cite{haider2020systematic, hanahan2022hallmarks} or circulating tumor
DNA (ctDNA) in the bloodstream. While these estimates are vital for early
diagnostics and monitoring disease progression~\cite{wang2023hot}, they often
struggle to capture complex cellular heterogeneity, such as the specific makeup
of the immune microenvironment~\cite{haider2020systematic}. Cell-type
deconvolution addresses this gap by determining the precise proportions of all
constituent cell types in an inhomogeneous sample. Because deconvolution
fundamentally captures both the target tumor proportion and its specific
cellular profile, the narrower problems of tumor classification and purity
estimation can be advantageously reduced to it. Thus, accurate deconvolution
covers a much broader range of clinical use cases than methods designed solely
for classification or purity estimation.

Existing methylome-based deconvolution methods typically operate on aggregated
methylation measurements ($\beta$-values~\cite{houseman2012dna},
$\alpha$-values~\cite{li2023comprehensive}, absolute counts of methylated
CpGs~\cite{caggiano2021comprehensive}) or fragment-level
counts~\cite{keukeleire2023cell, loyfer2023dna, ferro2024computational,
  stackpole2022cost}, omitting the differences in methylation pattern at the
level of the individual DNA fragments (or \textit{reads}) that are known to be
important for tumorigenesis~\cite{ehrlich2009dna}. Conversely, incorporating
local \textit{omics} signatures (potentially containing more than methylation)
and the broader nucleotide context may offer a more accurate representation of
cell-of-origin and may improve detection capacity~\cite{baker2026genome}.
Despite this potential, only three end-to-end classification-based
deconvolution methods exist~\cite{li2018cancerdetector, li2021dismir,
  jeong2025methylbert} and only two of them preserve the within-read methylation
pattern\cite{li2021dismir, jeong2025methylbert}. These methods are either unfit
to treat multi-class problems~\cite{li2018cancerdetector, li2021dismir}, or
wrongly apply the Bayesian framework to uncalibrated outputs of neural
networks, leading to highly biased deconvolution
results~\cite{li2021dismir,jeong2025methylbert}. Furthermore, none of these
methods has been demonstrated to work across dozens of different cell
types~\cite{loyfer2023dna}.

To address those limitations, we introduce \emph{Syto}\footnote{\emph{Syto} is
  the romanization of the Ukrainian word for \emph{sieve}, which fits our
  framework as it \emph{sieves} the different cell types.}: a modular framework
for read-level pattern-sensitive classification-based cell-type deconvolution
that scales to dozens of cell types. Our main contributions are as follows:

\begin{itemize}[noitemsep, topsep=-0.8em, leftmargin=*]
  \item We propose several labeling schemes to enable multi-class classification of
        omics-enriched DNA reads with a large number of classes. Among these, we
        introduce a biologically inspired, data-driven soft-labeling scheme -- the
        maximum-likelihood estimator of the true conditional distribution. This
        approach has potential applications in scenarios where the mapping from signal
        to class labels is many-to-many rather than many-to-one.
  \item We propose a linear calibration method for regression models with probability
        simplex outputs: a simple, hyperparameter-free post hoc correction that
        outperforms existing calibration baselines in our setting.
  \item We propose a modular framework with a decoupled classifier, deconvolver, and
        calibrator, allowing independent improvement of each component. We extend two
        existing read-level pattern-sensitive classifiers and one based on
        $\alpha$-values to operate within it and provide five deconvolvers.
  \item We conduct large-scale experiments on a 39-cell-type reference
        atlas~\cite{loyfer2023dna} and on an out-of-distribution dataset of 16
        tissues~\cite{li2023comprehensive}. \textit{Syto} consistently improves
        deconvolution accuracy over existing baselines.
\end{itemize}

\section{Related Work}

\paragraph{Read-level classification-based deconvolution methods.}
\label{sect:existing_rlcbd_methods}

\textit{CancerDetector}~\cite{li2018cancerdetector} is a tumor detection and fraction estimation method that models the joint methylation states of CpG
\footnote{CpG sites are places in the genomic sequence where a Cytosine is followed by Guanine in the 5' to 3' direction.
    These are the most prevalent methylation targets in mammals~\cite{ehrlich1982amount}.}
sites in DNA reads using a Beta-Bernoulli distribution.
For every cell-type specific genomic region, the model learns two separate Beta distributions conditioned on $\alpha$-values for tumoral and normal classes; these priors are then used to estimate the tumor-derived cfDNA fraction via maximum likelihood.
\textit{Dismir}~\cite{li2021dismir} is another tumor fraction estimation method; it uses a CNN-LSTM architecture with both DNA sequence and methylation state as input to assign each read a \emph{d-score} representing the probability that it originates from a tumor. The sample-level tumor fraction is then obtained by maximizing the posterior probability over all d-scores.
\textit{MethylBERT}~\cite{jeong2025methylbert} is an adaptation of the BERT architecture~\cite{devlin2019bert} to read-level classification.
The model is first pre-trained on 3-mer tokenized reference genome sequences, then fine-tuned by adding methylation embeddings encoding the CpG methylation state at each position and by training a read classifier.
Tumor purity is also obtained via maximum likelihood estimation.
\textit{MethylBERT} is the only existing read-level pattern-sensitive classification-based model that has demonstrated deconvolution with up to five cell types, using focal loss~\cite{lin2017focal} to mitigate class imbalance.

\paragraph{Label distribution learning and soft labeling.} \citet{geng2016label} has introduced label distribution learning as a computational task of estimating the degree to which each label describes the instance.
\citet{xu2021label} coined the term \textit{label enhancement} (LE) for the task of constructing a label distribution (or \textit{soft labels}) from one-hot encoded labels (or \textit{hard labels}).
They proposed Graph Laplacian Label Enhancement (GLLE) to learn the label distribution from hard labels by propagating label information through a graph constructed from the feature space.
However, GLLE anchors the label distribution of each instance to its original hard label, which is not suitable for DNA methylation data (cf.~\Cref{sect:labeling}).
\citet{wang2019classification} adapted label distribution learning to classification by, \textit{inter alia}, upweighting the loss of instances whose label distributions have high entropy.
\citet{szegedy2016rethinking} introduced label smoothing of hard labels to improve the generalization of computer vision models.
In their definition, the original hard label is replaced by a soft label constructed as a mixture of the one-hot encoded label and the uniform distribution.

\paragraph{Calibration.}
\citet{muller2019does} highlighted that applying label smoothing results in implicit confidence calibration that can be equated to temperature scaling~\cite{guo2017calibration} under certain parameters.
\citet{kull2019beyond} proposed a class-wise Dirichlet calibration method for classifiers applied on log-probabilities.
The calibration transfer function is of the form:
\inlineequation[eq:multiclass_calibration_standard_form]{\mathbf{\tilde{p}} \coloneq \textrm{softmax} (W \log(\mathbf{\hat{p}}) + \mathbf{b})},
where $\coloneq$ denotes a definition, \(\mathbf{\tilde{p}}\) is the calibrated prediction vector, \(\mathbf{\hat{p}}\) is the uncalibrated prediction vector, \(\log\) is applied element-wise, and \(W\) and \(\mathbf{b}\) are the parameters of the calibration model.
Dirichlet calibration is a generalization of both temperature scaling (\(W = tI\) and \(\mathbf{b} = \mathbf{0}\))\cite{guo2017calibration} and vector scaling (diagonal \(W\))~\cite{guo2017calibration, kull2019beyond}.

\section{Methods}

\subsection{\texorpdfstring{\textit{Syto}}{Syto}: a modular framework}

Existing methods for read-level classification-based cell-type deconvolution
follow three steps: selection of informative genomic regions for training and
inference, classification of individual reads, and estimation of target
cell-type proportions. This common structure offers an opportunity for
harmonization. In this work, we propose \textit{Syto}, a modular framework for
read-level classification-based deconvolution. \textit{Syto} specifically
subsumes the classification and proportions estimation steps; it can be applied
out of the box, regardless of the region selection procedure.

\paragraph{Inference.}
\textit{Syto} operates on a sequencing library \(\mathcal{D} \coloneq \left\{ (r_i, g_i, s_i) \right\}_{1 \leq i \leq N}\) consisting of $N$ DNA sequences (or \textit{reads}) $r_i$ obtained from a biological bulk sample.
Each read $r_i$ belongs to a single group of genomic regions (or \textit{GR group}) $g_i \in \mathcal{G} \coloneq \{1, \ldots, G\}$.
Differentially Methylated Regions (DMRs) are a common choice for GR groups, and we use them in this work.
Each read $r_i$ is also assigned an \textit{omics signature} $s_i$.
In this work, we focus on methylation, and so the signature is the set of pairs (position, methylation state) for each CpG site in the read.
The methylation state is either 1 (methylated) or 0 (unmethylated).
The omics signatures can be extended to contain information from gene expression, histone marks, chromatin accessibility, etc.
The first step is to classify all reads, producing a vector of probability over the classes ${\mathbf{u_i} \coloneq {(u_{i,1}, \ldots, u_{i,C})}^T \in \Delta^{C-1}}$ for every read $r_i$,
where $C \in \mathbb{N}$ is the number of cell types to deconvolve, and $\Delta^{C-1}$ is the ${(C-1)}$\Hyphdash* dimensional probability simplex.
The predictions are then aggregated into a \textit{feature matrix} (or \textit{prediction matrix}) $P \in \mathbb{R}^{G\times C}$ (cf.~\Cref{supp:predictions_aggregation} for further details):

\begin{equation}
  P_{g,c} \coloneq \frac{1}{\sum_{i=1}^N \delta_{g, g_i} w_i  }\sum_{i=1}^N \delta_{g, g_i} w_i u_{i,c},\quad \forall (g,c) \in \mathcal{G}\times \mathcal{C},
\end{equation}

where $\mathcal{C}\coloneq \{1, \ldots, C\}$ is the set of cell type indexes,
$w_i$ is a weight equal to the number of CpG sites in $r_i$, and $\delta_{x,y}$
is the Kronecker delta equal to~1 if $x=y$ and 0 otherwise. The prediction
matrix is fed into a \textit{deconvolver} model, which outputs mixture
proportions $\mathbf{\hat{p}} \in \Delta^{C-1}$. Optionally, a post-hoc
calibrator can be applied to obtain calibrated mixture proportions
$\mathbf{\tilde{p}} \in \Delta^{C-1}$.

\begin{figure}[tb]
  \centering
  \includegraphics[width=1\textwidth]{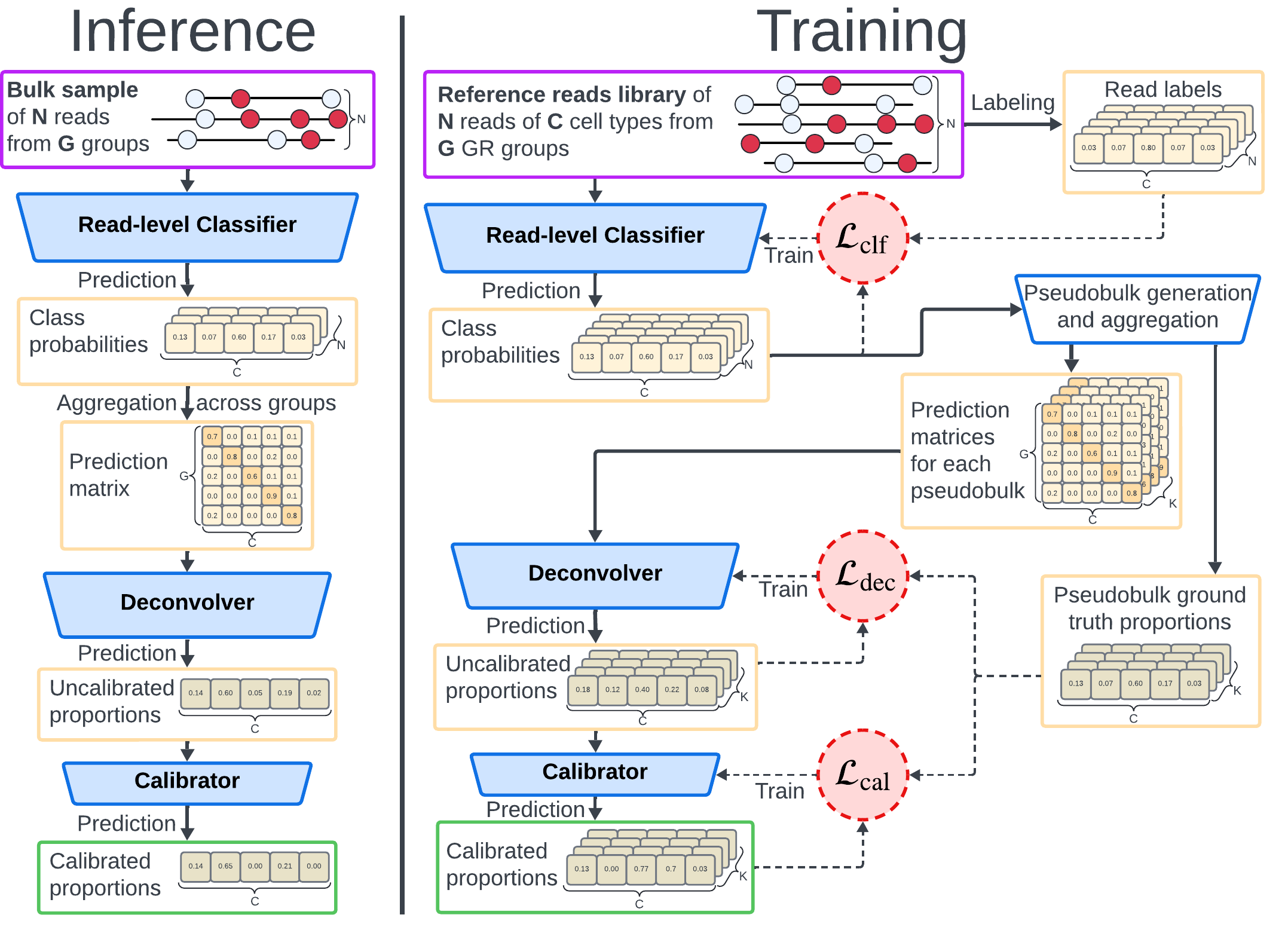}
  \caption{Proposed framework (\textit{Syto}) for read-level classification-based cell-type deconvolution.
  }
  \label{fig:framework}
\end{figure}

\paragraph{Training.}
\textit{Syto} has three trainable parts: the read-level classifier, the deconvolver, and the calibrator. Each of them can be independently improved.
The training dataset additionally contains, for each read $r_i$, the class label $c_i \in \mathcal{C}$, which is the cell type the read originates from.
To train the classifier, we assign a label vector to each read.
As we explain in \Cref{sect:labeling}, standard one-hot encoded labels (or \textit{hard labels}) are insufficient for model convergence.
We propose several labeling methods, namely hard labels with an additional background class, a smoothed version of these hard labels (cf.~\Cref{sect:hard_labels_with_background}), and a probabilistic labeling called \textit{data-driven soft-labeling} (cf.~\Cref{sect:soft_labeling}).
We extend and modify the three existing read-level classifiers (\textit{CancerDetector}, \textit{Dismir} and \textit{MethylBERT}) and provide custom losses that enable convergence across different label types.
Deconvolvers are trained on a dataset of \textit{pseudobulks}, \textit{i.e.} pairs of (ground truth pseudobulk proportions, prediction matrix) obtained by subsampling the original dataset with known proportions.
The calibrators are trained on the deconvolvers' outputs, using the pseudobulk ground-truth proportions as targets.
The inference and training of \textit{Syto} is summarized in \Cref{fig:framework}.

We deliberately omit the selection of informative regions to emphasize that
\textit{Syto} can be applied out of the box with different GR groups, as
illustrated in \Cref{fig:framework_specific}. In particular, \textit{Syto} can
be applied to long reads by treating the whole genome as a single GR group, as
in~\cite{baker2026genome}.

\subsection{Labeling}
\label{sect:labeling}

\begin{wrapfigure}{r}{0.41\textwidth}
  \vspace{-3.0em}
  \centering
  \includegraphics[width=0.4\textwidth]{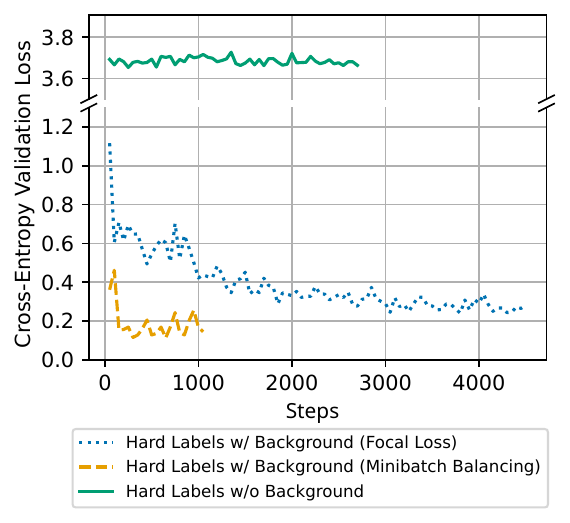}
  \caption{Validation cross-entropy loss for \textit{MethylBERT} using hard labeling with or without background class.}
  \label{fig:hard_labeling_loss_curves}
  \vspace{-1.6em}
\end{wrapfigure}

Using standard one-hot encoded labels (or \textit{hard labels}) as the
classification target when the number of cell types is large impedes model
convergence. Indeed, when deconvolving $C$ cell-types with an atlas where each
GR group $g$ is specific for a single cell type $c^{(g)}$, only about $1/C$
($\approx 3\%$ when $C=39$) of reads originate from the cell type for which
their GR group is specific (so-called \textit{on-target} reads) and thus carry
a distinctive omics signature. The remaining \textit{off-target} reads,
however, share similar omics signatures, despite being assigned to different
cell-type labels. This yields an impossible learning task and results in
degraded models that learn uniform class probabilities regardless of the
context. \Cref{fig:hard_labeling_loss_curves} shows clearly that
\textit{MethylBERT} does not converge when trained on standard hard labels. In
this section, we propose several labeling schemes to overcome this issue.

\subsubsection{Hard labels with background}
\label{sect:hard_labels_with_background}

\paragraph{Hard labels with background.}
To mitigate the detrimental effects of off-target reads, we can assign them to
an additional \textit{background} label $\bot \coloneq C+1$. Let $r_i$ be a DNA
read in GR group $g$, $c_i \in \mathcal{C}$ be its original cell-type label and
$c^{(g)}$ be the cell type for which GR group $g$ is specific. We define the
hard label with background $y_i$ as $c_i$ if $c_i = c^{(g)}$ (\textit{i.e.} if
$r_i$ is on-target) and $\bot$ otherwise. This can be seen as an explicit and
more flexible extension of the one-versus-all binary labeling scheme from
\cite{jeong2025methylbert}, in which on-target and off-target reads are
assigned labels $1$ and $0$, respectively.

\paragraph{Canonical label smoothing on hard Labels with background.}
We also apply the canonical label smoothing of \citet{szegedy2016rethinking} to
hard labels with background. Let $\epsilon \in [0,1]$ be the smoothing
parameter. The smoothed hard label of $r_i$ is the vector $\mathbf{\bar{y}_i}
  \in \Delta^{C}$ defined as:
\begin{equation}
  \bar{y}_{i, c} \coloneq (1-\epsilon)\delta_{c, y_i} + \frac{\epsilon}{C}, \quad \forall c\in \mathcal{C} \cup \{\bot\}.
\end{equation}

\subsubsection{Data-driven soft labeling}
\label{sect:soft_labeling}

\paragraph{Biological rationale.}
We argue that data-driven soft labels, constructed as probabilistic
representations that capture the conditional distribution over cell types for
each omics signature, are more appropriate than hard labels for read-level cell
type classification. The main reason is that the mapping between omics
signatures and cell types is not many-to-one, as in standard classification,
but rather many-to-many, because a single omics signature (and in particular a
single methylation pattern) in a given genomic region can be associated with
multiple cell types. This is due, among several reasons, to imperfect
propagation and maintenance of DNA methylation patterns across cell
divisions~\cite{ushijima2003fidelity}, as well as high methylome variation
across the cell cycle, with a sharp drop in average methylation during the S
phase~\cite{geisenberger2025single}. Both issues are further compounded by
technical noise: the purification of cell types prior to sequencing is
imperfect~\cite{loyfer2023dna}, and the techniques to infer the methylation
status of CpG sites are noisy~\cite{zhou2019systematic}. This ambiguity in the
mapping was previously resolved by selecting highly discriminative regions that
are hypomethylated in a specific cell type and methylated in all
others~\cite{zilbauer2013genome,stadler2011dna,loyfer2023dna,schubeler2015function}.
However, as we consider an increasing number of cell types, it becomes
necessary to resort to less specific regions, where methylation patterns can
only discriminate between non-singleton cell-type
groups~\cite{li2023comprehensive}. In this case, assigning a different hard
label to reads sharing the same methylation pattern is not biologically
meaningful and may impair the models' ability to learn the true conditional
distribution of cell types.

\paragraph{Data-driven soft labels as MLE of the true categorical distribution.}
Let $r_i$ be a read with omics signature $s_i$, its \emph{data-driven soft
  label} is defined as the empirical frequency of each class among the reads
sharing that signature. For all $c \in \mathcal{C}$, we denote by $n_{i,c}$ the
number of reads of class $c$ with signature $s_i$. The soft label $\mathbf{q_i}
  \coloneq (q_{i,1}, \ldots, q_{i,C})^T \in \Delta^{C-1}$ of $r_i$ is the vector
defined as:

\begin{equation}
  q_{i,c} \coloneq \frac{n_{i,c}}{\sum_{c' \in \mathcal{C}} n_{i,c'}}, \quad \forall c \in  \mathcal{C}.
\end{equation}

$\mathbf{q_i}$ is the unique maximum likelihood estimator of the true conditional categorical distribution $P(c \mid s_i)$.
This results from a direct application of the classical MLE for categorical distributions~\cite{murphy2012machine}.

\paragraph{Coverage normalization by pooling.} The construction of soft labels faces two challenges: class imbalance and low
per-signature read coverage. We handle imbalance by reweighting each class
count by the inverse of its global frequency in the training data. The second
issue is addressed by pooling class counts from the nearest
signatures\footnote{In this work, we use the Jaccard distance over the
  signatures (cf.~\Cref{sect:supp_jaccard_distance}). The extension of omics
  signatures to other signals such as chromatin accessibility, histone marks or
  gene expression will require adapting the distance.} within each genomic
region, until a minimum read count $\tau$ is reached or a maximum distance
threshold $\delta_{\max}$ is exceeded. Importantly, this count transfer between
signatures is \textit{not} symmetric. This procedure yields \emph{data-driven
  soft labels with pooling}. We summarize the soft labeling procedure with class
reweighting and pooling in \Cref{alg:soft_labeling}. Our adaptations
differentiate it from the more general \textit{KNN soft labels} by
\citet{el2006study}.

\paragraph{Confidence-weighted Cross Entropy (CE) loss.}
When $C$ is large, the majority of reads are off-target and thus have
high-entropy soft labels. In this case, the standard CE loss leads the model to
predict uniform distributions, which is not desirable. We use a
confidence-weighted CE loss. For a read $r_i$ with soft label $\mathbf{t}_i$ of
length $C$ and predicted class probabilities $\mathbf{u}_i$, the weight is
$\big(\max(\mathbf{t}_i) - C^{-1}\big) / \big(1 - C^{-1}\big)$ and the loss is:
\begin{equation}
  \mathcal{L}^{\mathrm{CW}}(\mathbf{u}_i, \mathbf{t}_i) := \left(\frac{\max(\mathbf{t}_i) - C^{-1}}{1 - C^{-1}}\right)\mathrm{CE}(\mathbf{u}_i, \mathbf{t}_i),
\end{equation}
where $\mathrm{CE}(\mathbf{u_i}, \mathbf{t_i}):= -\sum_{c} t_{i,c} \log u_{i,c}$ is the standard cross-entropy.

Our data-driven soft labels fit in the general field of Label
Enhancement~\cite{xu2021label}. However, to our knowledge, we are the first to
propose such a construction of soft labels. Moreover, in this setting, the most
informative reads are those with low entropy, which is the opposite situation
of the one in the existing literature~\cite{wang2019classification}.

\begin{algorithm}[t]
  \caption{Data-driven soft labeling with pooling}
  \label{alg:soft_labeling}
  \begin{algorithmic}[1]

    \Statex \textbf{Input:} A set of signatures $\{{s_i\}}_{1 \leq i \leq S }$; for each signature $s_i$, the class counts $\mathbf{n_i} \in \mathbb{N}^C$ of reads having signature $s_i$; minimal count threshold $\tau \in \mathbb{N}$; maximal distance $\delta_{\max} \in \mathbb{R}^+$

    \Statex \textbf{Output:} Soft label $\mathbf{q_i}^{\text{pooled}} \in \Delta^{C-1}$ for each signature $s_i$

    \For{each class $c = 1, \ldots, C$}
    \State $w_c \leftarrow \sum_{i=1}^S n_{i,c}$ \hfill  {\small $\triangleright$ compute class weights as global counts}
    \EndFor

    \For{each GR group $g$}            \hfill {\small $\triangleright$ aggregation is performed only within GR groups}
    \State Compute $J[i,j] \leftarrow \textsc{JaccardDist}(s_i, s_j)$ for all pairs of signatures $(s_i, s_j)$ in $g$

    \For{each signature $s_i$ in region $g$}
    \State $\mathbf{\tilde{n}_i} \leftarrow (0, \ldots, 0)^T \in \mathbb{N}^C$ \hfill {\small $\triangleright$ initialize pooled class counts for $s_i$}

    \For{$j$ in $\mathrm{argsort}(J[i,:])$} \hfill {\small $\triangleright$ iterate over signatures ($s_i$ included) from nearest to farthest}

    \If{$J[i,j] > \delta_{\max}$ or $\mathrm{sum}(\mathbf{\tilde{n}_i}) \geq \tau$} \textbf{break} \EndIf
    \State $\mathbf{\tilde{n}_i} \leftarrow \mathbf{\tilde{n}_i} + \mathbf{n}_j$            \hfill {\small $\triangleright$ add class counts of $s_j$ to the pooled counts of $s_i$}
    \EndFor
    \State $\hat{n}_{i,c} \leftarrow \tilde{n}_{i,c} \cdot w_c$ for each class $c$          \hfill {\small $\triangleright$ weight the pooled counts}
    \State $q_{i,c}^{\text{pooled}} \leftarrow \hat{n}_{i,c} \, / \, \sum_{c'=1}^C \hat{n}_{i,c'}$ for each class $c$ \hfill {\small $\triangleright$ normalize}
    \EndFor
    \EndFor
  \end{algorithmic}
\end{algorithm}

\subsection{Read-level classifiers}
\label{sect:read_level_classifiers}

{\setlength{\parskip}{0pt}
    \paragraph{Extension of existing classifiers.}
    We adapted \textit{CancerDetector}~\cite{li2018cancerdetector} to the
    multiclass setting by extending the Beta-Bernoulli read likelihood model to all
    $C$ cell types, with per-marker, per-class parameter estimation by MLE
    (cf.~\Cref{sect:supp_mat_cancer_detector}). We did a major refactoring of
    \textit{MethylBERT}~\cite{jeong2025methylbert} to harmonize it with the Hugging
    Face Transformers library~\cite{wolf2020transformers} and enable faster
    fine-tuning. We also substituted the original classification head with one with
    GR group attention (cf.~\Cref{supp:sect_dmr_attention_head}). The original
    \textit{Dismir}~\cite{li2021dismir} implementation does not integrate GR group
    context, making it untrainable on short reads with many classes, as the context
    cannot be inferred from the reads themselves. Therefore, we also use the
    classification head with GR group attention for \textit{Dismir}.

    \paragraph{Lookup classifier as baseline.}
    To serve as a baseline, we introduced \textit{Lookup Classifier}, a
    modification of 1-NN. For an input read $r$ with signature $s$, the predicted
    probability vector under soft labeling corresponds to the soft label computed
    on the training data. Under hard labeling with background, the prediction is
    instead the argmax over all reads sharing signature $s$, with ties resolved by
    averaging the one-hot encoded labels. If $s$ was never seen in the training
    data, for soft labels, the training signatures at minimal Jaccard distance to
    $s$ are aggregated, their normalized counts are pooled, and the resulting soft
    label is returned. For hard labels, we return the average of the one-hot
    encoded labels of the training signatures at minimal distance to $s$. If $r$
    belongs to no GR group in the training data, a uniform distribution is returned
    for soft labels, and the background class is returned for hard labels. }

\subsection{Deconvolution models}
\label{sect:methods_deconvolvers}

All previous works relied in some way or another on maximum likelihood
estimation (MLE) in order to infer the mixture proportions from the
classifiers' outputs. However, preliminary experiments showed that MLE methods
applied on top of the classifier outputs fail in our large-scale setting
(cf.~\Cref{sect:supp_mle_deconvolvers}). We therefore propose five distinct
deconvolution models (or \textit{deconvolvers}) to learn the transfer function
from the classifier output to the mixture proportions, which can all be used in
combination with every classifier. Three of those models are supervised machine
learning (ML) models, namely \emph{XGBoost}~\cite{chen2016xgboost} (referred to
as \emph{XGB}), and two neural networks: a multi-layer perceptron with three
hidden layers (referred to as \emph{MLP}) and a multi-layer perceptron with one
wider hidden layer (referred to as \text{shallow wide network} or \emph{SWN}),
both implemented with PyTorch~\cite{paszke2019pytorch}. \emph{XGB} is trained
with RMSE loss, while the neural networks are trained with $loss = MSE +
    0.5KL$. The other two methods are the widely used \emph{Non-negative Least
    Squares} or \emph{NNLS}, and a variant with sum-to-one constraint:
\emph{Probability Simplex Least Squares} or \emph{PSLS}; both implemented with
cvxpy~\cite{diamond2016cvxpy}. \emph{NNLS} and \emph{PSLS} are initialized on
the $C$ prediction matrices resulting from mixtures with a single cell type. ML
deconvolvers are fitted on the $K$ prediction matrices obtained from the
aggregation of the classifier predictions. For the pseudobulk experiment, we
have simulated $K=100{,}000$ pseudobulk mixtures for the training and
validation split, and $50{,}000$ for the test split, via the sampling procedure
described in \Cref{supp:pseudobulk_sampling}. We describe the models'
architectures and fitting parameters in detail in
\Cref{sec:supp_deconv_params}.

\subsection{Calibration of predicted proportions}
\label{sect:methods_calibration}

\vspace{-2mm}

The output of deconvolvers is not necessarily well calibrated. Noting that the
mapping between predicted and true proportions appears well approximated by a
linear function (cf.~\Cref{fig:supp_linear_cal_illustration}), we propose a
simple, hyperparameter-free method we call \textit{linear calibration}. For all
cell-type \(c \in \mathcal{C}\), we fit $a_c$ and $b_c$ the parameters of the
best fitting linear model mapping the validation set predicted proportions of
$c$ \((\hat{p}_{k,c})_{k=1}^K\) to the true proportions \((p_{k,c})_{k=1}^K\):
$p_{k,c} = a_c \hat{p}_{k,c} + b_c + \epsilon$, where \(\epsilon\) is the error
term. The fitted parameters are arranged in two vectors \(\mathbf{a} \coloneq
(a_1, \ldots, a_C)^T \in \mathbb{R}^C\) and \(\mathbf{b} \coloneq (b_1, \ldots,
b_C) \in \mathbb{R}^C\) and inference is performed as follows:

\begin{equation}
  \mathbf{\tilde{p}} \coloneq f(\mathbf{a} \odot \mathbf{\hat{p}} + \mathbf{b}),
\end{equation}

where $\odot$ is the Hadamard product, \(\mathbf{\hat{p}}\) is the uncalibrated
prediction vector, \(\mathbf{\tilde{p}}\) is the calibrated prediction vector
and \(f\) is a function that corrects the small deviations of $\mathbf{a} \odot
  \mathbf{\hat{p}} + \mathbf{b}$ from the probability simplex. We present two
variants of \(f\): \emph{clip-normalize}, where \(f : \mathbf{x} \in
\mathbb{R}^C \mapsto ((x_c)_+ / \sum_{c'\in \mathcal{C}} (x_{c'})_+)_{c \in
    \mathcal{C}}\) with \((x)_+ \coloneq \max(0, x)\), and
\emph{simplex-projection}, which projects the uncalibrated vector onto the
probability simplex using the threshold algorithm by
\citet{duchi2008efficient}. We performed preliminary experiments comparing the
performance of our method with established calibration methods from the
literature, namely Dirichlet calibration, vector scaling and temperature
scaling~\cite{guo2017calibration,kull2019beyond}. Results in
\Cref{sect:supp_mat_calibration} show that our method and vector scaling yield
better MSE\@. We proceed with these in the main experiments.

\vspace{-3mm}

\section{Experiments}

 {\setlength{\parskip}{0pt}
  \subsection{Experimental setups}
  \label{sect:experiments setup}

  \paragraph{External baselines.} We compared \textit{Syto} to \textit{UXM}~\cite{loyfer2023dna},
  \textit{CelFiE}~\cite{caggiano2021comprehensive}, \textit{Houseman
    CP}~\cite{houseman2012dna} and \textit{EpiDISH}~\cite{teschendorff2017epidish}.
  The motivation and implementation details are described in
  \Cref{sect:supp_baselines_description}.

  \paragraph{Pseudobulk experiment.}\label{sect:pbm_deconvolution}
  We chose the largest available (by number of cell-types) DNA methylation atlas
  of healthy adult human cell types for
  modeling~\cite{loyfer2023dna,moss2023megakaryocyte}. The dataset comprises 205
  WGBS (Whole Genome Bisulfite Sequencing) libraries from 39 major somatic cell
  types, originating from 137 donors, and is publicly available at GEO under
  accession number GSE186458. We preprocessed the data as described in
  \Cref{sect:supp_data_preprocessing}. We grouped the DMRs of
  \citet{loyfer2023dna} into 39 GR groups, each containing up to 25 DMRs specific
  to a single cell type. To fit the read-level classifiers, we first filtered the
  dataset to keep only reads that overlapped with at least 4 CpGs in the
  reference atlas. We split the data into training, validation and test sets
  following the procedure in \Cref{sect:supp_split_construction}, ensuring that
  every GR group has DNA reads in all splits while isolating biological sources
  whenever possible. Resulting split ratios were 62:19:19. Of the 205 biological
  samples, 4 were shared across all splits and 2 were shared between the
  validation and test splits. To mitigate the risk of overestimating the
  generalization capabilities of the methods, we also evaluated them on an OOD
  (Out-of-Distribution) RRBS (Reduced Representation Bisulfite Sequencing)
  dataset (see below). We trained and evaluated 360 possible permutations of
  methods of \textit{Syto} (cf.~\Cref{fig:recap_experimental_setups}) and
  measured the MSE, MAE, KL divergence and Limits of Agreement on the predicted
  proportions (cf.~metrics definitions in \Cref{sect:supp_deconv_metrics}). The
  fitting parameters for the classifiers, deconvolvers and calibrators are
  described in sections \Cref{sec:supp_classifier_training},
  \Cref{sec:supp_deconv_params} and \Cref{sect:supp_par_calibrators_fitting},
  respectively.

  \paragraph{Deconvolution of OOD RRBS dataset.}\label{sect:par_ood_rrbs_experiment}
  In the absence of a publicly available alternative dataset, we used the RRBS
  methylation atlas of 29 human tissues, composed of 521 non-cancer samples, by
  \citet{li2023comprehensive} (GEO accession number GSE233417). For each tissue,
  we constructed the expected cell type proportions using the \emph{Tabula
    Sapiens} dataset~\cite{quake2025tabula, the2022tabula} as a joining key
  (cf.~\Cref{sect:supp_mapping_cfsort_tissues_to_tabula_sapiens}) The resulting
  16 tissues had a matching rate in the range $88.7\% - 100\%$ and 260 biological
  samples in total (see tissue composition in \Cref{fig:cfsort-tabula-map}).
  \emph{Tabula Sapiens} cell type proportions cannot be treated as ground truth
  due to technical biases in data collection~\cite{the2022tabula}. Thus, instead
  of the usual deconvolution metrics, we use a proximal measure based on the
  insight that cell types not expected in the target tissue should not appear in
  the deconvolution results, or equivalently, that the predicted proportions of
  expected cell types should sum to a high value, ideally one. We call ``Tissue
  Concordance Score'' (or \textit{TCS}) the sum of predicted proportions of cell
  types expected in the tissue (cf.~\Cref{sect:supp_par_TCS}). To determine the
  set of expected cell types, we use matched cell types with individual
  contributions of at least 1\% per tissue in \emph{Tabula Sapiens}. For this
  experiment the train split is the fusion of the training and validation splits
  of the pseudobulk experiment, the validation split is the test split of the
  pseudobulk experiment, and the test split is the OOD RRBS dataset. Soft labels
  are computed using all splits of the pseudobulk experiment. We evaluated the
  same 360 permutations of \textit{Syto} featured in the pseudobulk experiment.
  Additionally, we calculated \textit{TCS} for \textit{UXM} operating with a
  $10\times$ more DMRs, denoted \textit{UXM U250}.}

\vspace{-3mm}

\subsection{Results}

\vspace{-2mm}

\paragraph{Pseudobulk experiment.}
The main results are summarized in \Cref{tab:main_results} and the integrality
of results is in \Cref{sect:supp_pseudobulk_exp_detailed_results}.
\textit{Syto}, equipped with our extended \textit{CancerDetector, MethylBERT}
or \textit{Dismir}, consistently outperforms the best baseline,
\textit{CelFiE}, achieving a $3.7\times$ reduction of the MSE (from $3.28
  \mathrm{e}{-4}$ to $0.88 \mathrm{e}{-4}$) with the best setting
(\textit{MethylBERT} + data-driven soft labels w/ pooling + \textit{NNLS} +
Linear calibrator with simplex projection). Our linear calibrator with simplex
projection appears as the best calibrator, uniformly improving the MSE,
including on baselines. Similarly, our soft labeling with pooling uniformly
outperforms alternative labeling schemes in terms of MSE\@.

{\setlength{\textfloatsep}{0pt}
\begin{table}[t]
  \centering
  \caption{Pseudobulk experiment test split deconvolution metrics.
    Best result in each section is in \textbf{bold}, second best is \underline{underlined}, best across all sections is {\color{teal}highlighted}.
    Best result per classifier across labeling schemes is \textit{italicized}.
    For all metrics, we indicate the 95\% confidence interval computed with BCa bootstrap~\cite{efron1987bca} with 10,000 resamples.
    Results are reported for deconvolvers fitted on the top 156 features
    unless \textsuperscript{\textdagger} is added to the deconvolver name,
    in which case they are fitted on the combination of diagonal + background column
    of the feature matrices (cf.~\Cref{supp:dec_features_selection}).}
  \label{tab:main_results}
  {\footnotesize
    \resizebox{\textwidth}{!}{%
      \begin{tabular}{l l l l l l l l}
        \toprule
        \multirow{2}{*}{\textbf{Classifier}}                                                                     & \textbf{Labeling Scheme} & \multirow{2}{*}{\textbf{Deconvolver}} & \multirow{2}{*}{\textbf{Calibrator}}
                                                                                                                 & $\mathbf{R}^\mathbf{2}$  & \textbf{MAE}                          & \textbf{MSE}                                                                                                                                                                                                                                              \\
                                                                                                                 & \textbf{(or prior type)} &                                       &                                      & ($\times 10^{-2}$)                                                     & ($\times 10^{-3}$)                                                  & ($\times 10^{-4}$)                                                  \\
        \midrule
        \multicolumn{7}{l}{\textit{(a) Syto variants with ablation of calibration — for each classifier and labeling/prior, the best deconvolver by test MSE is reported}}                                                                                                                                                                                                                                                                      \\
        \midrule
        \multirow{4}{*}{{\shortstack[l]{Dismir~\cite{li2021dismir}                                                                                                                                                                                                                                                                                                                                                                              \\(CNN+LSTM)}}}
                                                                                                                 & Hard labels w/ bckg.     & SWN                                   & None                                 & 97.98 {\scriptsize[97.94, 98.01]}                                      & 3.44 {\scriptsize[3.42, 3.45]}                                      & 1.68 {\scriptsize[1.65, 1.71]}                                      \\
                                                                                                                 & Canonical Soft Labels    & MLP                                   & None                                 & 97.49 {\scriptsize[97.45, 97.54]}                                      & 3.65 {\scriptsize[3.63, 3.67]}                                      & 2.08 {\scriptsize[2.05, 2.12]}                                      \\
                                                                                                                 & DD Soft Labels w/o pool. & SWN                                   & None                                 & 98.24 {\scriptsize[98.20, 98.27]}                                      & 3.19 {\scriptsize[3.18, 3.20]}                                      & 1.46 {\scriptsize[1.43, 1.49]}                                      \\
                                                                                                                 & DD Soft Labels w/ pool.  & SWN                                   & None                                 & \textit{98.32 {\scriptsize[98.29, 98.34]}}                             & \textit{3.15 {\scriptsize[3.14, 3.16]}}                             & \textit{1.40 {\scriptsize[1.38, 1.42]}}                             \\
        \cmidrule(l){2-7}
        \multirow{4}{*}{MethylBERT~\cite{jeong2025methylbert}}
                                                                                                                 & Hard labels w/ bckg.     & NNLS\textsuperscript{\textdagger}     & None                                 & 98.06 {\scriptsize[98.03, 98.08]}                                      & 3.53 {\scriptsize[3.52, 3.55]}                                      & 1.61 {\scriptsize[1.59, 1.63]}                                      \\
                                                                                                                 & Canonical Soft Labels    & NNLS\textsuperscript{\textdagger}     & None                                 & 98.12 {\scriptsize[98.10, 98.14]}                                      & 3.55 {\scriptsize[3.53, 3.56]}                                      & 1.56 {\scriptsize[1.54, 1.58]}                                      \\
                                                                                                                 & DD Soft Labels w/o pool. & SWN                                   & None                                 & 97.99 {\scriptsize[97.96, 98.01]}                                      & 3.49 {\scriptsize[3.48, 3.51]}                                      & 1.67 {\scriptsize[1.65, 1.69]}                                      \\
                                                                                                                 & DD Soft Labels w/ pool.  & SWN                                   & None                                 & \textit{\textbf{98.49 {\scriptsize[98.48, 98.51]}}}                    & \textit{3.08 {\scriptsize[3.07, 3.09]}}                             & \textit{\textbf{1.25 {\scriptsize[1.23, 1.26]}}}                    \\
        \cmidrule(l){2-7}
        \multirow{3}{*}{\shortstack[l]{Lookup Classifier                                                                                                                                                                                                                                                                                                                                                                                        \\(1-NN)}}
                                                                                                                 & Hard labels w/ bckg.     & SWN\textsuperscript{\textdagger}      & None                                 & 86.43 {\scriptsize[86.31, 86.56]}                                      & 10.29 {\scriptsize[10.25, 10.32]}                                   & 11.26 {\scriptsize[11.16, 11.37]}                                   \\
                                                                                                                 & DD Soft Labels w/o pool. & SWN                                   & None                                 & 88.62 {\scriptsize[88.49, 88.75]}                                      & 9.60 {\scriptsize[9.57, 9.64]}                                      & 9.45 {\scriptsize[9.33, 9.56]}                                      \\
                                                                                                                 & DD Soft Labels w/ pool.  & SWN                                   & None                                 & \textit{96.72 {\scriptsize[96.67, 96.76]}}                             & \textit{4.42 {\scriptsize[4.40, 4.44]}}                             & \textit{2.73 {\scriptsize[2.69, 2.77]}}                             \\
        \cmidrule(l){2-7}
        \multirow{2}{*}{CancerDetector~\cite{li2018cancerdetector}}
                                                                                                                 & Prior: Training counts   & PSLS                                  & None                                 & 98.41 {\scriptsize[98.39, 98.44]}                                      & \underline{3.05 {\scriptsize[3.04, 3.07]}}                          & 1.32 {\scriptsize[1.30, 1.34]}                                      \\
                                                                                                                 & Prior: Uniform           & PSLS                                  & None                                 & \underline{98.44 {\scriptsize[98.42, 98.46]}}                          & \textbf{3.02 {\scriptsize[3.01, 3.04]}}                             & \underline{1.30 {\scriptsize[1.28, 1.31]}}                          \\
        \midrule
        \multicolumn{7}{l}{\textit{(b) Syto variants with calibration — for each classifier and labeling/prior, the best (deconvolver + calibrator) by test MSE is reported}}                                                                                                                                                                                                                                                                   \\
        \midrule
        \multirow{4}{*}{{\shortstack[l]{Dismir~\cite{li2021dismir}                                                                                                                                                                                                                                                                                                                                                                              \\(CNN+LSTM)}}}
                                                                                                                 & Hard labels w/ bckg.     & NNLS\textsuperscript{\textdagger}     & Lin.\ simplex                        & 98.73 {\scriptsize[98.71, 98.74]}                                      & 2.92 {\scriptsize[2.91, 2.93]}                                      & 1.06 {\scriptsize[1.05, 1.07]}                                      \\
                                                                                                                 & Canonical Soft Labels    & NNLS\textsuperscript{\textdagger}     & Lin.\ simplex                        & 97.98 {\scriptsize[97.95, 98.01]}                                      & 3.39 {\scriptsize[3.38, 3.41]}                                      & 1.68 {\scriptsize[1.65, 1.70]}                                      \\
                                                                                                                 & DD Soft Labels w/o pool. & NNLS                                  & Lin.\ simplex                        & 98.89 {\scriptsize[98.88, 98.90]}                                      & 2.72 {\scriptsize[2.71, 2.73]}                                      & 0.92 {\scriptsize[0.91, 0.93]}                                      \\
                                                                                                                 & DD Soft Labels w/ pool.  & NNLS                                  & Lin.\ simplex                        & \underline{\textit{98.93 {\scriptsize[98.92, 98.94]}}}                 & \cellcolor{teal!27}\textit{\textbf{2.66 {\scriptsize[2.65, 2.67]}}} & \underline{\textit{0.89 {\scriptsize[0.88, 0.90]}}}                 \\
        \cmidrule(l){2-7}
        \multirow{4}{*}{MethylBERT~\cite{jeong2025methylbert}}
                                                                                                                 & Hard labels w/ bckg.     & NNLS\textsuperscript{\textdagger}     & Lin.\ simplex                        & 98.57 {\scriptsize[98.56, 98.59]}                                      & 3.09 {\scriptsize[3.08, 3.10]}                                      & 1.18 {\scriptsize[1.17, 1.20]}                                      \\
                                                                                                                 & Canonical Soft Labels    & PSLS                                  & Lin.\ simplex                        & 98.77 {\scriptsize[98.76, 98.78]}                                      & 3.01 {\scriptsize[3.00, 3.02]}                                      & 1.02 {\scriptsize[1.01, 1.03]}                                      \\
                                                                                                                 & DD Soft Labels w/o pool. & PSLS                                  & Lin.\ simplex                        & 98.62 {\scriptsize[98.60, 98.63]}                                      & 3.01 {\scriptsize[2.99, 3.02]}                                      & 1.15 {\scriptsize[1.14, 1.16]}                                      \\
                                                                                                                 & DD Soft Labels w/ pool.  & NNLS                                  & Lin.\ simplex                        & \cellcolor{teal!27}\textit{\textbf{98.94 {\scriptsize[98.93, 98.95]}}} & \underline{\textit{2.70 {\scriptsize[2.69, 2.71]}}}                 & \cellcolor{teal!27}\textit{\textbf{0.88 {\scriptsize[0.87, 0.89]}}} \\
        \cmidrule(l){2-7}
        \multirow{3}{*}{\shortstack[l]{Lookup Classifier                                                                                                                                                                                                                                                                                                                                                                                        \\(1-NN)}}
                                                                                                                 & Hard labels w/ bckg.     & SWN                                   & Lin.\ simplex                        & 91.56 {\scriptsize[91.46, 91.65]}                                      & 7.89 {\scriptsize[7.86, 7.92]}                                      & 7.01 {\scriptsize[6.93, 7.09]}                                      \\
                                                                                                                 & DD Soft Labels w/o pool. & NNLS                                  & Lin.\ simplex                        & 96.67 {\scriptsize[96.62, 96.72]}                                      & 4.56 {\scriptsize[4.54, 4.58]}                                      & 2.76 {\scriptsize[2.72, 2.80]}                                      \\
                                                                                                                 & DD Soft Labels w/ pool.  & PSLS                                  & Lin.\ simplex                        & \textit{98.27 {\scriptsize[98.25, 98.29]}}                             & \textit{3.27 {\scriptsize[3.26, 3.28]}}                             & \textit{1.44 {\scriptsize[1.42, 1.45]}}                             \\
        \cmidrule(l){2-7}
        \multirow{2}{*}{CancerDetector~\cite{li2018cancerdetector}}
                                                                                                                 & Prior: Training counts   & PSLS                                  & None                                 & 98.90 {\scriptsize[98.89, 98.91]}                                      & 2.74 {\scriptsize[2.73, 2.75]}                                      & 0.91 {\scriptsize[0.90, 0.92]}                                      \\
                                                                                                                 & Prior: Uniform           & PSLS                                  & None                                 & 98.91 {\scriptsize[98.90, 98.92]}                                      & 2.72 {\scriptsize[2.71, 2.73]}                                      & 0.91 {\scriptsize[0.90, 0.91]}                                      \\
        \midrule
        \multicolumn{7}{l}{\textit{(c) External baselines with and without our calibration}}                                                                                                                                                                                                                                                                                                                                                    \\
        \midrule
        \multicolumn{3}{c}{\multirow{2}{*}{\textit{CelFiE}~\cite{caggiano2021comprehensive}}}                    & None                     & 96.05 {\scriptsize[96.00, 96.10]}     & 6.42 {\scriptsize[6.40, 6.44]}       & 3.28 {\scriptsize[3.24, 3.32]}                                                                                                                                                                                     \\
                                                                                                                 &                          &                                       & Lin.\ simplex                        & \textbf{98.03 {\scriptsize[98.01, 98.06]}}                             & \textbf{3.47 {\scriptsize[3.45, 3.48]}}                             & \textbf{1.63 {\scriptsize[1.61, 1.65]}}                             \\
        \cmidrule(l){4-7}
        \multicolumn{3}{c}{\multirow{2}{*}{\textit{EpiDISH RPC}~\cite{teschendorff2017epidish}}}                 & None                     & 89.41 {\scriptsize[89.28, 89.54]}     & 10.20 {\scriptsize[10.16, 10.23]}    & 8.79 {\scriptsize[8.68, 8.90]}                                                                                                                                                                                     \\
                                                                                                                 &                          &                                       & Vec.\ scaling                        & 95.95 {\scriptsize[95.92, 95.98]}                                      & 6.02 {\scriptsize[6.00, 6.04]}                                      & 3.36 {\scriptsize[3.34, 3.38]}                                      \\
        \cmidrule(l){4-7}
        \multicolumn{3}{c}{\multirow{2}{*}{\textit{Houseman CP}~\cite{teschendorff2017epidish,houseman2012dna}}} & None                     & 90.94 {\scriptsize[90.83, 91.05]}     & 9.25 {\scriptsize[9.22, 9.28]}       & 7.52 {\scriptsize[7.42, 7.62]}                                                                                                                                                                                     \\
                                                                                                                 &                          &                                       & Lin.\ simplex                        & 96.51 {\scriptsize[96.47, 96.55]}                                      & 4.91 {\scriptsize[4.89, 4.92]}                                      & 2.90 {\scriptsize[2.86, 2.93]}                                      \\
        \cmidrule(l){4-7}
        \multicolumn{3}{c}{\multirow{2}{*}{\textit{UXM}~\cite{loyfer2023dna}}}                                   & None                     & 94.86 {\scriptsize[94.78, 94.93]}     & 6.49 {\scriptsize[6.47, 6.52]}       & 4.27 {\scriptsize[4.20, 4.34]}                                                                                                                                                                                     \\
                                                                                                                 &                          &                                       & Lin.\ simplex                        & \underline{96.74 {\scriptsize[96.68, 96.79]}}                          & \underline{4.32 {\scriptsize[4.30, 4.34]}}                          & \underline{2.71 {\scriptsize[2.66, 2.76]}}                          \\
        \bottomrule
      \end{tabular}%
    }
  }
\end{table}
}

\vspace{-2mm}

\paragraph{OOD RRBS dataset.}
\begin{wrapfigure}{R}{0.46\textwidth}
  \vspace{-6.0mm}
  \centering
  \includegraphics[trim={0.2cm 0.2cm 0.2cm 0.5cm}, clip, width=0.45\textwidth]{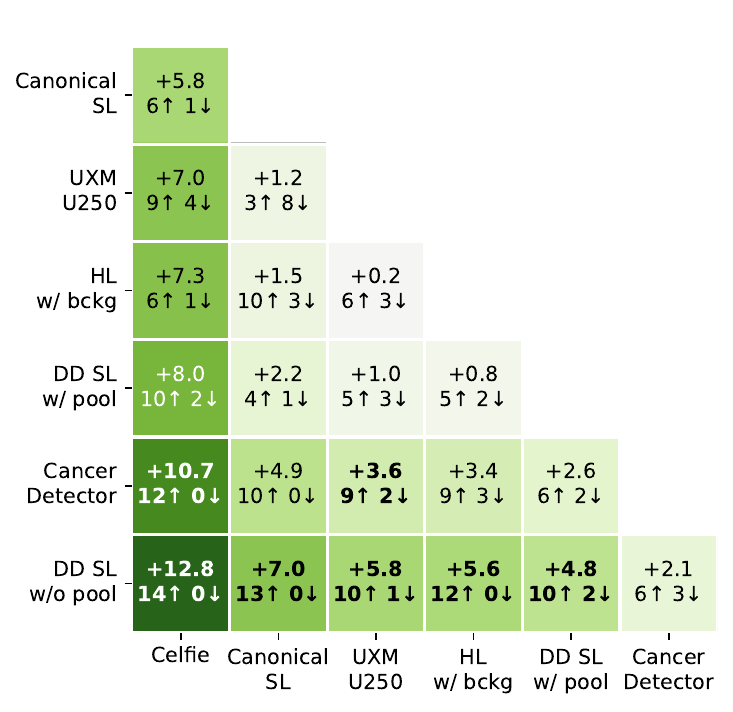}
  \caption{Pairwise mean differences in TCS (row minus column, best configuration per category).
    The values in parentheses indicate the number of tissues where significantly higher (↑) or lower (↓) TCS is observed for the row method relative to column method (Wilcoxon signed-rank test, $p \leq0.05$).}
  \label{fig:main_rrbs_pvalues}
  \vspace{-5mm}
\end{wrapfigure}

Each labeling scheme's best classifier-deconvolver-calibrator combination
yielded significantly higher TCS than the best baseline \textit{CelFiE} with
the original reference (\Cref{fig:main_rrbs_pvalues}) with \textit{Syto}
variants not using canonical soft labeling surpassing \textit{UXM U250}, which
uses $10\times$ more DMRs. Data-driven soft labeling without pooling with
\textit{Dismir} delivered a robust improvement: all 20 deconvolver-classifier
combinations outperformed \textit{CelFiE}, with 11 at $p\leq 0.05$.
\textit{CancerDetector} with training-counts prior was on par, also surpassing
\textit{CelFiE} in all 20 combinations and 11 at $p\leq 0.05$. Full counts are
in \Cref{tab:rrbs_config_counts}.

\vspace{-3mm}

\section{Discussion}

\vspace{-1mm}

\label{sect:discussion}

{\setlength{\parskip}{2pt}
    A clear pattern supported by the two experiments is that how much work the labeling
    schemes do depends on the classifier's inductive bias
    (see~\Cref{fig:soft_labels_contours,fig:methods_contours}). Starting from the
    pseudobulk experiment, our simple \textit{1\nobreakdash-NN Lookup} improves
    significantly when switched from hard labeling to data-driven soft labels and
    dramatically when pooling is enabled. \textit{Dismir} improves consistently,
    but the gain is much smaller. \textit{MethylBERT} is on par with
    \textit{Dismir} on improvement magnitude, but suffers when pooling is off. In
    the OOD experiment, \textit{1\nobreakdash-NN~Lookup} and \textit{MethylBERT}
    only work with soft labels and \textit{Dismir} is comparatively insensitive to
    the choice of labeling scheme. \textit{MethylBERT} is consistently better with
    pooling; however, its results without pooling are confounded by the necessity
    of using balanced minibatching (see \Cref{sec:supp_classifier_training}).
    \textit{1\nobreakdash-NN~Lookup} and \textit{Dismir} reach higher TCS without
    pooling, yet these gains must be viewed through the lens of TCS limitations
    (see \Cref{sect:tcs_analysis}). Importantly, soft labeling can elevate even
    limited classifiers: uncalibrated \textit{1\nobreakdash-NN~Lookup} fitted on
    soft labels with pooling and paired with \textit{SWN} deconvolver beats every
    examined uncalibrated external baseline in both experiments.

    Our labeling-free extension of \textit{CancerDetector}, while operating on
    $\alpha$-values, yields the second-best uncalibrated combination, challenging
    the need for pattern-level classifiers. This advantage is likely specific to
    the present dataset, whose short reads and selected regions make average
    methylation informative by design. The gap with pattern-sensitive methods is
    expected to increase for longer reads and more complex methylation
    patterns~\cite{jeong2025methylbert}.

    The choice of deconvolver and calibrator is less consequential than that of the
    labeling scheme and classifier, but is essential to distinguish between good
    and best combinations. \textit{NNLS} and \textit{PSLS} dominate pseudobulk
    performance when paired with our linear calibration with simplex projection,
    but loose most of their advantage in OOD\@. \textit{SWN} is stronger before
    calibration and stable across evaluation settings. \textit{XGB} performs worst
    in pseudobulk but second best in OOD, albeit with the largest variation in
    TCS\@. And \textit{MLP} is the only deconvolver absent from the Pareto front of
    the two metrics (see~\Cref{fig:joint_pareto_front,fig:methods_contours}). The
    rank correlation between pseudobulk MSE and OOD TCS is only $+0.37$, lower
    still for \textit{NNLS} and \textit{PSLS}. Therefore, deconvolver selection
    based on pseudobulk performance alone, as is common in the field, may be
    suboptimal.

    The usefulness of TCS depends on its own limitations. We screened it for the
    failure modes listed in \Cref{sect:limitations} (details in
    \Cref{sect:tcs_analysis}) and found two major artifacts: under-counted
    keratinocytes in skin and potential over-allocation to smooth muscle in the
    ovary. These penalise all methods equally in the first case and favour
    \textit{Syto} in the second; yet even when the ovary is excluded, the relative
    ranking of the labeling schemes is preserved and \textit{Syto} maintains its
    advantage over the baselines, albeit with a reduced TCS gap.

    Lastly, as discussed at length in \Cref{sect:supp_calibration_standalone}, the
    bias correction by calibrating on pseudobulk is useful beyond \textit{Syto}.
    All four examined baselines improve their scores and benefits translating to
    OOD\@. Systematic, linearly correctable bias therefore appears to be a general
    property of reference-based setups, and the calibrator can be adopted
    independently of the rest of the framework. }

\vspace{-1mm}

\section{Limitations}
\label{sect:limitations}

\vspace{-1mm}

The reported performance for learnable classifiers (\textit{Dismir},
\textit{MethylBERT}) and deconvolvers (\textit{XGB}, \textit{MLP},
\textit{SWN}) should be considered a lower bound, as optimal hyperparameters
were not pursued.

We rely on pseudobulk mixtures, a field standard. Evaluation with real samples
containing known ground-truth proportions would be preferable, but no suitable
read-level dataset was available at the time. We therefore use an additional
OOD experiment with TCS as a measure to strengthen our benchmark. TCS has
several prominent failure modes: (i) when a cell type with relatively modest
contribution is overpredicted (TCS rewards the wrong method), (ii) when a cell
type truly present in a tissue is not listed as expected (TCS penalizes the
right method), or (iii) when a cell type is listed as expected in a majority of
tissues and a method is biased to overly predict it (TCS rewards it). We
screened for these issues and found limited impact (see
\Cref{sect:tcs_analysis}).

As noted in~\cite{deridder2024benchmarking,sun2024systematic}, the choice of
genomic regions affects deconvolution performance. Our main experiments rely on
a large set of regions selected with
\texttt{wgbstools}~\cite{loyfer2026wgbstools}. We feature a supplementary
experiment focusing on a single alternative strategy for the regions selection
(see \Cref{sect:gr_sensitivity_analysis}), far from an exhaustive sensitivity
analysis. Furthermore, our training data comes from a single WGBS study based
on FACS-purified samples of 137 donors, many diagnosed with various cancers
despite atlas focus on normal human cells (samples use tumour-adjacent normal
tissue material). The purity estimates in this data are questionable,
confounded by marker selection, and are below 80\% for 17 samples out of 205.
The OOD evaluation rests on a single RRBS dataset: broader benchmarking data
would be desirable.

The 39 modeled cell types come from various levels of the ontological tree,
mostly higher in hierarchy and therefore not suitable for clinical analyses
where granular assignment is a differentiating factor. As with most
reference-based methods, \textit{Syto} in its published version cannot predict
cell types absent from the reference. Necessary extensions are foreseeable and
are planned for future work. The absence of malignant cell types in the
reference limits the application of the \textit{Syto} models trained on it.
Such models can in principle flag cancer indirectly~\cite{jeong2025methylbert},
but only as a secondary check and not as a reliable estimate. While there is
some evidence that \textit{Syto} works well with more heterogeneous and related
markers characteristic of malignant cell types (see
\Cref{sect:gr_sensitivity_analysis}), the reference must be extended with
target tumours before models trained on it can be used for tumour purity
estimation and cancer classification. We therefore frame these experiments as a
methodological proof of concept, leaving translational validation for future
work.

\vspace{-1mm}

\vspace{-2mm}

\section{Conclusion}

\vspace{-2mm}

This study presents \textit{Syto}, a modular framework for cell-type
classification. It extends and includes existing read-level classifiers to
support a broad range of cell types and introduces learnable deconvolvers and
post-hoc calibrators. We propose several labeling schemes to enable
convergence in large multi-class settings. Our data-driven soft-labeling
approach outperforms all other evaluated labeling strategies, including our
generalization of hard labels with background, achieving superior results in
both pseudobulk and OOD scenarios, and surpassing the best examined baseline
\textit{CelFiE} by a substantial margin. 
Beyond improving deconvolution accuracy, scaling read-level approaches to larger 
reference panels enables analyses that retain cell-type information at the level of 
individual reads, providing access to local epigenetic variation that 
aggregate-based methods cannot capture.
The promising results observed across 39
healthy cell types suggest the feasibility of progressively expanding to larger
and more complex reference panels. A primary extension involves incorporating
malignant cell types, enabling \textit{Syto} to address tumor classification
and purity estimation as specific instances of deconvolution. Furthermore, 
integration of multi-omics signatures and
third-generation long-read sequencing data will facilitate the capture
of more complex epigenetic patterns and enhance differentiating capabilities, 
potentially enabling new biological insights and, following appropriate validation, 
supporting clinical decision-making in diseases associated with epigenetic alterations.

\label{sect:funding}
\begin{ack}
We acknowledge funding from the KU Leuven BOFZAP starting grant (STG/22/024) and FWO Runner-up project MUTAHIST-PoP (G0ATW25N) to P.L.
\end{ack}

{
  \small
  \bibliographystyle{unsrtnat}
  \bibliography{references}
}

\newpage
\appendix
\setcounter{figure}{0}
\renewcommand{\thefigure}{S\arabic{figure}}
\setcounter{table}{0}
\renewcommand{\thetable}{S\arabic{table}}

\newpage
\section{Source code and experiments data}
\label{sect:supp_code_and_data}
The production code that can be used to reproduce the paper results is distributed in this frozen release: \url{https://github.com/CompEpigen/syto/releases/tag/v1.0.0}.

The run configs for every step in each experiment, along with preprocessed data, fitted models, and generated pseudobulks, are deposited in the KU Leuven Research Data Repository (DOI: \href{https://doi.org/10.48804/JM4BOS}{10.48804/JM4BOS}).
The full mirror of the same data can be found at \url{https://huggingface.co/datasets/CompEpigen/syto.1.0}.

\newpage

\section{Supplementary material on methods}
\label{sect:supp_methods}

\subsection{General framework}

\begin{figure}[H]
  \centering
  \includegraphics[width=1\textwidth]{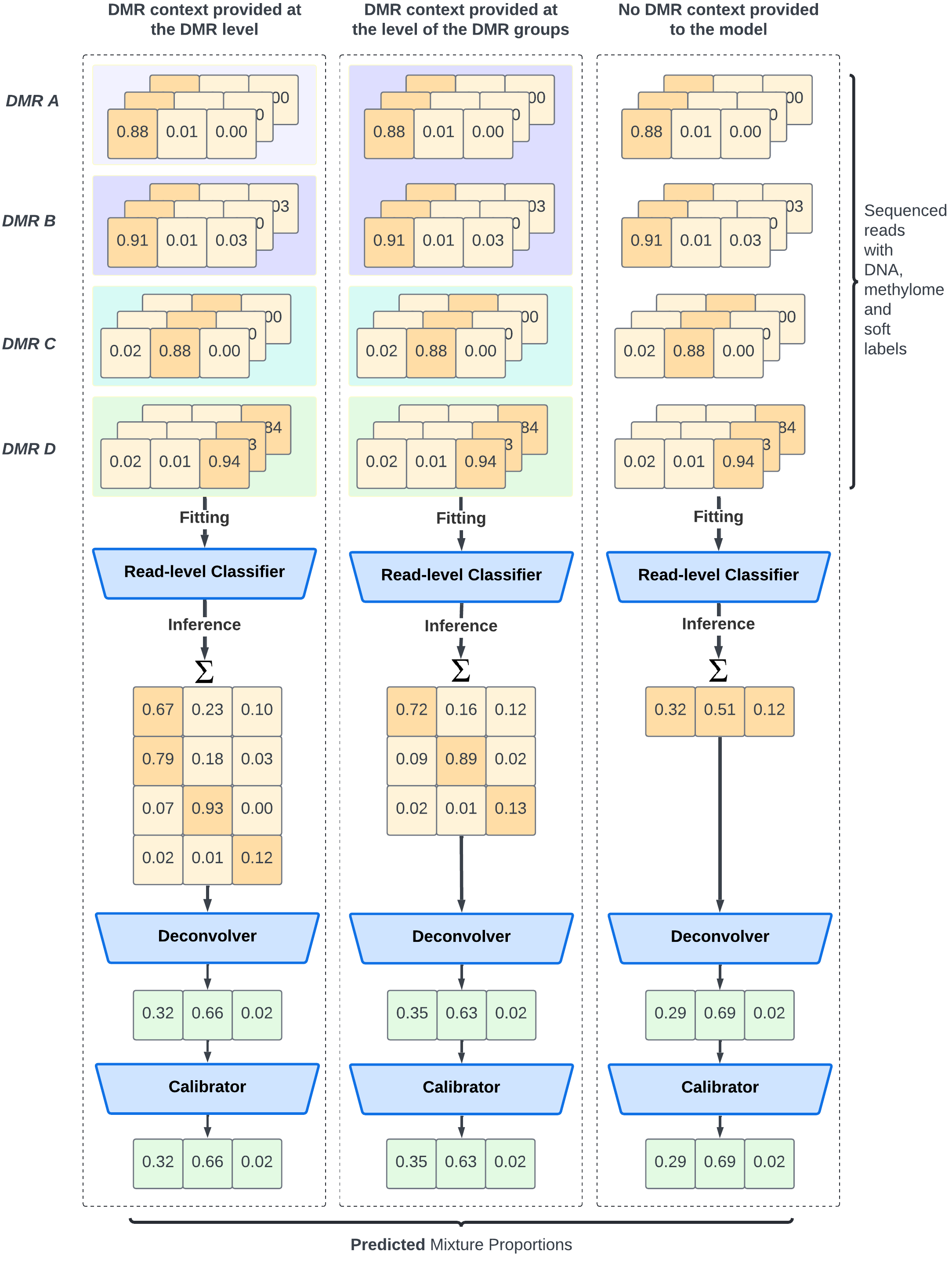}
  \caption{Implementation of proposed framework for different reference aggregation levels.
    This work utilizes DMR groups (center).}
  \label{fig:framework_specific}
\end{figure}

\subsection{Jaccard distance-based smoothing of soft labels}
\label{sect:supp_jaccard_distance}

As mentioned in the main paper, for signatures with low read coverage, we pool
the counts of similar signatures using the Jaccard distance. In this section,
we describe precisely how we apply the Jaccard distance to signatures.

In our setting, we define an omics signature as a set of pairs (position,
methylation status) where position is the position of the CpG site in the human
genome (in accordance with selected reference) and methylation status is either
0 for unmethylated or 1 for methylated. Since every signature is a set, we can
define the Jaccard distance between two signatures \(s_1\) and \(s_2\) in the
classical way as:

\begin{equation}
  \label{eq:jaccard_distance}
  \textsc{JaccardDist}(s_1, s_2) \coloneq 1 - \frac{|s_1 \cap s_2|}{|s_1 \cup s_2|} = \frac{| s_1 \Delta s_2 |}{|s_1 \cup s_2|},
\end{equation}

where \(\Delta\) denotes the symmetric difference between sets and \(|\cdot|\)
denotes the cardinality of a set.

\subsection{Sensitivity of \textit{Syto} to the choice of pooling parameters} \label{sect:supp_pooling_sensitivity}

In our data-driven soft labeling approach, two parameters control how the
pooling is performed: the Jaccard distance threshold $\delta$ and the minimum
count threshold $\tau$ to reach per signature.

In this section, we discuss the heuristics that lead to the choice of these
parameters and we show the results of a sensitivity analysis on the choice of
these parameters.

In all of our experiments, we set $\delta=0.41$ and $\tau=30$. Those choices
were made based on heuristics derived from the training data:
\begin{itemize}
    \item Distance threshold of $\delta=0.41$: More than 80\% of reads in our training
          data contain 4--7 CpGs. A Jaccard distance threshold of 0.41 allows reads in
          this range to differ by at most one CpG, reflecting the high local correlation
          of CpG methylation. Differences in two or more CpGs are more likely to
          represent a biological change. This threshold choice groups 4-CpG reads with
          methylation levels of 0 and 0.25, as well as 0.75 and 1.0, similarly to how UXM
          operates.
    \item Minimum count threshold of $\tau=30$: We argue that the required read depth
          should depend on the number of modeled classes while still providing sufficient
          support. Setting the threshold too high may unnecessarily discard informative
          variation. We used 30 reads, slightly below the number of cell types.
\end{itemize}

We additionally perform a sensitivity analysis on the choice of these
parameters. We varied each of the two parameters $\delta$ and $\tau$ while
keeping the other fixed to the default value. We trained the \textit{Syto}
variant Dismir + Soft Labeling with Pooling. To isolate the effect of the
pooling parameters as much as possible, we did not apply calibration. All other
experimental settings were kept the same as in the main pseudobulk experiment.

\Cref{tab:supp_pooling_sensitivity} shows the results of this sensitivity analysis.
First of all, we observe that, overall, the variations in performance between different
parameter settings are small compared to the difference with external baselines
(CelFiE --- the previous SoTA --- had an MSE of $3.27\times 10^{-4}$),
which hints at the robustness of our method to the pooling parameters.

We also observe that the best pooling parameters are different depending on the
deconvolver. However, if we focus on \textit{NNLS} and \textit{PSLS}, which
have the desirable property of having no training procedure and thus are not
prone to variability due to random initialization and training dynamics
(contrarily to \textit{XGB} \footnote{XGB is conditionally determenistic, but
    it is also consistently worse than \textit{NNLS} and \textit{PSLS}},
\textit{SWN} and \textit{MLP}), we see that our heuristic choice landed in a
sweet spot for both parameters, confirming our original intuition.

\begin{table}[htbp]
    \centering
    \caption{MSE ($\times 10^{-4}$) and 95\% confidence intervals for varying distance threshold $\delta$ and minimum count threshold $\tau$ parameters for
        Dismir + Soft Labels with Pooling for all deconvolvers, without calibration.
        Best results per deconvolver are in \textbf{bold}, second best are \underline{underlined}.}
    \label{tab:supp_pooling_sensitivity}
    {\footnotesize
        \resizebox{\textwidth}{!}{%
            \begin{tabular}{l ccccc}
                \multirow{2}{*}{\textbf{Deconvolver}} & $\mathbf{\boldsymbol{\delta}=0.41}$      & $\mathbf{\boldsymbol{\delta}=0.2}$    & $\mathbf{\boldsymbol{\delta}=0.41}$   & $\mathbf{\boldsymbol{\delta}=0.6}$ & $\mathbf{\boldsymbol{\delta}=0.41}$ \\
                                                      & $\mathbf{\boldsymbol{\tau}=15}$          & $\mathbf{\boldsymbol{\tau}=30}$       & $\mathbf{\boldsymbol{\tau}=30}$       & $\mathbf{\boldsymbol{\tau}=30}$    & $\mathbf{\boldsymbol{\tau}=45}$     \\
                \midrule
                \textit{NNLS}                         & \underline{1.53 \scriptsize[1.50, 1.55]} & 1.65 \scriptsize[1.62, 1.68]          & \textbf{1.47 \scriptsize[1.45, 1.50]} & 1.66 \scriptsize[1.64, 1.69]       & 1.61 \scriptsize[1.58, 1.64]        \\
                \textit{PSLS}                         & \underline{1.60 \scriptsize[1.58, 1.63]} & 1.79 \scriptsize[1.75, 1.82]          & \textbf{1.55 \scriptsize[1.52, 1.57]} & 1.73 \scriptsize[1.70, 1.76]       & 1.70 \scriptsize[1.67, 1.74]        \\
                \textit{MLP}                          & \underline{2.16 \scriptsize[2.11, 2.20]} & \textbf{1.60 \scriptsize[1.58, 1.62]} & 2.17 \scriptsize[2.14, 2.19]          & 2.22 \scriptsize[2.19, 2.25]       & 1.91 \scriptsize[1.89, 1.93]        \\
                \textit{SWN}                          & \underline{1.24 \scriptsize[1.23, 1.26]} & \textbf{1.20 \scriptsize[1.19, 1.22]} & 1.40 \scriptsize[1.38, 1.42]          & 1.67 \scriptsize[1.63, 1.70]       & 1.49 \scriptsize[1.46, 1.52]        \\
                \textit{XGB}                          & \underline{2.10 \scriptsize[2.06, 2.14]} & \textbf{1.98 \scriptsize[1.95, 2.01]} & 2.12 \scriptsize[2.08, 2.16]          & 2.27 \scriptsize[2.23, 2.32]       & 2.18 \scriptsize[2.15, 2.22]        \\
                \bottomrule
            \end{tabular}%
        }}
\end{table}

\subsection{Multiclass extension of \texorpdfstring{\textit{CancerDetector}}{CancerDetector}}
\label{sect:supp_mat_cancer_detector}

\textit{CancerDetector} is a method by \citet{li2018cancerdetector} to estimate the proportion of reads originating from tumor cells in a bulk sample of cell-free DNA (cfDNA).

\paragraph{\textit{CancerDetector} for binary deconvolution.} In the original method, the \textbf{training phase}
proceeds in two steps. First, identifying methylation markers specific to the
cancer under scrutiny by identifying eligible markers --- \emph{i.e.} markers
with sufficient coverage --- and then keeping the markers whose methylation
levels can differentiate tumor samples from normal tissue samples and normal
plasma samples. Second, for each marker, for each class (tumoral or normal),
the parameters of the beta distribution of the \(\alpha\)-values of the reads
in that marker from the given class are computed by maximum likelihood
estimation (MLE).

In the \textbf{deconvolution phase}, for each read in the bulk sample, the
likelihood of the read to belong to each class is computed according to the
beta distribution of the each class and the marker from which the read
originates. Finally, the proportion of each class in the mixture is computed by
MLE.

\citet{li2018cancerdetector} also rely on the prior knowledge that they are deconvolving a mixture of normal and tumoral reads where the tumoral reads are expected to be in low proportion. They apply a process called ``removal of `confounding' markers'' where they estimate the tumor proportion of each marker and iteratively remove the markers whose tumor proportion is more than one standard deviation above the mean tumor proportion, until convergence.

\paragraph{\textit{CancerDetector} for multiclass classification.} We extend \textit{CancerDetector} to the
multiclass classification setting of the \citet{loyfer2023dna} dataset. The
original marker selection is replaced by the selection of DMR already performed
for the construction of the dataset~\cite{loyfer2023dna}.

The dataset \(\mathcal{D} \coloneqq \{(r_i, g_i, s_i, c_i)\}_{i=1}^N\) consists
of \(N\) independent and identically distributed (i.i.d.) samples, where
\(r_i\) is the read, \(g_i\) is the GR group to each $r_i$ belong, $s_i$ is the
omics signature of $r_i$ and \(c_i\) is its class label. The independence
assumption is common in the literature
\cite{li2018cancerdetector,jeong2025methylbert}. Since the particular position
is irrelevant in this section, we will model the signatures as the tuples of
the methylation states of the reads' CpG sites.

For a given read $r_i$, its signature \(s_i \coloneq (s_{i,1}, \ldots,
s_{i,L_i}) \in \{0, 1\}^{L_i}\) is modeled by the Beta-Bernouilli distribution,
as in~\cite{li2018cancerdetector}. Specifically, the methylation state
\(s_{i,j}\) is distributed as \(s_{i,j} \sim \text{Bernouilli}(p)\) where \(p
\sim \text{Beta}(\eta_{g_i}^{c_i}, \rho_{g_i}^{c_i})\) is the prior of
methylation rates of CpG sites for the reads in GR group \(g_i\) and
originating from cell type \(c_i\).

During the \textbf{training phase}, for every GR group \(g \in \mathcal{G}\)
that contains at least one read in the dataset, and for every classe \(c \in
\mathcal{C}\), we estimate the parameters $\eta_g^c$ and $\rho_g^c$ of the Beta
distribution \(\text{Beta}(\eta_g^c, \rho_g^c)\), which models the
\(\alpha\)-values of the reads belonging to GR group \(g\) with class label
\(c\). We estimate the parameters \(\eta_g^c\) and \(\rho_g^c\) by maximum
likelihood estimation (MLE) when there is at least 3 reads for the GR
group-class combination, and when all reads do not have the same methylation
rate. In case there are at least 3 reads but all of them have the same
methylation rate (which happens in our data only with methylation rate 0 and
1), then the MLE cannot converge, and so we fallback to bayesian estimation
with uniform prior. In case there is 2 reads or less for the GR group-class
combination, we deem the distribution as being uniform.

The \textbf{classification phase} consists in:
\begin{enumerate}
      \item Filtering the reads of the bulk sample to deconvolve by keeping only the reads
            in GR groups that were seen in the training data, obtaining the following
            dataset: \(\mathcal{D}' \coloneq \{(r_i, g_i, s_i)\}_{i=1}^{N'}\).
      \item Computing, for all \( (r_i, g_i, s_i) \in \mathcal{D}'\), the likelihood of the
            read to belong to each class \(c \in \mathcal{C}\) according to the beta
            distribution parameters of GR group \(g_i\) and class \(c\):
            \begin{align}
                  \ell(r_i|c) & = \prod_{j=1}^{L_i} P(s_{i,j} \mid \text{Beta}(\eta_{g_i}^c, \rho_{g_i}^c)),                                                                                                   \\
                              & = \prod_{j=1}^{L_i} \int_{0}^{1}p^{s_{i,j}}(1-p)^{1-s_{i,j}} \frac{p^{\eta_{g_i}^c - 1}(1-p)^{\rho_{g_i}^c - 1}}{B(\eta_{g_i}^c, \rho_{g_i}^c)} dp,                            \\
                              & = \prod_{j=1}^{L_i} \frac{B(s_{i,j} + \eta_{g_i}^c, 1 - s_{i,j} + \rho_{g_i}^c)}{B(\eta_{g_i}^c, \rho_{g_i}^c)},                                                               \\
                              & = \frac{B(1 + \eta_{g_i}^c, \rho_{g_i}^c)^{\sum_{j=1}^{L_i} s_{i,j}} B(\eta_{g_i}^c, 1 + \rho_{g_i}^c)^{\sum_{j=1}^{L_i} (1 - s_{i,j})}}{B(\eta_{g_i}^c, \rho_{g_i}^c)^{L_i}},
            \end{align}

            where \(B\) is the Beta function.
      \item Computing, for all \( (r_i, g_i, s_i) \in \mathcal{D}'\), the posterior
            probability of the read to belong to each class \(c \in \mathcal{C}\) according
            to Bayes' rule:
            \begin{equation}
                  P(c \mid r_i, g_i, s_i) = \frac{\ell(r_i|c) P(c)}{\sum_{c' \in \mathcal{C}} \ell(r_i|c') P(c')},
            \end{equation}
            where \(P(c)\) is the prior probability of class \(c\).
\end{enumerate}

In our experiments, we use as class prior both the distribution of classes in
the training data (which is the mathematically correct choice) and the uniform
distribution.

\subsection{GR-group-attention classification head}
\label{supp:sect_dmr_attention_head}

We replace the vanilla fully connected classification head of
\textit{MethylBERT}~\cite{jeong2025methylbert} with a cross-attention mechanism
conditioned on the GR group identity. A preliminary comparison of the two
classification heads showed slightly better losses for our formulation, which
we use in all experiments. The intuition is that the same methylation pattern
may be informative for a given cell type in one GR group but uninformative in
another; conditioning on the GR group allows the classifier to attend to the
relevant sequence positions accordingly.

Let $H = (\mathbf{h}_1, \dots, \mathbf{h}_L) \in \mathbb{R}^{L \times d}$ be
the encoder output for a read of length~$L$, and let $g \in \mathcal{G}$ be the
GR group label. We embed $g$ into a learnable vector $\mathbf{e}_g =
    E_{\text{gr}}[g] \in \mathbb{R}^d$, where $E_{\text{gr}} \in \mathbb{R}^{G
        \times d}$ is the GR embedding table.

\paragraph{Cross-attention.}
We compute a single-head cross-attention in which the GR group embedding serves
as the query and the encoder outputs serve as keys and values:
\begin{align}
    \mathbf{q} & \coloneq W_Q\,\mathbf{e}_g \in \mathbb{R}^d, \qquad
    \mathbf{k}_j \coloneq W_K\,\mathbf{h}_j \in \mathbb{R}^d, \qquad
    \mathbf{v}_j \coloneq W_V\,\mathbf{h}_j \in \mathbb{R}^d, \quad \forall j \in \{1, \dots, L\},                                                                                                                                  \\
    \alpha_j   & \coloneq \frac{\exp\!\bigl(\mathbf{q}^\top \mathbf{k}_j / \sqrt{d}\,\bigr)}{\sum_{j'=1}^{L} \exp\!\bigl(\mathbf{q}^\top \mathbf{k}_{j'} / \sqrt{d}\,\bigr)} \in \Delta^{L-1}, \quad \forall j \in \{1, \dots, L\}, \\
    \mathbf{z} & \coloneq \sum_{j=1}^{L} \alpha_j \mathbf{v}_j \in \mathbb{R}^d,
\end{align}
where $W_Q, W_K, W_V \in \mathbb{R}^{d \times d}$ are learnable projection matrices for queries, keys and values respectively, $\boldsymbol{\alpha}$ is the attention distribution over sequence positions and $\mathbf{z}$ is the attended representation.
Positions corresponding to padding tokens are masked by setting their pre-softmax logits to $-10^{4}$.

\paragraph{Context fusion.}
The attended representation $\mathbf{z}$ is concatenated with the original GR
group embedding and passed through a feed-forward fusion block:
\begin{equation}
    \mathbf{f} \coloneq \mathrm{Dropout}\!\Bigl(\mathrm{ReLU}\!\bigl(\mathrm{LayerNorm}(W_f\,[\mathbf{z};\,\mathbf{e}_g] + \mathbf{b}_f)\bigr)\Bigr),
\end{equation}
where $W_f \in \mathbb{R}^{d \times 2d}$ and $\mathbf{b}_f \in \mathbb{R}^{d}$ are the parameters of the linear layer, and $[\,\cdot\,;\,\cdot\,]$ denotes concatenation.

\paragraph{Classifier.}
The fused features are projected to class logits via a two-layer \textit{MLP}:
\begin{equation}
    \boldsymbol{\ell} \coloneq W_2\,\mathrm{Dropout}\!\bigl(\mathrm{ReLU}(W_1\,\mathbf{f} + \mathbf{b}_1)\bigr) + \mathbf{b}_2,
\end{equation}
where $W_1 \in \mathbb{R}^{(d/2) \times d}$, $W_2 \in \mathbb{R}^{C' \times (d/2)}$, $\mathbf{b}_1 \in \mathbb{R}^{d/2}$ and $\mathbf{b}_2 \in \mathbb{R}^{C'}$ are the parameters of the \textit{MLP}, and $C'$ equals~$C$ under soft labeling or $C+1$ under hard labeling with background class.

As indicated in the main paper, we extended use of this mechanism to
\textit{Dismir}~\cite{li2021dismir}, which lacks any conditioning on GR groups
as model input in its original implementation. Depending on the underlying
classifier, the attention weights $\boldsymbol{\alpha}$ provide means for
interpretability, indicating which encoded positions are most relevant for
classification given the GR group context.

\subsection{Read-Level Classifier Training}
\label{sec:supp_classifier_training}

Both \textit{Dismir}~\cite{li2021dismir} and \textit{MethylBERT}
\cite{jeong2025methylbert} use the GR group-attention-based classification
head. This replaces the original vanilla fully connected classifier with a
cross-attention mechanism conditioned on GR group label
(cf.~\Cref{supp:sect_dmr_attention_head}). Input sequences are fixed to a
maximum length of 160 nucleotides (\textit{Dismir}) or 160 k-mers
(\textit{MethylBERT}).

\paragraph{\textit{Dismir}.}
The \textit{Dismir} encoder consists of a 1D convolution (100 filters, kernel size 10) followed by ReLU activation and max-pooling, a bidirectional LSTM with hidden size 75 per direction, a second 1D convolution (100 filters, kernel size 3) with max-pooling and dropout ($p = 0.2$) after each pooling layer.
The encoder output is passed to the GR group-attention classifier.

Training used the Adam optimizer with a learning rate of $10^{-3}$, weight
decay of $10^{-6}$, batch size of 128, and early stopping with patience of 7
epochs over a maximum of 100 epochs. For the soft-labeling scheme, the model
was trained with 39 output classes using confidence-weighted cross-entropy
loss. For the hard-labeling scheme, the model was trained with 40 output
classes (39 cell types plus a background class) using standard cross-entropy
loss.

\paragraph{\textit{MethylBERT}.}
We fine-tuned the 12-layer pre-trained \textit{MethylBERT} model (\texttt{methylbert\_hg19\_12l}) with the GR group-attention-based classification head.
Fine-tuning used the AdamW optimizer with mixed-precision (FP16) training. The learning rate was $4 \times 10^{-4}$, $(\beta_1, \beta_2) = (0.9, 0.98)$ and $\epsilon = 10^{-6}$. Weight decay was 0.1. A linear warmup of 100 steps was used. Gradient clipping was set at 1.0.
The batch size was 512 per device, with evaluation and checkpointing every 200 steps, early stopping patiance of 1000 steps over a maximum of 5 epochs. The best checkpoint was selected based on validation loss.

For the data-driven soft-labeling scheme, the model was trained with 39 output
classes using confidence-weighted cross-entropy loss. For the hard-labeling and
canonical soft-labeling schemes, the model was trained with 40 output classes
(39 cell types plus a background class) using standard cross-entropy loss. We
additionally applied minibatch balancing for \textit{MethylBERT} when trained
using canonical soft, hard, and data driven soft labels without pooling. Each
training batch was composed of 50\% on-target reads (reads originating from the
cell type corresponding to their GR group) and 50\% background reads. We used a
custom batch sampler that cycles through the background pool to maintain this
ratio throughout training.

\subsection{MLE deconvolvers}
\label{sect:supp_mle_deconvolvers}

In previous works, the mapping from the classification results to the mixture
proportions was systematically performed via a form of Maximum Likelihood
Estimation (MLE)~\cite{li2018cancerdetector,li2021dismir,jeong2025methylbert}.
In this section, we derive two MLE formulations for multiclass deconvolution
and compare their performance with \textit{UXM} and our deconvolvers. The main
takeaway is that MLE deconvolvers produce results that are orders of magnitude
worse than other deconvolvers, which is why they are not featured in the main
paper.

We use the notations introduced in the main paper. We additionaly assume the
bulk biological sample from which the set of reads $\{{(r_i, g_i)}\}_{i \in {1,
        \dots, N}}$ originates has true underlying cell types proportions
$\boldsymbol{\theta} \coloneq (\theta_1, \dots, \theta_C) \in \Delta^{C-1}$,
where $\theta_c$ is the proportion of cell type $c$ in the sample, $r_i$ is the
read and $g_i$ is the GR group to which $r_i$ belongs. For all GR group $g$, we
denote by $c^{(g)}$ the cell type for which it is specific.

\paragraph{MethylBERT-style MLE.}
The way \citet{jeong2025methylbert} compute likelihood $L^{MB}$ is by modeling,
for each read $r_i$ falling in GR group $g_i$, the probability $P(c^{(g_i)}
    \mid r_i )$ that the read belong to cell type $c^{(g_i)}$. In our framework,
the classifier outputs probabilities for all classes: $\mathbf{u_i} \in
    \Delta^{C-1}$, and in particular $P(c^{(g_i)} \mid r ) = u_{i,c^{(g_i)}}$. We
denote by $R_c \coloneq \{ i \mid c^{g_i} = c \}$ the set of indexes of reads
belonging to a GR group specific to cell type $c$. By independence of the
reads\footnote{The independence assumption is common in the literature
    ~\cite{li2018cancerdetector,jeong2025methylbert}.} and Bayes' theorem:

\begin{align}
    L^{MB}(\boldsymbol{\theta}) & = \prod_{c=1}^C \prod_{i \in R_c} \left[\theta_c P(r_i \mid c) + (1-\theta_c) P(r_i \mid \text{not } c) \right],                                                   \\
                                & = \prod_{c=1}^C \prod_{i \in R_c} \left[\theta_c \frac{P(c \mid r_i)P(r_i)}{P(c)} + (1-\theta_c) \frac{P(\text{not } c \mid r_i)P(r_i)}{P(\text{not } c)} \right], \\
                                & = \prod_{c=1}^C \prod_{i \in R_c} P(r_i) \left[\theta_c \frac{P(c \mid r_i)}{P(c)} + (1-\theta_c) \frac{1 - P(c \mid r_i)}{1 - P(c)} \right].
\end{align}

It should be noted that this formulation of the likelihood is the one used in
\textit{MethylBERT}
\textit{implementation}\footnote{\href{https://github.com/CompEpigen/methylbert/blob/main/src/methylbert/deconvolute.py}{https://github.com/CompEpigen/methylbert/blob/main/src/methylbert/deconvolute.py}},
but it differs from what is \textit{written} in Equation (19) of the original
paper. In fact, there are two errors related to this Equation (19): the first
is that $P(r_i \mid \text{not } c)$ is equated with $1-P(r_i\mid c)$, which is
true iff $P(c)=0.5$, but which is false in general. The second error is that in
the paper, $R_c$ is defined as the set of reads classified as $c$, whereas in
the implementation, it is the set of reads that fall into GR groups specific to
$c$.

The log-likelihood writes as follows:

\begin{equation}
    \log L^{MB}(\boldsymbol{\theta}) = \sum_{c=1}^C \sum_{i \in R_c} \log \left( \theta_c \frac{u_{i,c} - P(c)}{P(c)(1-P(c))} + \frac{1 - u_{i,c}}{1 - P(c)} \right) +  \sum_{c=1}^C \sum_{i \in R_c} \log P(r_i).
\end{equation}

The $P(r_i)$ terms factor out and do not play any role in the optimization. The
prior terms $P(c)$ are computed as the ratios of classes in the data
effectively processed by the classifier during training (it is only different
from the ratio of class in the training set when we use minibatch balancing).
We also experiment with a uniform prior. The maximum likelihood estimates
$\boldsymbol{\theta}^*$ are obtained by solving the following convex
optimization problem using the Python package SciPy:

\begin{align}
    \boldsymbol{\theta}^* & = \underset{\boldsymbol{\theta}}{\textrm{argmax}} \log L^{MB}(\boldsymbol{\theta}),                     \\
                          & \textrm{s.t.} \quad \sum_{c=1}^{C} \theta_c = 1, \qquad \theta_c\geq 0, \quad \forall c\in \mathcal{C},
\end{align}

We apply the MLE deconvolver to pseudobulks corresponding to pure mixtures
(\textit{i.e.}, a single cell type in the mixture), the simplest mixtures to
deconvolve. We use 3 different classifiers settings to generate the classifier
outputs on which we apply the MLE deconvolver. First is \textit{MethylBERT}
trained with focal loss on hard labels with background, second is
\textit{MethylBERT} trained with minibatch balancing on hard labels with
background and third is \textit{MethylBERT} trained with focal loss on the
binary label scheme of \citet{jeong2025methylbert} (where the label of a read
is 1 if it belongs to the cell type specific to its GR group and 0 otherwise).
The last setting is the one used in the \textit{MethylBERT} paper
\cite{jeong2025methylbert}. Results are presented in
\Cref{tab:supp_mle_methylbert_style_deconvolver_results}.

\begin{table}[hbt!]
    \centering
    \caption{Deconvolution performances of \textit{UXM} and of MLE \textit{MethylBERT}-style deconvolver when applied on 39 pure mixtures composed from the test set predictions of \textit{MethylBERT} for two training regimes (focal loss on hard labels with background class and hard labels with background class with balanced minibatches).}
    \label{tab:supp_mle_methylbert_style_deconvolver_results}
    {\footnotesize
        \resizebox{\textwidth}{!}{%
            \begin{tabular}{c cccccccc}
                \toprule
                \multirow{2}{*}{\textbf{Deconvolver}} & \multirow{2}{*}{\textbf{Training of MethylBERT}} & \multirow{2}{*}{\textbf{Prior}} & $\mathbf{R}^\mathbf{2}$ & \textbf{LoA}       & \textbf{LoA worst-class} & \textbf{MAE}       & \textbf{MSE}       & \textbf{KL}        \\
                                                      &                                                  &                                 & ($\times 10^{-2}$)      & ($\times 10^{-2}$) & ($\times 10^{-2}$)       & ($\times 10^{-3}$) & ($\times 10^{-4}$) & ($\times 10^{-2}$) \\
                \midrule
                MLE (\textit{MethylBERT}-style)       & Hard labels w/ Focal loss                        & Uniform prior                   & -10.29                  & [-32.55, 32.55]    & [-33.95, 28.82]          & 49.09              & 275.55             & 1313.22            \\
                MLE (\textit{MethylBERT}-style)       & Hard labels w/ Focal loss                        & Train freq prior                & -8.02                   & [-32.21, 32.21]    & [-27.64, 35.76]          & 49.08              & 269.87             & 1166.60            \\
                MLE (\textit{MethylBERT}-style)       & Hard labels w/ Minibatch balancing               & Uniform prior                   & 33.12                   & [-25.34, 25.34]    & [-33.83, 28.73]          & 37.83              & 167.08             & 149.94             \\
                MLE (\textit{MethylBERT}-style)       & Hard labels w/ Minibatch balancing               & Effective train freq prior      & 39.56                   & [-24.09, 24.09]    & [-33.36, 28.75]          & 36.82              & 151.00             & 139.39             \\
                MLE (\textit{MethylBERT}-style)       & Binary labels w/ Focal loss                      & Train freq prior                & 94.76                   & [-7.09,7.09]       & [-26.80, 23.47]          & 6.23               & 13.09              & 16.44              \\
                MLE (\textit{MethylBERT}-style)       & Binary labels w/ Focal loss                      & Uniform prior                   & 92.34                   & [-8.58,8.58]       & [-31.30, 26.70]          & 7.27               & 19.15              & 21.63              \\
                \textit{UXM}                          & N/A                                              & N/A                             & 98.69                   & [-3.54, 3.54]      & [-9.85, 8.46]            & 3.74               & 3.27               & 7.86               \\
                \bottomrule
            \end{tabular}%
        }}
\end{table}

The performance is worse than \textit{UXM}. At first glance, this may seem
surprising, as \citet{jeong2025methylbert} report that the MLE deconvolver can
deconvolve five cell types with performance similar to \textit{UXM} (panel H,
Figure 4 in~\cite{jeong2025methylbert}). However, the experimental setup is not
described in sufficient detail for us to reproduce this comparison, and several
aspects of the two settings differ. First, our experiments consider 39 cell
types covering the whole body, whereas the \textit{MethylBERT} study considers
only five leukocyte cell types. Second, the reported outputs are compared with
\textit{UXM}'s estimates rather than with the ground-truth mixture proportions.
Third, the pseudobulk proportions in the \textit{MethylBERT} study span
relatively narrow ranges: \emph{Blood-B} and \emph{Blood-NK} are always present
at less than 10\%, while \emph{Blood-Mono+Macro} is restricted to a narrow
range. \emph{Blood-Granul} and \emph{Blood-T} are the only cell types with
substantial variation in their proportions, and neither spans the full $[0,1]$
range. This differs substantially from our setting, in which all 39 cell types
can occur at arbitrary proportions.

\paragraph{General MLE.} The extension of likelihood to multiclass by \citet{jeong2025methylbert} is
actually based on the binary class setting, and relies on the fact that in the
original paper, \textit{MethylBERT} only outputs the probability of a read
being of the cell type which is specific to the GR group the read belongs to.
We derive the likelihood for multiclass in the context where for a single read
$r$, the model outputs a posterior probability distribution $P(c \mid r)$ for
all classes $c$, also using the read independence assumption and Bayes'
theorem:
\begin{align}
    L(\boldsymbol{\theta}) & = \prod_{i=1}^N P(r_i \mid \boldsymbol{\theta}) = \prod_{i=1}^N \sum_{c=1}^C \theta_c P(r_i \mid c),                                      \\
                           & = \prod_{i=1}^N \sum_{c=1}^C \theta_c \frac{P(c \mid r_i)P(r_i)}{P(c)} = \prod_{i=1}^N P(r_i) \sum_{c=1}^C \theta_c \frac{u_{i,c}}{P(c)}.
\end{align}

The $P(r_i)$ terms factorize out in the log-likelihood and thus play no role in
the optimization:

\begin{equation}
    \log L(\boldsymbol{\theta}) = \sum_{i=1}^N \log P(r_i) + \sum_{i=1}^N \log \left( \sum_{c=1}^C \theta_c \frac{u_{i,c}}{P(c)} \right).
\end{equation}

We obtain the mixture proportion estimates by minimizing the negative
log-likelihood subject to positivity and sum-to-one constraints, using the
Python package SciPy. \Cref{tab:supp_mle_general_deconvoler_results} shows the
performance of the main paper deconvolvers (without calibration) and of this
general MLE deconvolver with two different priors: the class frequency in the
train set (which is the mathematically correct choice), and the uniform prior.
As in the previous experiment, the performance of MLE methods is orders of
magnitude worse than that of other methods.

\begin{table}[tbhp]
    \centering
    \caption{Deconvolution performances for UXM and MLE general deconvolver when applied on 39 pure pseudobulks composed from the test set predictions of \textit{MethylBERT} with soft labels with pooling.}
    \label{tab:supp_mle_general_deconvoler_results}
    {\footnotesize
        \resizebox{\textwidth}{!}{%
            \begin{tabular}{c cccccc}
                \toprule
                \multirow{2}{*}{\textbf{Deconvolver}} & $\mathbf{R}^\mathbf{2}$ & \textbf{LoA}       & \textbf{LoA worst-class} & \textbf{MAE}       & \textbf{MSE}       & \textbf{KL}        \\
                                                      & ($\times 10^{-2}$)      & ($\times 10^{-2}$) & ($\times 10^{-2}$)       & ($\times 10^{-3}$) & ($\times 10^{-4}$) & ($\times 10^{-2}$) \\
                \midrule
                MLE (general, train freq prior)       & -65.15                  & [-39.82,39.84]     & [-5.10, 152.04]          & 48.20              & 412.61             & 544.13             \\
                MLE (general, uniform prior)          & -20.20                  & [-33.97,33.98]     & [38.65, 105.00]          & 39.46              & 300.29             & 253.97             \\
                \textit{UXM}                          & 98.69                   & [-3.54, 3.54]      & [-9.85, 8.46]            & 3.74               & 3.27               & 7.86               \\
                \bottomrule
            \end{tabular}%
        }}
\end{table}

One possible reason why MLE deconvolvers fail is that they are applied to
neural network outputs, which are known not to be calibrated and therefore not
representative of the true posterior probabilities~\cite{guo2017calibration}.
Another issue is the high sensitivity of the deconvolvers to the prior over
classes. Indeed, we see in \Cref{tab:supp_mle_general_deconvoler_results} that
using a uniform prior, even if not mathematically correct, leads to better
results. It also appears that each combination of type of output on which the
model was trained and prior elicits a specific bias cell type, for which the
predicted proportion is always high. For instance, in the case of MLE
MethylBERT-style, when applied to the outputs of \textit{MethylBERT} trained on
hard labels with focal loss and using the train frequency prior, the average
predicted proportion of \emph{Gallbladder} across all pure mixtures is 24.0\%.
A perfect deconvolver would have given an average prediction across all pure
mixtures of $1/39 \approx 2.6\%$ for all cell types.
\Cref{tab:supp_mle_deconvolver_bias_ctypes} summarizes the biases of all MLE
settings presented in this section, as well as \textit{UXM} bias.

\begin{table}[htbp]
    \centering
    \caption{Biases of the different MLE methods.}
    \label{tab:supp_mle_deconvolver_bias_ctypes}
    {\footnotesize
        \resizebox{\textwidth}{!}{%
            \begin{tabular}{c cccc}
                \toprule
                \textbf{MLE flavour} & \textbf{Training of MethylBERT}    & \textbf{Prior}            & \textbf{Bias cell type} & \textbf{Avg prediction of bias cell type over pure mixtures} \\
                \midrule
                MethylBERT-style     & Hard labels w/ Focal loss          & Train frequency           & \emph{Gallbladder}      & 24.0\%                                                       \\
                MethylBERT-style     & Hard labels w/ Focal loss          & Uniform                   & \emph{Blood-T    }      & 24.0\%                                                       \\
                MethylBERT-style     & Hard labels w/ Minibatch balancing & Effective Train frequency & \emph{Neuron     }      & 15.0\%                                                       \\
                MethylBERT-style     & Hard labels w/ Minibatch balancing & Uniform                   & \emph{Neuron     }      & 22.7\%                                                       \\
                MethylBERT-style     & Binary labels w/ Focal loss        & Train frequency           & \emph{Dermal-Fibro}     & 3.3\%                                                        \\
                MethylBERT-style     & Binary labels w/ Focal loss        & Uniform                   & \emph{Endothel     }    & 7.1\%                                                        \\
                General              & Soft Labels w/ Pooling             & Train frequency           & \emph{Dermal-Fibro}     & 76.0\%                                                       \\
                General              & Soft Labels w/ Pooling             & Uniform                   & \emph{Bladder-Ep   }    & 74.3\%                                                       \\
                \midrule
                ---                  & \textit{UXM}                       & ---                       & \emph{Fallopian-Ep}     & 3.0\%                                                        \\
                \bottomrule
            \end{tabular}%
        }}
\end{table}

\subsection{Fitting parameters of deconvolvers}
\label{sec:supp_deconv_params}

All deconvolution models were trained on the same feature-selected input
matrices derived from in-silico pseudobulk mixtures
(cf.~\Cref{supp:dec_features_selection}). Below we list the hyperparameters
used for each deconvolver.

\paragraph{XGBoost (\textit{XGB}).}
A gradient-boosted tree ensemble was trained with 500 estimators, a maximum depth of 15 and a learning rate of 0.05.
Row subsampling was set to 0.7, with full-column sampling per tree.
Regularization was controlled via a minimum child weight of 5, L1 penalty $\alpha = 0.1$ and L2 penalty $\lambda = 1.0$.
Outputs were clipped and normalized to produce valid mixture proportions.

\paragraph{Single-layer Wide Network (\textit{SWN}).}
A single hidden layer of 1024 units with GELU activation and 20\% dropout, followed by a softmax output layer.
Training ran for up to 5000 epochs with a batch size of 512, a learning rate of $5 \times 10^{-5}$, weight decay of $10^{-3}$ and early stopping with a patience of 1000 epochs based on validation MAE.
No learning rate scheduler was used.

\paragraph{Multi-Layer Perceptron (\textit{MLP}).}
Two hidden layers of 512 and 256 units (each with GELU activation and 20\% dropout), followed by a projection back to the input dimensionality with 10\% dropout and a softmax output layer.
Training used the same schedule as \textit{SWN}: up to 5000 epochs, batch size 512, learning rate $5 \times 10^{-5}$, weight decay $10^{-3}$ and early stopping with patience 1000 on validation MAE.
No learning rate scheduler was used.

\paragraph{Non-Negative Least Squares (\textit{NNLS}) and Probability Simplex Least Squares (\textit{PSLS}).}
Both solvers used a reference matrix constructed from the training-split pure profiles after applying the feature mask.
Both \textit{NNLS} and \textit{PSLS} were solved using cvxpy~\cite{diamond2016cvxpy}.
Neither method has learnable hyperparameters beyond the reference matrix itself.

\subsection{Calibration methods}
\label{sect:supp_mat_calibration}

\paragraph{Calibration methods analysis.}
In addition to the simple linear calibration scheme that we introduced, we also
experimented with the existing calibration schemes for multiclass
classification, as described in \Cref{sect:methods_calibration}.

The parameters $W$ and $\mathbf{b}$ from
\Cref{eq:multiclass_calibration_standard_form} are optimized via Adam with
cross-entropy loss between the true and calibrated proportions.

The parameters are regularized with an $\ell_2$ penalty of strength $\lambda$
in the case of temperature scaling and vector scaling. For Dirichlet
calibration, when $\mu$ is None, we apply $\ell_2$ regularization on all
parameters with strength $\lambda$. When $\mu$ is not None, we apply the ODIR
regularization of~\cite{kull2019beyond} with strength $\lambda$ on the
off-diagonal parameters of $W$ and with strength $\mu$ on $\mathbf{b}$.

Training hyperparameters (learning rate and regularization strengths) are
selected by 3-fold cross-validation grid search over the mixture proportions of
\textit{UXM} on the validation set, retaining the configuration with the lowest
mean validation loss. During training, we apply a reduce-on-plateau learning
rate scheduler (factor 0.5, patience 10) and early stopping (patience 15,
tolerance $10^{-4}$), both monitored on the validation loss. We show the
parameters of the grid search and the selected hyperparameters in
\Cref{tab:supp_calibration_hyperparams}.

The results of the calibrators trained on the whole validation set with the
best hyperparameters are shown in \Cref{tab:supp_calibrator_comparison}.

Results show that our method and vector scaling achieve the best results in
terms of $R^2$, LoA, MAE and MSE\@. We use them in our main experiments.

\begin{table}[htbp]
    \centering
    \caption{Hyperparameter selection of the learned calibration functions on the validation set. The selected hyperparameters are underlined.}
    \label{tab:supp_calibration_hyperparams}
    {\footnotesize
        \begin{tabular}{cccc}
            \toprule
            \textbf{Model}                                          & \textbf{Reg.\ $\lambda$}                            & \textbf{Reg.\ $\mu$}         & \textbf{LR}                            \\
            \midrule
            Temperature scaling~\cite{guo2017calibration}           & [\underline{0}, 0.0001]                             & N/A                          & [0.0001, \underline{0.001}, 0.01, 0.1] \\
            Vector scaling~\cite{guo2017calibration,kull2019beyond} & [0, \underline{0.001}, 0.01, 0.1]                   & N/A                          & [0.0001, \underline{0.001}, 0.01]      \\
            Dirichlet calibration~\cite{kull2019beyond}             & [0, 100, 1000, \underline{2000}, 4000, 5000, 10000] & [\underline{None}, 0, 1, 10] & [1e-4, \underline{1e-5}, 1e-6, 1e-7]   \\
            \bottomrule
        \end{tabular}
    }
\end{table}

\begin{table}[htbp]
    \centering
    \caption{Calibrators performance on \textit{UXM} predictions on the test set. Best result is \textbf{bold}, second best result is \underline{underlined}.}
    \label{tab:supp_calibrator_comparison}
    {\footnotesize
        \resizebox{\textwidth}{!}{%
            \begin{tabular}{c cccccc}
                \toprule
                \multirow{2}{*}{\textbf{Calibrator}}                    & $\mathbf{R}^\mathbf{2}$ & \textbf{LoA}              & \textbf{LoA worst-class}    & \textbf{MAE}       & \textbf{MSE}       & \textbf{KL}        \\
                                                                        & ($\times 10^{-2}$)      & ($\times 10^{-2}$)        & ($\times 10^{-2}$)          & ($\times 10^{-3}$) & ($\times 10^{-4}$) & ($\times 10^{-2}$) \\
                \midrule
                No calibration                                          & 94.86                   & [-4.05, 4.05]             & [-10.82, 14.02]             & 6.49               & 4.27               & 14.98              \\
                Lin.\ Cal.\ clip-norm (Ours)                            & 95.91                   & [-3.61, 3.61]             & [-10.89, 13.04]             & 5.33               & 3.40               & 12.33              \\
                Lin.\ Cal.\ simplex-projection (Ours)                   & \underline{96.74}       & \underline{[-3.23, 3.23]} & [-10.78, 12.98]             & \textbf{4.32}      & \underline{2.71}   & 10.91              \\
                Temperature scaling~\cite{guo2017calibration}           & 96.21                   & [-3.48, 3.48]             & [-10.03, 12.83]             & 5.33               & 3.14               & \underline{10.24}  \\
                Vector scaling~\cite{guo2017calibration,kull2019beyond} & \textbf{96.90}          & \textbf{[-3.14, 3.14]}    & \underline{[-10.44, 11.23]} & \underline{4.99}   & \textbf{2.57}      & \textbf{9.74}      \\
                Dirichlet calibration~\cite{kull2019beyond}             & 95.77                   & [-3.67, 3.67]             & \textbf{[-9.56, 9.61]}      & 6.09               & 3.51               & 13.11              \\
                \bottomrule
            \end{tabular}%
        }}
\end{table}

\paragraph{Calibrators fitting.}\label{sect:supp_par_calibrators_fitting}
In the main experiments, we fit vector scaling on the validation split and
select $\lambda$ and the learning rate for each classifier-labeling-deconvolver
combination by 3-fold cross-validation grid search across sets
$\{10^{-4},10^{-3}, 10^{-2}\}$ for learning rate and $\{0, 10^{-4}, 10^{-3},
    10^{-2}\}$ for $\lambda$. Our method is hyperparameter-free and thus does not
require a selection procedure.

\begin{figure}[htbp]
    \centering
    \includegraphics[width=1\textwidth]{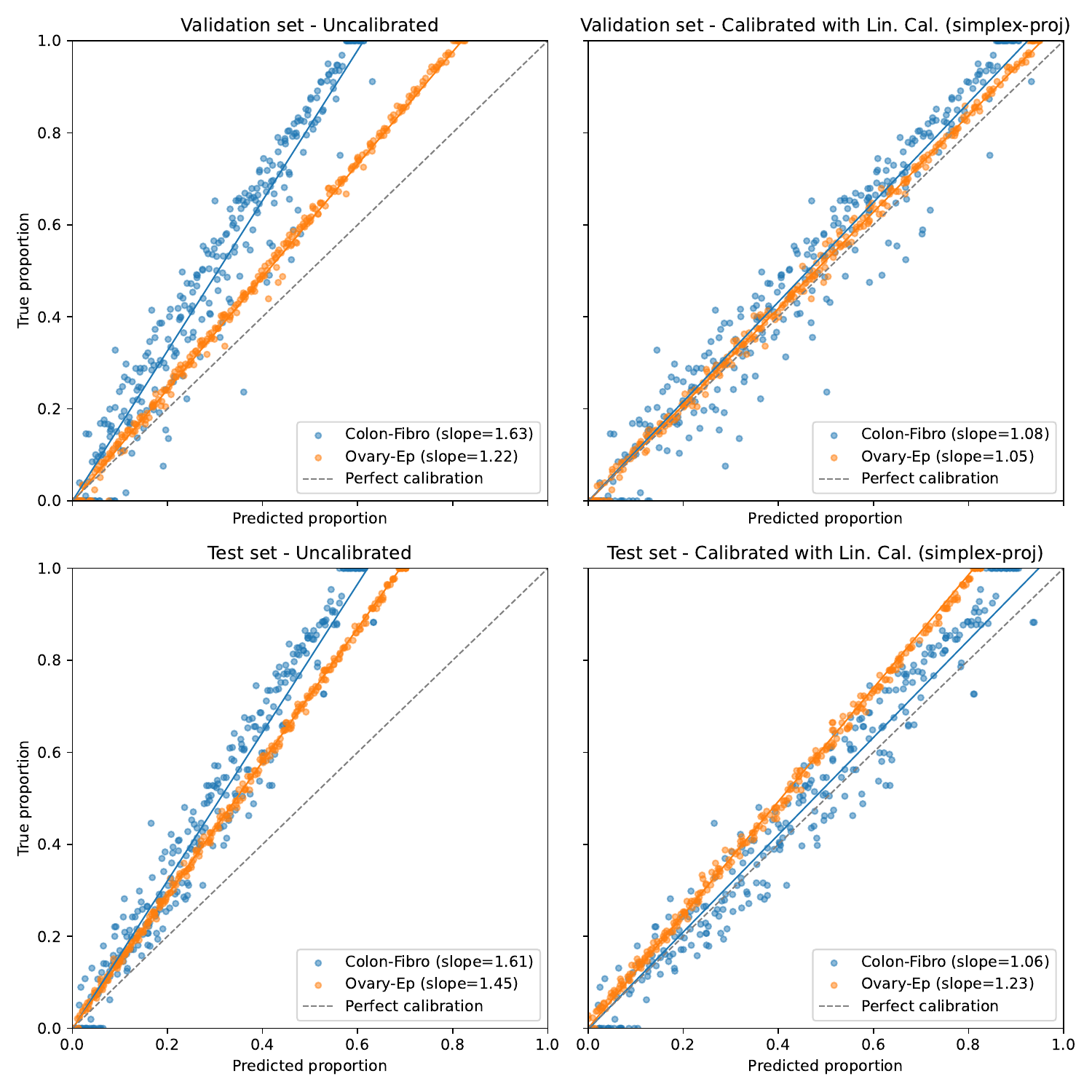}
    \caption{Uncalibrated and calibrated proportions (using linear calibrator with simplex projection) predicted by Dismir + Soft Labels w/ pooling + PSLS on the validation and test set for the two cell types with worst calibration.The regression lines and slopes are computed on the whole sets, but for visualization purposes, a subset of 300 pseudobulks have been selected for each cell type in such a way that the scattered points span uniformly the $[0,1]$ range.}
    \label{fig:supp_linear_cal_illustration}
\end{figure}

\subsection{Baselines}

\label{sect:supp_baselines_description}

\paragraph{Motivation for selection.} \textit{UXM} is the method proposed by~\citet{loyfer2023dna} alongside their dataset that we rely on for training.
It is also one of the best methods for WGBS data ~\cite{giuili2025benchmark}. \textit{CelFiE}~\cite{caggiano2021comprehensive}. is reported as the best method in the DecoNFlow
benchmark~\cite{giuili2025benchmark} and is expected to perform well on the chosen atlas~\cite{sun2024systematic}. The \textit{Houseman CP}~\cite{houseman2012dna} is one of the earliest methylation-based deconvolution algorithms, the best
beta-value-based method in the DecoNFlow benchmark ~\cite{giuili2025benchmark}, strong in several other benchmarks ~\cite{jeong2022systematic,song2022systematic}, and is one of the most frequently used for comparison.
And \textit{EpiDISH}~\cite{teschendorff2017epidish} is reported as the best method in multiple benchmarks~\cite{deridder2024benchmarking,song2022systematic}.

\paragraph{UXM~\cite{loyfer2023dna}.}

\emph{UXM} is an unsupervised reference-based deconvolution method operating on the aggregated counts of labeled DNA fragments.
\emph{UXM} requires a reference atlas matrix $A \in \mathbb{R}^{G\times C}$ where $A_{g,c}$ is equal to the ratio of reads in GR group $g$ that are unmethylated (\textit{i.e.} with at most $25\%$ methylated CpGs).
At inference, the ratios $b_g$ of unmethylated reads in GR group $g$ are aggregated in a vector $\mathbf{b} \in \mathbb{R}^G$ and the mixture proportions $\mathbf{x} \in \Delta^{C-1}$ are inferred using \textit{NNLS}, minimizing $||A\mathbf{x} - b||_2^2$. $\mathbf{x}$ is normalized to $1$ a posteriori.
In our work, we use four different atlases, built on the region sets described in \Cref{sect:supp_region_sets}.

1. In the pseudobulk experiment, we first selected genomic region groups from the \textit{U25} atlas for the \textit{hg38} reference, published by \citet{loyfer2023dna}. We then computed $A_{g,c}$ using the train split, utilizing a $25\%$ methylation value to label fragments as unmethylated, consistent with the UXM default threshold.
Limiting estimation of the unmethylated reads ratio to the train split is motivated by avoiding information leak that would otherwise be exerted from validation and test sets.

2. In the OOD RRBS experiment, we used \textit{U25} atlas for \textit{hg19} reference, published by \citet{loyfer2023dna} as our main atlas for inference, with a set of GR groups and corresponding $A_{g,c}$ as in the original work.
Additionally, we used \textit{U250} atlas for \textit{hg19} reference, published by \citet{loyfer2023dna} to compute \emph{UXM U250} TCS\@. We did it to generate a comparison point where \emph{UXM}
has access to $\approx10 \times$ more GR groups and reads at inference and to see if Syto could match this result running on \textit{U25}. Finally, the \textit{U25 GSS sorted} atlas
was used to test if better results can be obtained for \emph{UXM} by selecting different GR groups. We reported the score achieved with this atlas under \textit{UXM U25 GSS sorted} (see \Cref{fig:uxm_rrbs_results_all}). We also used the \textit{U25 GSS sorted} atlas to run an auxiliary GR group selection sensitivity experiment for \textit{Syto} (see \Cref{sect:gr_sensitivity_analysis}).

For running \emph{UXM}, we ported the original Python code (see
\Cref{sect:licensing}). Since our pipeline requires converting .pat files into
reads, we additionally validated that \emph{UXM} CLI run on .pat and our port
run on recovered reads produce the same results by comparing the distribution
of TCS\@. The mean absolute difference across 521 evaluated OOD samples was
0.000369, or $6.6\%$ of the \emph{UXM} uncalibrated pseudobulk MAE which
suggests that our port produced results sufficiently close to the original CLI
implementation.

\paragraph{CelFiE~\cite{caggiano2021comprehensive}.}
\emph{CelFiE} is an Expectation Maximization (EM) based algorithm, operating on the individual CpGs or GR groups aggregated $\beta$-values.
Each row in the reference features the number of methylated CpGs and the total number of CpGs for each cell type $c \in C$. At inference time, \emph{CelFiE} jointly optimizes reference atlas $\beta$-values (parameterized as $\gamma$) and derived cell-type proportions $\alpha$, conditioned on
initial atlas and $s \in S$ bulk mixtures to deconvolute. It can also work with an arbitrary number of unknown components $n_{unknown}$, by extending the atlas with $n_{unknown} \times 2$ columns, and the number of methylated CpGs and total number of CpGs for each unknown cell type initialized with zeroes.
We used a set of GR groups from the \textit{U25} atlas for the \textit{hg38} reference, published by \citet{loyfer2023dna}, and a training set to compute the atlas for the pseudobulk experiment.
For the OOD RRBS experiment, we used GR groups from the \textit{U25} atlas for the \textit{hg19} reference, published by \citet{loyfer2023dna}, and recovered reads overlapping GR groups from all three pseudobulk splits to compute the atlas.
For running \emph{CelFiE}, we first ported the original \emph{CelFiE} Python code (see \Cref{sect:licensing}). We kept inference parameters at default values: $random\_restarts=10; convergence\_criteria=0.0001$. We increased the number of EM iterations from the default $1000$ to $10000$ to guarantee convergence.
We tried running \emph{CelFiE} using individual CpGs and GR groups aggregated $\beta$-values. Based on a pseudobulk experiment, using aggregated signal yielded significantly stronger results, so we set inference to GR groups, validating a previous finding stated by \citet{caggiano2021comprehensive}.
Despite the aforementioned settings, our initial experiments positioned \emph{CelFiE} inferior to \emph{UXM}, in contrast to~\cite{giuili2025benchmark}. After careful analysis, we have found that optimization of the reference at inference time significantly degrades results. Specifically, a variant of
\emph{CelFiE} with enabled optimization of reference atlas yielded MSE of $5.11 \times 10^-4$, while dissabling updates of $\beta$-values reduced MSE to $3.28 \times 10^-4$.
Therefore, we freeze $\gamma$ at its initial values across experiments to strengthen the baseline. We argue that re-estimating the atlas is necessary in modeling with unknown components and feasible when several independent samples per deconvolution task are available (the last condition is violated in our pseudobulks).
In all our settings, we deconvolute one sample at a time.

\paragraph{EpiDISH~\cite{teschendorff2017epidish}.}

\emph{EpiDISH} is a singleton method and an R package implementing four different reference-based algorithms, each operating on the individual CpGs $\beta$-values.
In this paper, we consider two algorithms: (1) robust partial correlations (RPC) with a posteriori elimination of negative proportions followed by scaling such that the sum of proportions equates to one.
(2) linear constraint projection, which enforces proportion positivity and, optionally, normalization to ensure the sum of prportions to one at inference. Algorithm (1) is \emph{EpiDISH} proper, and we refer to it as is throughout our work.
Algorithm (2) is an implementation of the Houseman~\cite{houseman2012dna} method, and we refer to it as \emph{Houseman CP} or \emph{Houseman Ineq.} interchangeably. Relaxing the inequality constraint produces better results in accordance with~\cite{giuili2025benchmark}, so we run it accordingly.
To integrate these methods in our experimental pipelines, we translated the original R code (see \Cref{sect:licensing}) into Python. We then generated 5 synthetic samples with 4 cell types each as test fixtures and computed their derived proportions using an R package. We did the same computation.
for our Python port and compared the two. Obtained proportions differed at most by 0.000233 and 0.000327 for \emph{EpiDISH} and \emph{Houseman CP}, which we deemed acceptable.
For each experiment, we ran \emph{EpiDISH} and \emph{Houseman CP} methods using atlases as derived for \emph{CelFiE} with $\beta$-values computed on the individual CpGs.

\newpage
\section{Data preparation for the experiments}
\label{sect:supp_data_preparation}

\subsection{Data preprocessing of .pat files}
\label{sect:supp_data_preprocessing}

The raw data featured in our experiments is provided in ultra-compact
\textit{*.pat} format where reads with the same covered CpG pattern are
collapsed in single data record without aligned sequences
\cite{loyfer2026wgbstools}. Since the classification models expect a
combination of read-level DNA sequence and methylation pattern, we recovered
this information using our modified version of the tool by
\citet{jeong2025methylbert}. The original version of the tool for recovering
mapped reads from raw .pat files kept only the reads lying strictly within the
selected genomic regions, discarding reads overlapping region boundaries. In
the earlier versions of this paper, we did a whole range of experiments relying
on the data filtered this way. Subsequent checks demonstrated that such
stringent read inclusion criteria lead to detrimental results across
experiments, equally for Syto and baselines. We therefore modified the read
recovery tool to keep all reads overlapping target regions, with start and end
trimmed to region boundaries. This led to a uniform improvement of MSE in the
pseudobulk experiment by 33\% for \textit{Syto} and \textit{UXM}, the only
baseline we had at a time we introduced this change.

The genomic regions against which the reads were recovered, and the reference
assemblies involved, are described in \Cref{sect:supp_region_sets}.

\subsection{Marker region sets used in the experiments}
\label{sect:supp_region_sets}

As regards the cell-type atlas~\cite{loyfer2023dna}, we downloaded all data in
the aligned format using both reference assemblies, hg19 and hg38. In the case
of the tissue atlas~\cite{li2023comprehensive}, only the hg19 aligned format
was available. For each alignment, the reads were recovered from the genomic
positions overlapping with the corresponding \textit{"top-250 unmethylated
    markers atlases"} published by \citet{loyfer2023dna} (namely
\textit{Atlas.U250.l4.hg19.full.tsv} and
\textit{Atlas.U250.l4.hg38.full.tsv}\footnote{Downloaded from
    \href{https://github.com/nloyfer/uxm\_deconv/tree/main/supplemental}{https://github.com/nloyfer/uxm\_deconv/tree/main/supplemental}}).
Recovering reads against the \textit{U250} region sets makes a single recovery
pass sufficient for all our experiments: every other region set we use is
nested within \textit{U250} for the same assembly, so the corresponding reads
are obtained by subsetting. \Cref{tab:supp_region_sets} summarizes the four
resulting region sets. Within a given experiment, the same set of GR groups was
used for all baseline methods and all Syto permutations. The way each method
turns those regions into its own reference is method-specific
(cf.~\Cref{sect:supp_baselines_description}).

\begin{table}[H]
    \centering
    \caption{Marker region sets used in our experiments. All of them are derived from the unmethylated marker atlases of \citet{loyfer2023dna}.}
    \label{tab:supp_region_sets}
    {\footnotesize
        \begin{tabular}{llll}
            \toprule
            \textbf{Name}           & \textbf{Assembly} & \textbf{DMRs per GR group} & \textbf{Used in}                               \\
            \midrule
            \textit{U25}            & hg38              & up to 25                   & pseudobulk experiment, all methods             \\
            \textit{U25}            & hg19              & up to 25                   & OOD RRBS experiment, all methods               \\
            \textit{U250}           & hg19              & up to 250                  & OOD RRBS experiment, \emph{UXM U250} only      \\
            \textit{U25 GSS sorted} & hg19              & up to 25                   & OOD RRBS experiment, \emph{UXM U25 GSS sorted} \\
                                    &                   &                            & and Syto region selection sensitivity          \\
            \bottomrule
        \end{tabular}%
    }
\end{table}

We generated \textit{U25 GSS sorted} by subsetting up to 25 top GR groups from
the \emph{U250} atlas for \textit{hg19} reference using proxy gap specificity
score (see \Cref{sect:supp_atlas_analysis}).

We initially trained \textit{MethylBERT} on \textit{hg19}-aligned and
\textit{hg38}-aligned data and observed no significant differences in the
attained losses. Based on this observation and the fact that hg38 had more
overlaping reads, we disregarded hg19, and our training effort was done on
hg38. For inference in our main experiments, we used
\textit{Atlas.U25.l4.hg38.full.tsv} in the pseudobulk experiment and
\textit{Atlas.U25.l4.hg19.full.tsv} in the OOD experiment; mixing assemblies
was possible for \textit{Dismir} and \textit{MethylBERT}, because we united
DMRs into cell-type groups, and for those classifiers training and infering on
different alignments was possible. \textit{Lookup Classifier} is not alignment
agnostic, because constructed lookup keys are position specific, hence, for
this classifier only we used hg38 aligned reads for training in pseudobulk
experiment and hg19 aligned reads for training in ODD experiment.

\subsection{Analysis of the DNA methylation atlas of normal human cell types} \label{sect:supp_atlas_analysis}

\begin{figure}[h]
  \centering
  \includegraphics[width=1\textwidth]{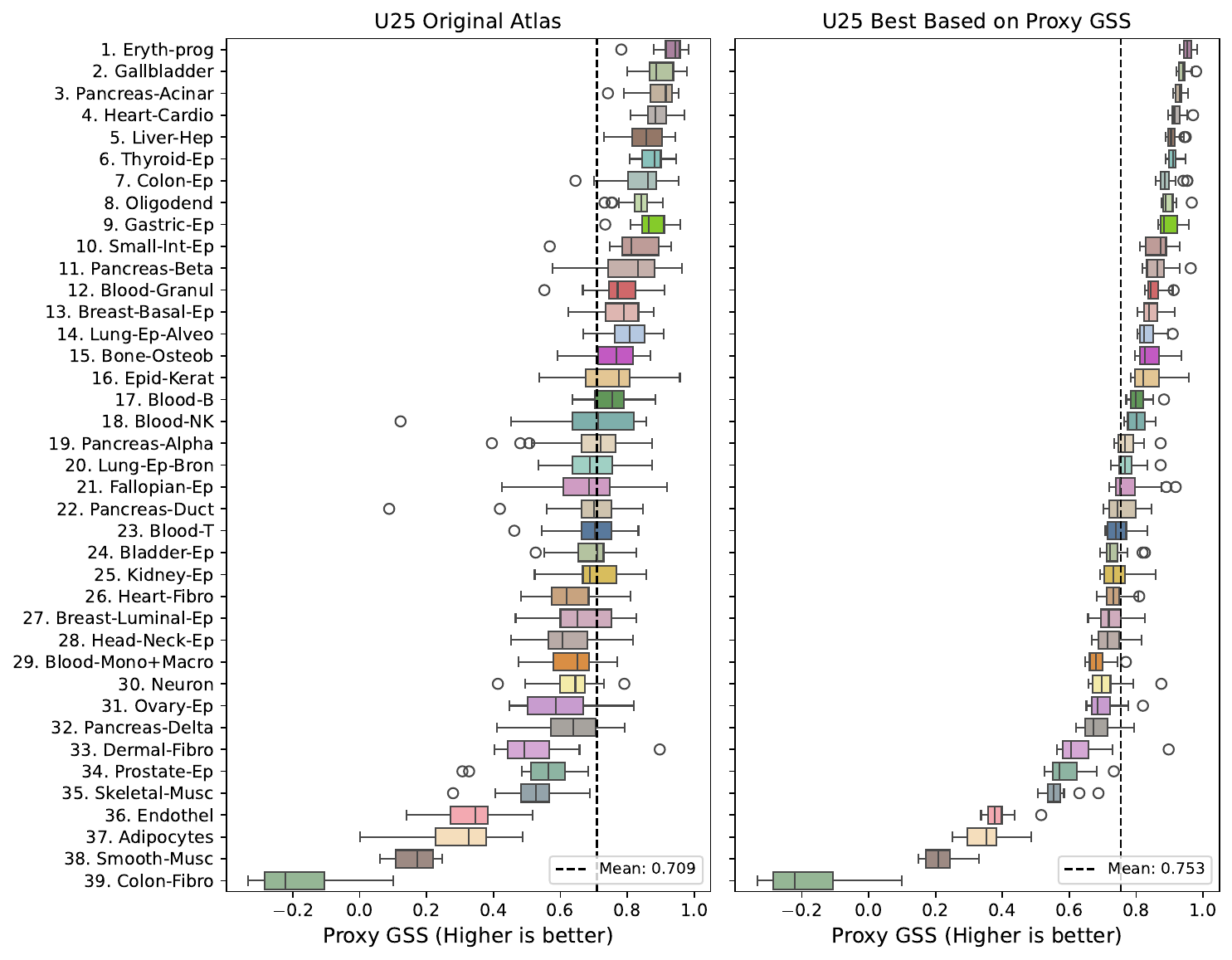}
  \caption{Proxy Gap Specificity Scores or difference between proportion of on-target unmethylated reads and maximal proportion of unmethylated reads from any other cell type in the same region (best appearing on top).
    We selected U25 Best Based on Proxy GSS (right graph) by subsetting the Top-25 regions per cell type from the U250 atlas, ordered by Proxy GSS.
  }
  \label{fig:GSS-atlas}
\end{figure}

To better understand the structure of the atlas, we calculated a \textit{proxy
  gap specificity score} for each cell-type-specific marker region as the
difference between the proportion of unmethylated reads for the target cell
type (the cell type for which the region is specific) and the maximal fraction
of unmethylated reads stemming from other cell-types in the same region (taking
inspiration from the work by \citet{guo2026guidelines}). The analysis revealed
a previously unreported\footnote{ \citet{sun2024systematic} featured the
  \textit{overall} specificity of different sets of markers, including those used
  in this study; but no in-depth analysis \textit{per cell-type group} was done}
heterogeneity in the specificity of the reference regions, with a significant
tail of lower-quality groups (\Cref{fig:GSS-atlas}). This proven to be
important later when assesed in supplemntary experiment (see
\Cref{fig:uxm_rrbs_results_all}) and contributed to understanding of the
relevance of regions selection to the overall performance.

\newpage

\subsection{Mapping of cfSort tissues to Tabula Sapiens atlas}
\label{sect:supp_mapping_cfsort_tissues_to_tabula_sapiens}

\begin{figure}[H]
  \centering
  \includegraphics[width=1\textwidth]{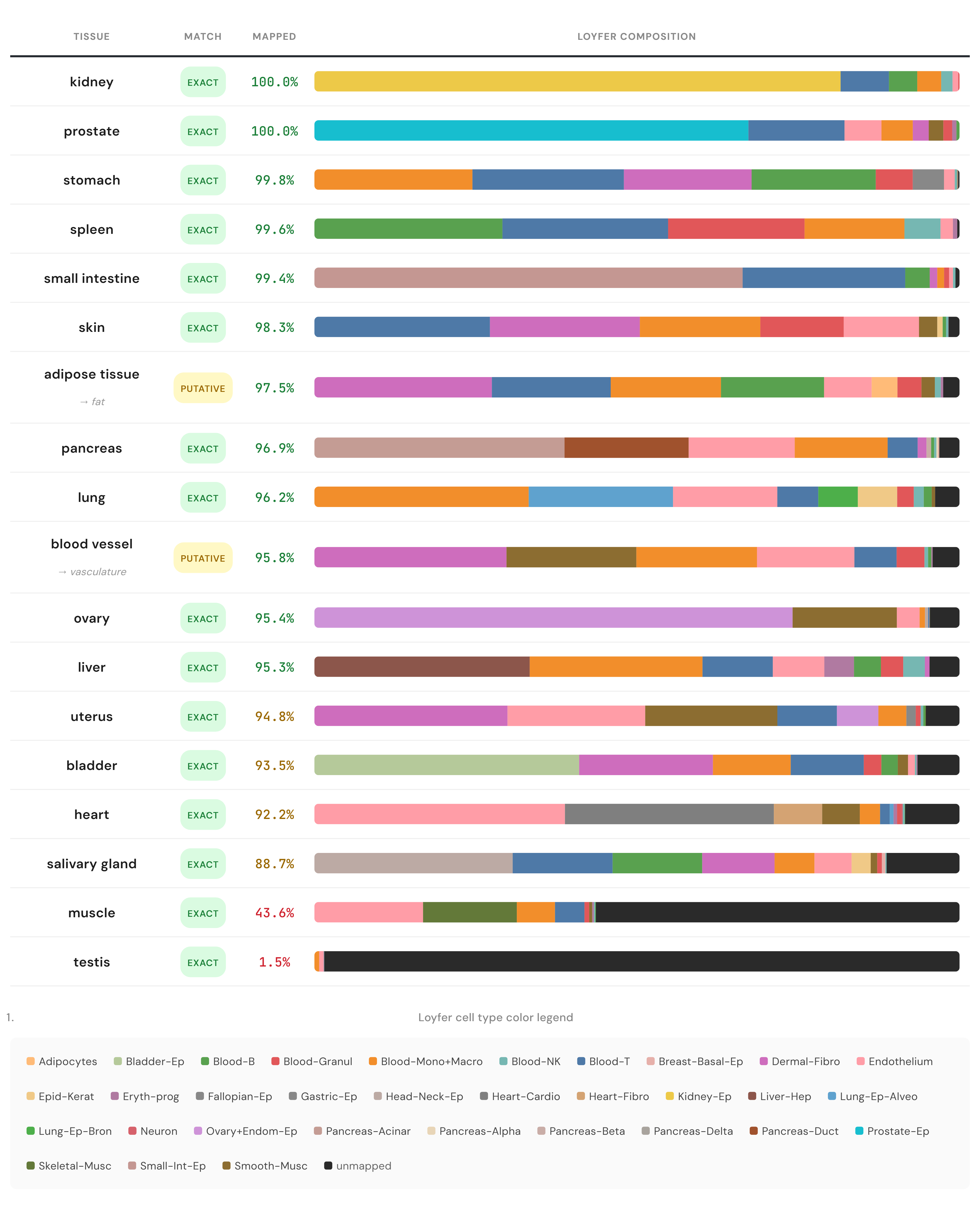}
  \caption{16+2 mapped cfSort tissues and their derived cell type contributions in terms of \citet{loyfer2023dna} atlas. 11 tissues had no corresponding mapping and are not displayed here.}
  \label{fig:cfsort-tabula-map}
\end{figure}

To perform mapping, we first matched the tissue types from the OOD dataset to
those in the single-cell transcriptomic atlas of human tissues \emph{Tabula
  Sapiens}~\cite{quake2025tabula, the2022tabula} using UBERON codes from OLS4
\cite{mclaughlin2025ols4}. The intersection yielded 16 exact and 2 putative
matches corresponding to 289 samples. We then mapped the cell types from the
\emph{Tabula Sapiens} project to the corresponding cell types from our
reference atlas~\cite{loyfer2023dna} using CL codes from OLS4
\cite{mclaughlin2025ols4} (cf.~\Cref{tab:loyfer_subtypes}). This additional
mapping allowed us to estimate the percentage of cells with direct
correspondence between the transcriptomic and methylation atlases for each
mapped tissue. Two tissues (muscle and testis) exhibited low mapping rates
($43.6\%$ and $1.5\%$) and were excluded from the analysis, resulting in a
final set of 14 exact and 2 putative matches with matching rates in the range
$88.7\% - 100\%$, and 260 samples in total. \Cref{fig:cfsort-tabula-map} shows
the inferred cell type distribution per tissue and \Cref{tab:loyfer_subtypes}
shows cell types mapping. The mapping was performed only for those \emph{Tabula
  Sapiens} cell types that contributed to the tissues overlaping with OOD
dataset.

\newpage

\newpage
\setlength{\LTcapwidth}{\linewidth}
\begin{longtable}[H]{p{2.6cm} p{2.6cm} p{2cm} >{\raggedright\arraybackslash}p{6cm}}
  \caption[Loyfer cell-type subtypes with their curated Cell Ontology term]{%
    Loyfer cell-type subtypes with their curated Cell Ontology (CL) term and the Tabula Sapiens labels that strictly-match each subtype (\textbf{bold} and (e) = exact match with CL term; (e) = exact, (m) = manual match, (em) = both).
    Exact matches are rare because cell types in Loyfer$'$s atlas are in fact cell-type groups, while Tabula Sapiens operates at a more granular level.
    There are exceptions: for example, the Tabula Sapiens \emph{fibroblast} matches the more specific Loyfer Colon-Fibro and Dermal-Fibro subtypes.
    In such cases, the choice was tissue-specific; for our setup, we used Dermal-Fibro across matched tissues, as it is more generic than Colon-Fibro (which is colon-specific and absent from our OOD set).
    Several matches were established by considering Tabula Sapiens cell types in the context of their originating tissue: \emph{ciliated epithelial cell} (uterus in Tabula Sapiens) was attributed to Fallopian-Ep, and \emph{epithelial cell} (stomach in Tabula Sapiens) to Gastric-Ep.
  } \label{tab:loyfer_subtypes}                                                                                                                \\
  \toprule
  \textbf{Loyfer cell type} & \textbf{Source tissues} & \textbf{CL terms} & \textbf{Tabula Sapiens strict matches}                             \\
  \midrule
  \endfirsthead
  \multicolumn{4}{l}{\emph{(continued)}}                                                                                                       \\
  \toprule
  Source Tissue             & Cell type               & CL term           & Tabula Sapiens strict matches                                      \\
  \midrule
  \endhead
  \midrule
  \multicolumn{4}{r}{\emph{(continued on next page)}}                                                                                          \\
  \endfoot
  \bottomrule
  \endlastfoot
  Adipocytes                & [Abdominal Subcut.]     & [CL:0000136]      & [mesenchymal stem cell of adipose tissue (m)]                      \\
  \hline
  Bladder-Ep                & [Bladder]               & [CL:1001428]      & [\textbf{bladder urothelial cell (em)}]                            \\
  \hline
  Blood-B                   & [Blood]                 & [CL:0000236,      & [\textbf{B cell (em)},                                             \\
                            &                         & CL:0000787]       & plasma cell (m)]                                                   \\
  \hline
  Blood-Granul              & [Blood]                 & [CL:0000094]      & [\textbf{granulocyte (em)},                                        \\
                            &                         &                   & basophil (m),                                                      \\
                            &                         &                   & mast cell (m),                                                     \\
                            &                         &                   & neutrophil (m)]                                                    \\
  \hline
  Blood-Mono+Macro          & [Blood,                 & [CL:0000091,      & [\textbf{monocyte (em)},                                           \\
                            & Colon,                  & CL:0000576,       & classical monocyte (m),                                            \\
                            & Liver,                  & CL:0000583,       & intermediate monocyte (m),                                         \\
                            & Lung alveolar,          & CL:0009038,       & macrophage (m),                                                    \\
                            & Lung interstitial]      & CL:1001603]       & mononuclear phagocyte (m),                                         \\
                            &                         &                   & non-classical monocyte (m),                                        \\
                            &                         &                   & tissue-resident macrophage (m)]                                    \\
  \hline
  Blood-NK                  & [Blood]                 & [CL:0000623]      & [\textbf{natural killer cell (em)}]                                \\
  \hline
  Blood-T                   & [Blood]                 & [CL:0000084,      & [CD4-positive alpha-beta T cell (m),                               \\
                            &                         & CL:0000492,       & \textbf{CD8-positive alpha-beta T cell (em)},                      \\
                            &                         & CL:0000625,       & \textbf{T cell (em)},                                              \\
                            &                         & CL:0000895,       & \textbf{naive thymus-derived CD4-positive alpha-beta T cell (em)}, \\
                            &                         & CL:0000900,       & \textbf{naive thymus-derived CD8-positive alpha-beta T cell (em)}, \\
                            &                         & CL:0000910,       & activated CD4-positive alpha-beta T cell (m),                      \\
                            &                         & CL:0000913,       & activated CD8-positive alpha-beta T cell (m),                      \\
                            &                         & CL:0001087,       & gamma-delta T cell (m),                                            \\
                            &                         & CL:0002678]       & mature NK T cell (m),                                              \\
                            &                         &                   & regulatory T cell (m)]                                             \\
  \hline
  Bone-Osteob               & [Bone]                  & [CL:0000062]      & ---                                                                \\
  \hline
  Breast-Basal-Ep           & [Breast]                & [CL:0002324]      & [basal cell (m),                                                   \\
                            &                         &                   & myoepithelial cell (m)]                                            \\
  \hline
  Breast-Luminal-Ep         & [Breast]                & [CL:0002326]      & ---                                                                \\
  \hline
  Colon-Ep                  & [Colon]                 & [CL:0000164,      & [\textbf{enteroendocrine cell (em)}]                               \\
                            &                         & CL:0011108]       &                                                                    \\
  \hline
  Colon-Fibro               & [Colon]                 & [CL:4052007]      & [fibroblast (m)]                                                   \\
  \hline
  Dermal-Fibro              & [Ad-dermal]             & [CL:0002551]      & [fibroblast (m)]                                                   \\
  \hline
  Endothelium               & [Aorta,                 & [CL:0000115,      & [\textbf{endothelial cell (em)},                                   \\
                            & Kidney glomerular,      & CL:0002543,       & \textbf{endothelial cell of artery (em)},                          \\
                            & Kidney tubular,         & CL:1000398,       & \textbf{vein endothelial cell (em)},                               \\
                            & Liver,                  & CL:1000413,       & capillary endothelial cell (m),                                    \\
                            & Lung alveolar,          & CL:1000892,       & cardiac endothelial cell (m),                                      \\
                            & Pancreas,               & CL:1001005,       & endothelial cell of arteriole (m),                                 \\
                            & Vascular saphenous]     & CL:4028001]       & endothelial cell of lymphatic vessel (m),                          \\
                            &                         &                   & endothelial cell of vascular tree (m),                             \\
                            &                         &                   & endothelial cell of venule (m)]                                    \\
  \hline
  Epid-Kerat                & [Skin]                  & [CL:0000312]      & [basal cell (m),                                                   \\
                            &                         &                   & keratocyte (m),                                                    \\
                            &                         &                   & sebocyte (m)]                                                      \\
  \hline
  Eryth-prog                & [Bone marrow]           & [CL:0000038]      & [erythrocyte (m),                                                  \\
                            &                         &                   & hematopoietic precursor cell (m),                                  \\
                            &                         &                   & hematopoietic stem cell (m)]                                       \\
  \hline
  Fallopian-Ep              & [Fallopien tubes]       & [CL:4030007]      & [ciliated epithelial cell (m)]                                     \\
  \hline
  Gallbladder               & [Gallbladder]           & [CL:1000415]      & ---                                                                \\
  \hline
  Gastric-Ep                & [Gastric,               & [CL:0000164,      & [\textbf{enteroendocrine cell (e)},                                \\
                            & Gastric antrum,         & CL:0002178]       & epithelial cell (m)]                                               \\
                            & Gastric body,           &                   &                                                                    \\
                            & Gastric fundus]         &                   &                                                                    \\
  \hline
  Head-Neck-Ep              & [Esophagus,             & [CL:0002252,      & [acinar cell of salivary gland (m),                                \\
                            & Larynx,                 & CL:0002633,       & glandular secretory epithelial cell (m),                           \\
                            & Pharynx,                & CL:1001573,       & salivary gland cell (m)]                                           \\
                            & Tongue,                 & CL:1001576,       &                                                                    \\
                            & Tonsil palatine,        & CL:1001577]       &                                                                    \\
                            & Tonsil pharyngeal]      &                   &                                                                    \\
  \hline
  Heart-Cardio              & [Heart]                 & [CL:0000746]      & [regular atrial cardiac myocyte (m),                               \\
                            &                         &                   & ventricular cardiac muscle cell (m)]                               \\
  \hline
  Heart-Fibro               & [Heart]                 & [CL:0002548]      & [\textbf{fibroblast of cardiac tissue (em)}]                       \\
  \hline
  Kidney-Ep                 & [Kidney glomerular,     & [CL:0000653,      & [\textbf{kidney epithelial cell (em)}]                             \\
                            & Kidney tubular]         & CL:0002518,       &                                                                    \\
                            &                         & CL:1000507]       &                                                                    \\
  \hline
  Liver-Hep                 & [Liver]                 & [CL:0000182]      & [\textbf{hepatocyte (em)}]                                         \\
  \hline
  Lung-Ep-Alveo             & [Lung alveolar,         & [CL:0000077,      & [\textbf{mesothelial cell (e)},                                    \\
                            & Lung pleural]           & CL:0002062]       & \textbf{pulmonary alveolar type 1 cell (em)},                      \\
                            &                         &                   & pulmonary alveolar type 2 cell (m)]                                \\
  \hline
  Lung-Ep-Bron              & [Lung bronchus]         & [CL:0002328]      & [club cell (m),                                                    \\
                            &                         &                   & lung multiciliated epithelial cell (m),                            \\
                            &                         &                   & pulmonary ionocyte (m),                                            \\
                            &                         &                   & respiratory tract goblet cell (m),                                 \\
                            &                         &                   & serous cell of epithelium of bronchus (m)]                         \\
  \hline
  Neuron                    & [Brain]                 & [CL:0000540]      & [\textbf{neuron (em)}]                                             \\
  \hline
  Oligodend                 & [Brain]                 & [CL:0000128]      & ---                                                                \\
  \hline
  Ovary+Endom-Ep            & [Endometrium,           & [CL:0002132,      & [\textbf{stromal cell of ovary (em)},                              \\
                            & Ovary]                  & CL:4030040]       & epithelial cell of uterus (m),                                     \\
                            &                         &                   & ovarian surface epithelial cell (m),                               \\
                            &                         &                   & theca cell (m)]                                                    \\
  \hline
  Pancreas-Acinar           & [Pancreas]              & [CL:0002064]      & [\textbf{pancreatic acinar cell (em)}]                             \\
  \hline
  Pancreas-Alpha            & [Pancreas]              & [CL:0000171]      & [\textbf{pancreatic A cell (em)}]                                  \\
  \hline
  Pancreas-Beta             & [Pancreas]              & [CL:0000169]      & [\textbf{type B pancreatic cell (em)}]                             \\
  \hline
  Pancreas-Delta            & [Pancreas]              & [CL:0000173]      & [\textbf{pancreatic D cell (em)}]                                  \\
  \hline
  Pancreas-Duct             & [Pancreas]              & [CL:0002079]      & [\textbf{pancreatic ductal cell (em)}]                             \\
  \hline
  Prostate-Ep               & [Prostate]              & [CL:0002231]      & [basal cell of prostate epithelium (m),                            \\
                            &                         &                   & luminal cell of prostate epithelium (m)]                           \\
  \hline
  Skeletal-Musc             & [Skeletal muscle]       & [CL:0000188]      & [fast muscle cell (m),                                             \\
                            &                         &                   & skeletal muscle satellite stem cell (m),                           \\
                            &                         &                   & slow muscle cell (m)]                                              \\
  \hline
  Small-Int-Ep              & [Jejunum,               & [CL:0000164,      & [BEST4+ enterocyte (m),                                            \\
                            & Small intestine]        & CL:1000334]       & \textbf{enteroendocrine cell (em)},                                \\
                            &                         &                   & enterocyte of epithelium proper of duodenum (m),                   \\
                            &                         &                   & enterocyte of epithelium proper of ileum (m),                      \\
                            &                         &                   & enterocyte of epithelium proper of jejunum (m),                    \\
                            &                         &                   & enterocyte of epithelium proper of small intestine (m),            \\
                            &                         &                   & enteroendocrine cell of small intestine (m),                       \\
                            &                         &                   & intestinal crypt stem cell of small intestine (m),                 \\
                            &                         &                   & intestinal tuft cell (m),                                          \\
                            &                         &                   & paneth cell of epithelium of small intestine (m),                  \\
                            &                         &                   & small intestine goblet cell (m),                                   \\
                            &                         &                   & transit amplifying cell of small intestine (m)]                    \\
  \hline
  Smooth-Musc               & [Aorta,                 & [CL:0002539,      & [\textbf{bronchial smooth muscle cell (em)},                       \\
                            & Bladder,                & CL:0002592,       & blood vessel smooth muscle cell (m),                               \\
                            & Bronchial,              & CL:0002597,       & myometrial cell (m),                                               \\
                            & Cornary artery,         & CL:0002598,       & smooth muscle cell (m),                                            \\
                            & Prostate]               & CL:1000487]       & vascular associated smooth muscle cell (m)]                        \\
  \hline
  Thyroid-Ep                & [Thyroid]               & [CL:0002257]      & ---                                                                \\
  \bottomrule
\end{longtable}

\newpage
\section{Details of experimental setups}
\label{sect:supp_details_experimental_setups}

\subsection{Split construction}
\label{sect:supp_split_construction}

To construct the training, validation and test splits, we follow a procedure
that ensures every target cell type is represented in all splits while
isolating biological sources as much as possible. At the beginning of the
procedure, we aim for a 70:15:15 ratio between the splits. Target cell types do
not have an equal number of biological samples covering them. Therefore, we
establish the following rules, which we follow in the stated order, and in such
a way that rule $j$ is only applied in a way that does not contradict all rules
$i < j$:

\begin{enumerate}
      \item If a target cell type is covered by a single sample, then the data from this
            sample is divided across all three splits.
      \item If a target cell type is covered by exactly two samples, one is allocated to
            train, and the other is split between validation and test sets.
      \item If a target cell type is covered by three or more biological samples, those
            samples must not be individually broken down into separate splits, and the
            validation and test splits must receive at least one full sample.
\end{enumerate}

Applying the described procedure results in the final train/val/test ratios
62:19:19 with four samples shared across all three splits (GEO
\textit{GSM5652204, GSM5652218,GSM5652321, GSM5652358}) and two samples shared
across validation and test split (GEO \textit{GSM5652199, GSM5652205}) from a
total of 205 samples.

\subsection{Pseudobulk generation procedure}
\label{supp:pseudobulk_sampling}

\Cref{alg:pseudo_bulk} describes the sampling procedure used to generate the pseudobulk mixtures employed in this paper.
The key parameter is the total size~$N$ of the pseudobulk, \textit{i.e.} the number of reads to subsample per split.
Across all splits, the number of reads per originating cell type ranged from~$4{,}270$ to~$464{,}002$, with a mean of~$59{,}977$.
We used $N = 475{,}000$ across all experiments, resulting in~$475{,}000 / 39 \approx 12{,}179$ reads per GR group, per pseudobulk, per split.
This value was set heuristically to balance three competing objectives: sufficient representation of each cell type in the resulting mixtures (for which a higher sample count is beneficial); useful sampling noise to improve generalization of the trained model (for which sample counts that do not match the original cell-type contributions are beneficial); and computational efficiency (for which a lower sample count is beneficial).
The generated pseudobulks are aggregated using \Cref{alg:aggregate_dmr}.

\begin{algorithm}[hbtp]
    \caption{Pseudobulk mixture generation}\label{alg:pseudo_bulk}
    \begin{algorithmic}[1]
        \Statex \textbf{Input:} Set of splits $\mathcal{S} = \{S_1, \dots, S_K\}$, each containing read-level predictions;
        \textit{UXM} atlas $A$ and reference cell names $\mathbf{r}$;
        Number of GR groups $G$;
        Total number of reads to sample $N$;
        Total number of cell types $C \geq 10$
        \Statex \textbf{Output:} The cell types mixed in the pseudobulk, their effective proportions and the sampled reads for each split.
        \Statex
        \State $C_{mix} \sim \text{Uniform}(\{1, \ldots, 10\})$ \Comment{Sample number of cell types to mix}
        \State $c_1, \dots, c_{C_{mix}} \gets$ sample $C_{mix}$ elements without replacement from $\{1, \dots, C\}$
        \State $\boldsymbol{\pi} \gets$ \Call{SampleProportions}{$C_{mix}$} \Comment{$C_{mix}$ uniform samples then normalized s.t.\ $\sum_j \pi_j = 1$}
        \State $n_j \gets \lfloor N \cdot \pi_j \rfloor$ for $j = 1, \dots, C_{mix}$ \Comment{Number of reads to sample for each cell type}
        \State $N' \gets \sum_{j=1}^{C_{mix}} n_j$ \Comment{Effective total number of reads}
        \State $\hat{\boldsymbol{\pi}} \gets \mathbf{0}^{C_{mix}}$ \Comment{Effective proportion vector}
        \For{$j = 1, \dots, C_{mix}$}
        \State $\hat{\pi}_{c_j} \gets n_j / N'$
        \EndFor
        \ForAll{split $S_k \in \mathcal{S}$}
        \State $\mathcal{R}^k \gets \emptyset$ \Comment{Set of sampled reads for split $S_k$}
        \For{$j = 1, \dots, C_{mix}$} \Comment{Iterate over cell types to mix}
        \State $(m_{j,1}, \dots, m_{j,G}) \sim \text{Multinomial}(n_j;\, \tfrac{1}{G}, \dots, \tfrac{1}{G})$ \Comment{Allocate read counts to GR groups}
        \For{$g = 1, \dots, G$} \Comment{Iterate over GR groups}
        \State $\mathcal{R}^k_g \gets \{r \in S_k \mid r \text{ originates from } c_j \wedge r \text{ is in GR group } g\}$
        \State $\mathcal{R}^k \gets \mathcal{R}^k \cup \{\text{sampled } m_{j,g} \text{ reads from } \mathcal{R}^k_g \text{ with replacement}\}$
        \EndFor
        \EndFor
        \EndFor
        \State \Return $(c_1, \dots, c_{C_{mix}}),\; \hat{\boldsymbol{\pi}},\; \{\mathcal{R}^k\}_{1 \leq k \leq K}$
    \end{algorithmic}
\end{algorithm}

\subsection{Predictions aggregation}
\label{supp:predictions_aggregation}

Before deconvolution, we aggregate the model's per-read predictions into a
compact matrix, where each row corresponds to a GR group and each column to a
candidate cell type of origin. This matrix summarises the cell-of-origin
information (or prediction information) that the model extracted from the
mixture at the read level.

We aggregate using a weighted average, where each read's weight is proportional
to the number of CpGs it covers. The intuition is that reads carrying more CpGs
convey more evidence about their cell of origin and are statistically more
reliable. For long reads with many CpGs, this weighting is essentially a no-op,
but in our setting, most reads cover only a handful of CpGs (between four to
seven), so down-weighting the shortest reads has a measurable effect.

A second issue arises at inference time. During training, every GR group is
well represented, and aggregation is straightforward. In real samples, however,
it is common for no read to map to a particular GR group, leaving an entire row
of the matrix undefined. In the ODD dataset analyzed in this work, 235 of 260
samples containe one GR group specific to colon fibroblasts with zero reads,
and remaining 25 samples contain no more than five reads from this group. This
is because the RRBS protocol produces less output than the WGBS data used for
training, and coverage gaps are likely due to our employed atlas having up to
25 DMRs per GR group (with each of our GR groups being specific to a single
cell type and only 3 DMRs for colon fibroblasts).

To handle this, we precompute a \emph{uniform prior} matrix that serves as a
sensible default for under-observed rows. Concretely, we generate $C$ pure
pseudobulk mixtures, one per cell type, each made entirely of reads from a
single cell type $c$. Then we aggregate each of them and average the resulting
matrices. The result approximates the aggregated matrix one would obtain from a
mixture in which all cell types contribute equally, and provides our best
uninformed guess for any GR group we fail to observe. Each row of the
aggregated matrix obtained from prediction is then shrunk toward the
corresponding prior row, with the strength of shrinkage determined by how many
reads were actually observed for that GR group. Thus, well-covered rows are
barely affected, while empty rows are replaced entirely by the prior. More
sophisticated imputation schemes that exploit conditional dependencies between
rows are possible, but lie outside the scope of this work.

Algorithm~\ref{alg:aggregate_dmr} describes the aggregation step in detail, and
Algorithm~\ref{alg:uniform_prior} describes how the uniform prior is computed.

\begin{algorithm}[hbtp]
    \caption{Aggregate read-level predictions at the GR group level}\label{alg:aggregate_dmr}
    \begin{algorithmic}[1]

        \Statex \textbf{Input:} Dataset $\{(r_i, \mathbf{u_i}, g_i, w_i)\}_{i=1}^N$ where $r_i$ is a read, $\mathbf{u_i} \in \Delta^{C-1}$ is the vector of model predictions for that read, $g_i$ is the GR group it belongs to and $w_i$ is the number of CpGs sites covered by the read;
        Uniform prior matrix $Q \in \mathbb{R}^{G \times C}$ (output of \Cref{alg:uniform_prior});
        Prior weight $w_{\text{prior}}\geq 0$ (default is $w_{\text{prior}} = 1$)
        \Statex \textbf{Output:} Aggregated prediction matrix with weighted-average predictions per GR group, blended with prior
        \Statex
        \State $P \gets$ matrix of dimension $G \times C$ initialized to zero \Comment{Aggregated prediction matrix}
        \ForAll{GR group g in $\{1, \dots, G\}$}
        \State $I_g \gets \{i \mid g_i = g\}$ \Comment{Indices of reads in GR group $g$}
        \State $W_g \gets \sum_{i \in I_g} w_i$ \Comment{Total weight for GR group $g$}
        \If{$W_g > 0$}
        \ForAll{c in $\{1, \dots, C\}$}
        \State $P_{g,c} \gets \frac{1}{W_g} \sum_{i \in I_g} w_i \, u_{i,c}$ \Comment{Weighted avg. prediction for GR group $g$ and cell type $c$}
        \State $n_g \gets |I_g|$ \Comment{Number of reads in GR group $g$}
        \EndFor
        \EndIf
        \EndFor

        \State \textbf{// Prior blending: shrink every row towards the uniform prior}
        \ForAll{GR group $g \in \{1, \dots, G\}$}
        \ForAll {c in $\{1, \dots, C\}$}
        \State $P_{g,c} \gets \dfrac{n_g}{n_g + w_{\text{prior}}} \, P_{g,c} + \dfrac{w_{\text{prior}}}{n_g + w_{\text{prior}}}\, Q_{g,c}$
        \EndFor
        \EndFor

        \State \Return $P$
    \end{algorithmic}
\end{algorithm}

\begin{algorithm}[hbtp]
    \caption{Compute uniform prior matrix from pure pseudobulk profiles}\label{alg:uniform_prior}
    \begin{algorithmic}[1]

        \Statex \textbf{Input:} Number of cell types $C$;
        Reads per pure profile $N_{\text{pure}}$
        \Statex \textbf{Output:} Uniform prior matrix $Q \in \mathbb{R}^{G \times C}$
        \Statex
        \For{$c = 1, \dots, C$} \Comment{Generate one pure prediction matrix per cell type}
        \State $\mathcal{R}_c \gets $ sampled reads from \Cref{alg:pseudo_bulk} where the mixture is made entirely of cell type $c$
        \State $P^{(c)} \gets$ \Call{AggregatePredictionsByGRGroup}{$\mathcal{R}_c$} \Comment{\Cref{alg:aggregate_dmr} without prior blending}
        \EndFor

        \Statex \text{//Average pure profiles to get uniform prior}
        \State $Q \gets \frac{1}{C} \sum_{c=1}^C P^{(c)}$

        \State \Return $Q$
    \end{algorithmic}
\end{algorithm}

\subsection{Deconvolution features selection}
\label{supp:dec_features_selection}

\begin{figure}[htbp]
    \centering
    \includegraphics[width=1\textwidth]{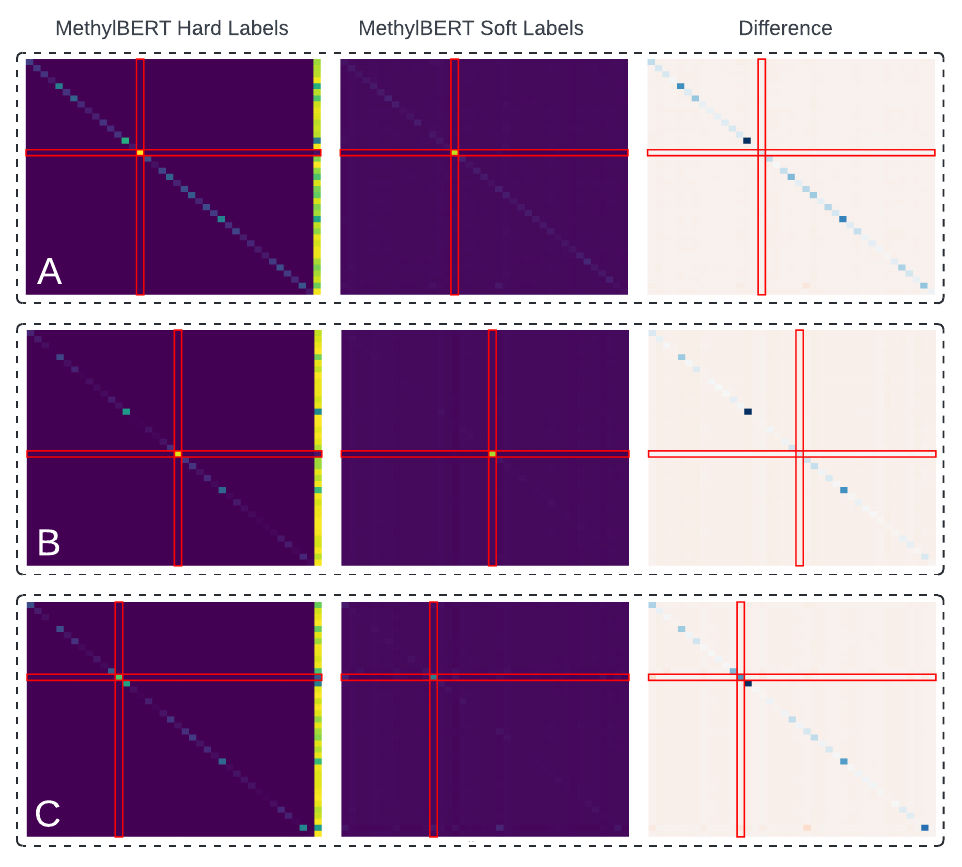}
    \caption{Differences in aggregated prediction matrices for in-silico pure mixtures (the background class column specific to hard labels is omitted). The row and column corresponding to the target cell type are highlighted. \textbf{A}: Mixture is composed solely of Erythrocyte Progenitor reads.
        \textbf{B}: Mixture is composed solely of Heart Cardio reads. \textbf{C}: Mixture is composed solely of Dermal Fibroblast reads.}
    \label{fig:features_heat_maps}
\end{figure}

As depicted in \Cref{fig:framework}, the prediction aggregation step produces a
matrix with dimensions $(G, C)$. In our experiments with soft labels, it yields
a $(39,39)$ matrix and a $(39,40)$ matrix for the models trained on hard labels
with an additional background class. However, not every element in the
resulting matrix is equally informative. First, for well-separated cell types
where the resulting data-driven soft labels will not be much different from the
hard labels for the same instance, the diagonal will be the most informative
part of the matrix regardless of the underlying labeling scheme. Second, for
the hard labeling scheme, no cross-talk off-diagonal is expected; even for the
soft labeling scheme, considering that most of the regions in the atlas are
highly specific (cf.~\Cref{sect:supp_atlas_analysis}), only some of the
off-diagonal features will be informative. \Cref{fig:features_heat_maps}
illustrates this considerations. For example, \textit{erythrocyte progenitors},
the most specific entry in the atlas according to the Proxy GSS, also confound
GR groups specific to other cell types, as evidenced by higher diagonal scores
in the pure mixture across both hard and soft labeling schemes. In contrast,
\textit{heart cardiomyocytes}, another well-separated cell type, almost does
not "fire" anywhere else except in its target region when the soft labeling
scheme is applied. Finally, \textit{dermal fibroblast}, which represents one of
the least specific cell types in the reference used, shows some off-diagonal
signal captured only by the soft labeling scheme. The hard labeling scheme,
however, might provide additional information in the background class.

Considering that most of the data in the resulting matrices is not informative,
we decided to apply a filter before deconvolution, specific to each labeling
scheme. For soft labeling, we keep only 156 (roughly 10\%) of the most relevant
features. To score each feature, we computed the feature matrices for all
in-silico mixtures with a single target cell type. We then calculated the
maximum and mean values for each matrix element across the resulting 39
profiles. The final score was calculated as the ratio between these two values.
We ranked all features by descending score, selected all diagonal features
regardless of their scores, and added the remaining top 117 features, resulting
in 156 features in total. For hard labeling, in addition to using the
\textit{Top 156} features (excluding the background column), we also use a
feature selection scheme that selects only the diagonal and the background
column. Indeed, as can be seen from \Cref{tab:supp_features_selection},
\textit{Top 156} performs almost as well as the full feature matrix for the
soft labeling scheme, while for hard labeling, the performance is no different
when either only the diagonal or any number of additional features is used.
This corroborates our hypothesis that the hard labeling scheme is steering the
classifier to eliminate any cross-talk by concentrating on separating on-target
from off-target. Therefore, the \textit{PSLS} deconvolver treats off-diagonal
information as noise and ignores it; also, it has to be noted that this is not
guaranteed, as one can envision scenarios where spurious correlations may still
occur between off-diagonal values and true proportions. Furthermore, by looking
at the detailed results (\Cref{sect:supp_pseudobulk_exp_detailed_results}), one
can observe that another unsupervised deconvolver, \textit{NNLS}, obtains
better results under the \textit{Diagonal + Bkg.} feature scheme compared to
\textit{Top 156} features ($1.61 \times 10^-4$ vs $1.82 \times 10^-4$).
\textit{PSLS} robustness is likely to stem from the enforced at optimization
time sum-to-one constraint, which acts as regularization.

\begin{figure}[htbp]
  \centering
  \includegraphics[width=1.0\textwidth]{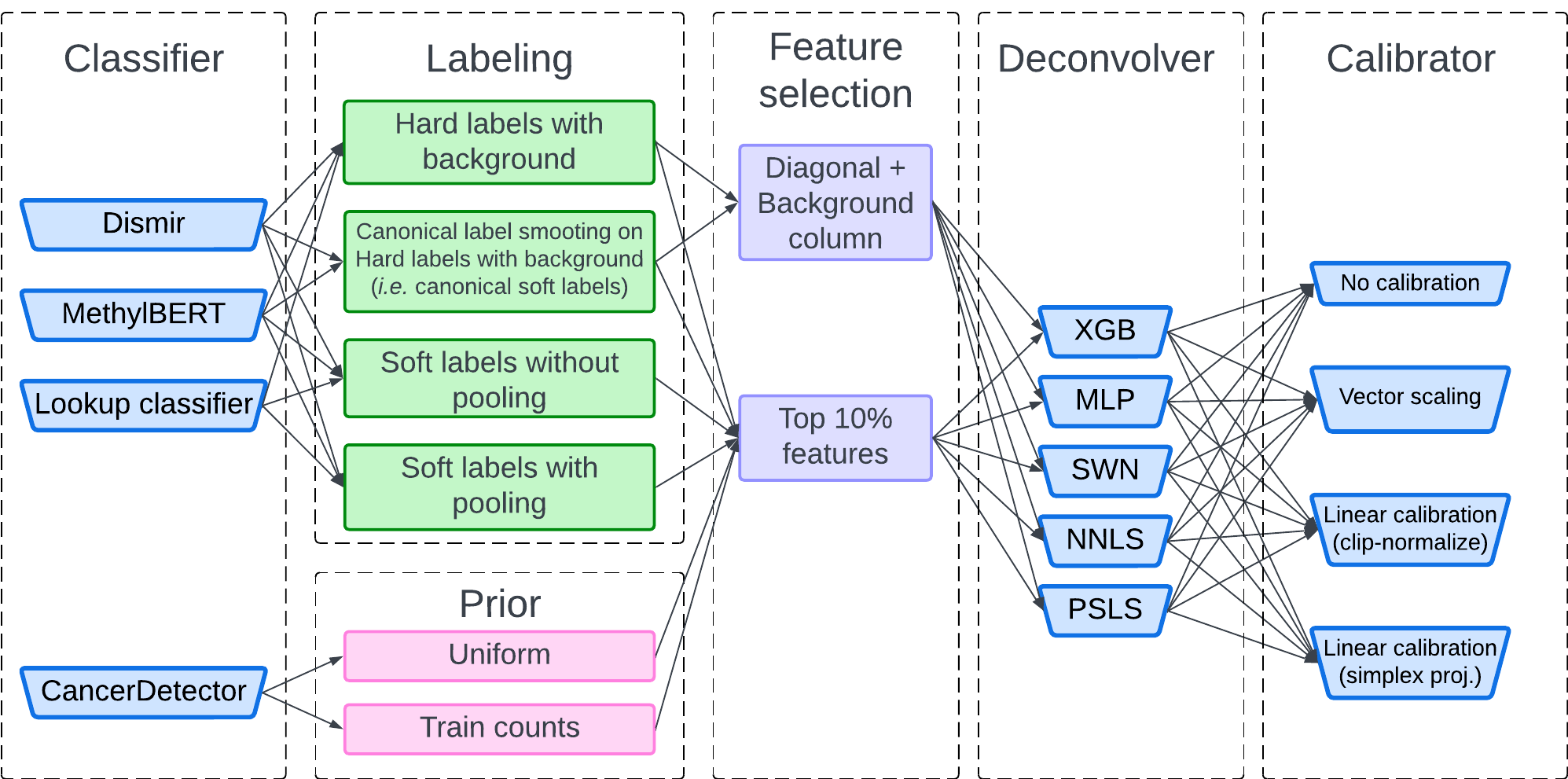}
  \caption{Representation of the design choices when constructing an instance of \textit{Syto}.
    Each path from classifier to calibrator is a possible choice.
    There is a total of 360 possible paths, and we explore all of them in our experiments ($360 = (2\times(2\times2+2\times1)+1\times(1\times2+2\times1)+1\times2\times1)\times5\times4$).
  }
  \label{fig:recap_experimental_setups}
\end{figure}

As for the soft labeling scheme, it allows cross-talk information to propagate
explicitly, which is why adding more features monotonically improves the
deconvolution accuracy. Although using the full matrix would give us slightly
better results for the soft labeling scheme, we never use it in our other
experiments as the computational cost is prohibitive, especially with the high
number of setups to test (cf.~\Cref{fig:recap_experimental_setups}). Instead,
selecting the top 156 features is considered asymptotically equivalent to using
the entire feature matrix, which is the only setting we utilize across
experiments featuring data-driven soft labels.

\begin{table}[H]
    \centering
    \caption{Comparison of feature selection schemes for \textit{MethylBERT} with \textit{PSLS} deconvolver.}
    \label{tab:supp_features_selection}
    {\footnotesize
        \resizebox{\textwidth}{!}{%
            \begin{tabular}{ll cc cccc cc}
                \toprule
                \multirow{2}{*}{\textbf{Classifier}} & \multirow{2}{*}{\textbf{Feature Selection Scheme}} & \multirow{2}{*}{\textbf{Deconvolver}} & \multirow{2}{*}{\textbf{Dec.\ Calib.}}
                                                     & $\mathbf{R}^\mathbf{2}$                            & \textbf{LoA}                          & \textbf{LoA worst-class}               & \textbf{MAE}
                                                     & \textbf{MSE}                                       & \textbf{KL}                                                                                                                                                 \\
                                                     &                                                    &                                       &
                                                     & ($\times 10^{-2}$)                                 & ($\times 10^{-2}$)                    & ($\times 10^{-2}$)                     & ($\times 10^{-3}$)
                                                     & ($\times 10^{-4}$)                                 & ($\times 10^{-2}$)                                                                                                                                          \\
                \midrule
                \multicolumn{10}{c}{\textit{Soft Labels with Pooling}}                                                                                                                                                                                                  \\
                \midrule
                \multirow{3}{*}{\textit{MethylBERT}}
                                                     & Diagonal                                           & \textit{PSLS}                         & None                                   & 96.73              & [-3.22, 3.22] & [-12.65, 9.04] & 3.837 & 2.71 & 13.62 \\
                \cmidrule(l){2-10}
                                                     & Top 156                                            & \textit{PSLS}                         & None                                   & 97.99              & [-2.53, 2.53] & [-9.94, 7.46]  & 3.175 & 1.67 & 7.36  \\
                \cmidrule(l){2-10}
                                                     & Full Matrix                                        & \textit{PSLS}                         & None                                   & 98.04              & [-2.50, 2.50] & [-9.62, 7.25]  & 3.149 & 1.63 & 7.09  \\
                \midrule
                \multicolumn{10}{c}{\textit{Hard Labels}}                                                                                                                                                                                                               \\
                \midrule
                \multirow{4}{*}{\textit{MethylBERT}}
                                                     & Diagonal                                           & \textit{PSLS}                         & None                                   & 98.03              & [-2.50, 2.50] & [-6.75, 7.36]  & 3.584 & 1.63 & 8.73  \\
                \cmidrule(l){2-10}
                                                     & Diagonal + Bkg.                                    & \textit{PSLS}                         & None                                   & 98.03              & [-2.50, 2.50] & [-6.75, 7.36]  & 3.585 & 1.63 & 8.71  \\
                \cmidrule(l){2-10}
                                                     & Top 156 (w/o background)                           & \textit{PSLS}                         & None                                   & 98.03              & [-2.50, 2.50] & [-6.75, 7.37]  & 3.584 & 1.63 & 8.71  \\
                \cmidrule(l){2-10}
                                                     & Full Matrix                                        & \textit{PSLS}                         & None                                   & 98.04              & [-2.50, 2.50] & [-6.75, 7.37]  & 3.585 & 1.63 & 8.71  \\
                \bottomrule
            \end{tabular}%
        }}
\end{table}

\newpage
\subsection{Deconvolution metrics}
\label{sect:supp_deconv_metrics}

In this section, we describe the metrics used to evaluate the deconvolution
results.

\paragraph{Deconvolution metrics for the pseudobulk experiment.}
We recall the notations used in the main paper. Let $K$ be the number of
pseudobulks, $(\mathbf{p}_k)_{1 \leq k \leq K}$ be the ground truth mixture
proportions and $(\mathbf{\tilde{p}}_k)_{1 \leq k \leq K}$ be the final
predictions.

$R^2$ is computed as the coefficient of determination across all
pseudobulks, variance-weighted per cell-type. It is computed as
\verb|sklearn.metrics.r2_score(y_true, y_pred, multioutput="variance_weighted")|
in the \texttt{scikit-learn} library and the close form writes as:

\begin{equation}
  R^2 \coloneq 1 - \frac{\sum_{k=1}^{K} \sum_{c=1}^C (p_{k,c} - \tilde{p}_{k,c})^2}{\sum_{k=1}^{K} \sum_{c=1}^C \left(p_{k,c} - \frac{1}{K}\sum_{k'=1}^K p_{k',c}\right)^2}.
\end{equation}

The \textbf{MSE} is computed as the average Mean Squared Error over all
pseudobulks:

\begin{equation}
  \text{MSE} \coloneq \frac{1}{K} \sum_{k=1}^{K} \frac{1}{C} \sum_{c=1}^C (p_{k,c} - \tilde{p}_{k,c})^2.
\end{equation}

Similarly, the \textbf{MAE} is computed as the average Mean Absolute Error over
all pseudobulks:

\begin{equation}
  \text{MAE} \coloneq \frac{1}{K} \sum_{k=1}^{K} \frac{1}{C} \sum_{c=1}^C |p_{k,c} - \tilde{p}_{k,c}|.
\end{equation}

The \textbf{KL} divergence is also computed as the average Kullback-Leibler
divergence over all pseudobulks:

\begin{equation}
  \text{KL} \coloneq \frac{1}{K} \sum_{k=1}^{K} \sum_{c=1}^C p_{k,c} \log\left( \frac{p_{k,c}}{\tilde{p}_{k,c}} \right).
\end{equation}

where both $p_{k,c}$ and $\tilde{p}_{k,c}$ are lower clipped to a small
$\epsilon$ for numerical stability in the logarithm (we used $\epsilon =
  10^{-8}$).

\textbf{LoA} denotes the overall Limits of Agreements defined as $\text{LoA} \coloneq [\text{LoA}_\text{lower}, \text{LoA}_\text{upper}]$ where:
\begin{align}
  \text{LoA}_\text{lower} & \coloneq \text{bias} - 1.96 \times \text{std},                                              \\
  \text{LoA}_\text{upper} & \coloneq \text{bias} + 1.96 \times \text{std},                                              \\
  \text{bias}             & \coloneq \frac{1}{K} \sum_{k=1}^{K} \frac{1}{C} \sum_{c=1}^C (p_{k,c} - \tilde{p}_{k,c}),   \\
  \text{std}              & \coloneq \sqrt{\frac{1}{KC - 1} \sum_{k=1}^{K} \sum_{c=1}^C (p_{k,c} - \tilde{p}_{k,c})^2}.
\end{align}

\textbf{LoA worst-class} denotes the worst Limits of Agreements per class. For all $c \in \mathcal{C}$ we define the class LoA as $\text{LoA}_c \coloneq [\text{LoA}_{\text{lower}, c}, \text{LoA}_{\text{upper}, c}]$ where:

\begin{align}
  \text{LoA}_{\text{lower}, c} & \coloneq \text{bias}_c - 1.96 \times \text{std}_c,                            \\
  \text{LoA}_{\text{upper}, c} & \coloneq \text{bias}_c + 1.96 \times \text{std}_c,                            \\
  \text{bias}_c                & \coloneq \frac{1}{K} \sum_{k=1}^{K} (p_{k,c} - \tilde{p}_{k,c}),              \\
  \text{std}_c                 & \coloneq \sqrt{\frac{1}{K - 1} \sum_{k=1}^{K} (p_{k,c} - \tilde{p}_{k,c})^2}.
\end{align}

Then, for all classes $c \in \mathcal{C}$, we then compute the LoA width
$\text{LoA}_{\text{width},c} \coloneq \text{LoA}_{\text{upper}, c} -
  \text{LoA}_{\text{lower}, c}$ and we report the class LoA with largest width.

\paragraph{Deconvolution metrics for the OOD RRBS dataset.}
\label{sect:supp_par_TCS}
As mentionned in the main paper, the OOD dataset is a dataset of samples originating from known tissues, but with unknown cell type proportions.
We inferred the expected cell types in each tissue following the procedure described in \Cref{sect:par_ood_rrbs_experiment} and \Cref{sect:supp_mapping_cfsort_tissues_to_tabula_sapiens}.
To evaluate the deconvolution results, we rely on the insight that cell types which are not expected in the sample's tissue should not appear in the deconvolution results.
This is equivalent to saying that the predicted proportions of cell types expected in the sample's tissue should sum to a high value, ideally one.
We define the ``Tissue Concordance Score'' (or \textit{TCS}) as the sum of predicted proportions of cell types expected in the sample's tissue.
Formally, let $\mathbf{\tilde{p}} \in \Delta^{C-1}$ be the predicted mixture proportions for a sample of tissue $t$ and let $\mathcal{C}_t \subseteq \mathcal{C}$ be the set of cell types expected in tissue $t$ according to the mapping described in \Cref{sect:par_ood_rrbs_experiment} and \Cref{sect:supp_mapping_cfsort_tissues_to_tabula_sapiens}.
Then, the Tissue Concordance Score is defined as:
\begin{equation}
  \text{TCS} \coloneq \sum_{c \in \mathcal{C}_t} \tilde{p}_c.
\end{equation}

\subsection{Possible combinations of elements in \texorpdfstring{\textit{Syto}}{Syto}}
\label{sect:supp_combinations_elements_Syto}

\newpage
\section{Integral experimental results}

\raggedbottom
\makeatletter
\renewcommand{\topfraction}{0.9}
\renewcommand{\bottomfraction}{0.6}
\renewcommand{\textfraction}{0.07}
\renewcommand{\floatpagefraction}{0.85}
\setlength{\@fptop}{0\p@}
\setlength{\@fpsep}{12\p@}
\setlength{\@fpbot}{0\p@ \@plus 1fil}
\makeatother

\subsection{Joint evaluation of pseudobulk and OOD results}
\label{sect:bulk_tcs_joint_evaluation}

\begin{figure}[H]
  \centering
  \includegraphics[width=\textwidth]{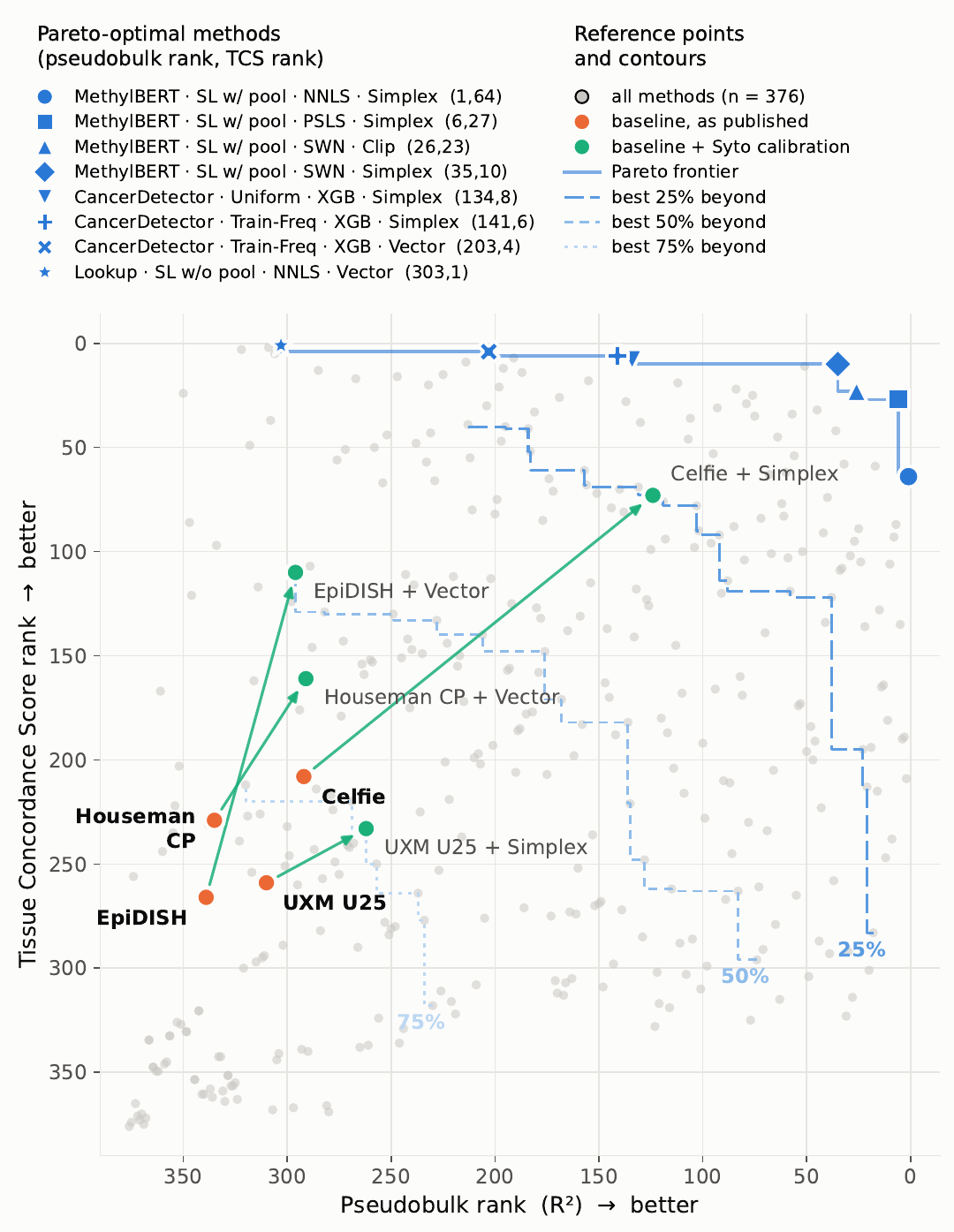}
  \caption{Joint ranking of examined 360 \textit{Syto} variants and \textit{baselines} as measured by pseudobulk's $R^2$ and OOD TCS\@.
    The best methods are in the top right corner.}
  \label{fig:joint_pareto_front}
\end{figure}

\begin{figure}[H]
  \centering
  \includegraphics[width=\textwidth]{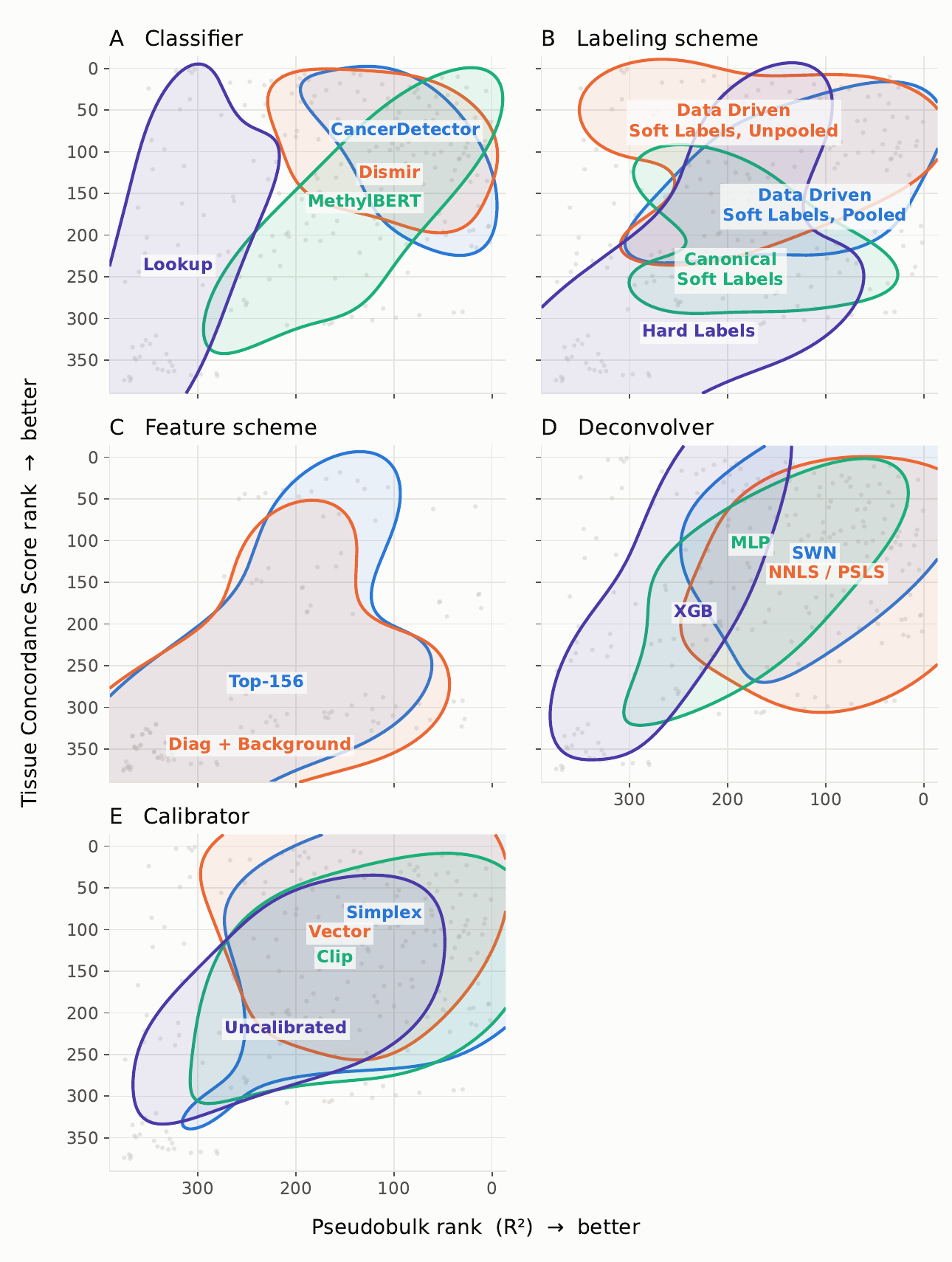}
  \caption{KDE contours of joint metrics plane for the methods' results grouped by different \textit{Syto} configuration choices.
    Contours outline the highest-density region containing 50\% of the fitted KDE probability mass represented on the grid, drawn from a two-dimensional Gaussian kernel density estimate (\texttt{scipy.stats.gaussian\_kde}, bandwidth factor $0.70$, evaluated on
    a $140\times140$ grid). Contour labels are anchored to their corresponding groups' medians.
    All panels exclude \textit{Diag + Background} feature scheme, except panel \textbf{(C)} that is built
    solely on \textit{Hard Labels} results, where both \textit{Diag + Background} and \textit{Top-156} choices are present.
    \textit{Canonical Soft Labels} results are only reflected in panel \textbf{(B)} and excluded elsewhere.
    \textit{NNLS} and \textit{PSLS} in panel \textbf{(D)} are statistically indistinguishable and therefore represented with the same contour.}
  \label{fig:methods_contours}
\end{figure}

\begin{figure}[H]
  \centering
  \includegraphics[width=\textwidth]{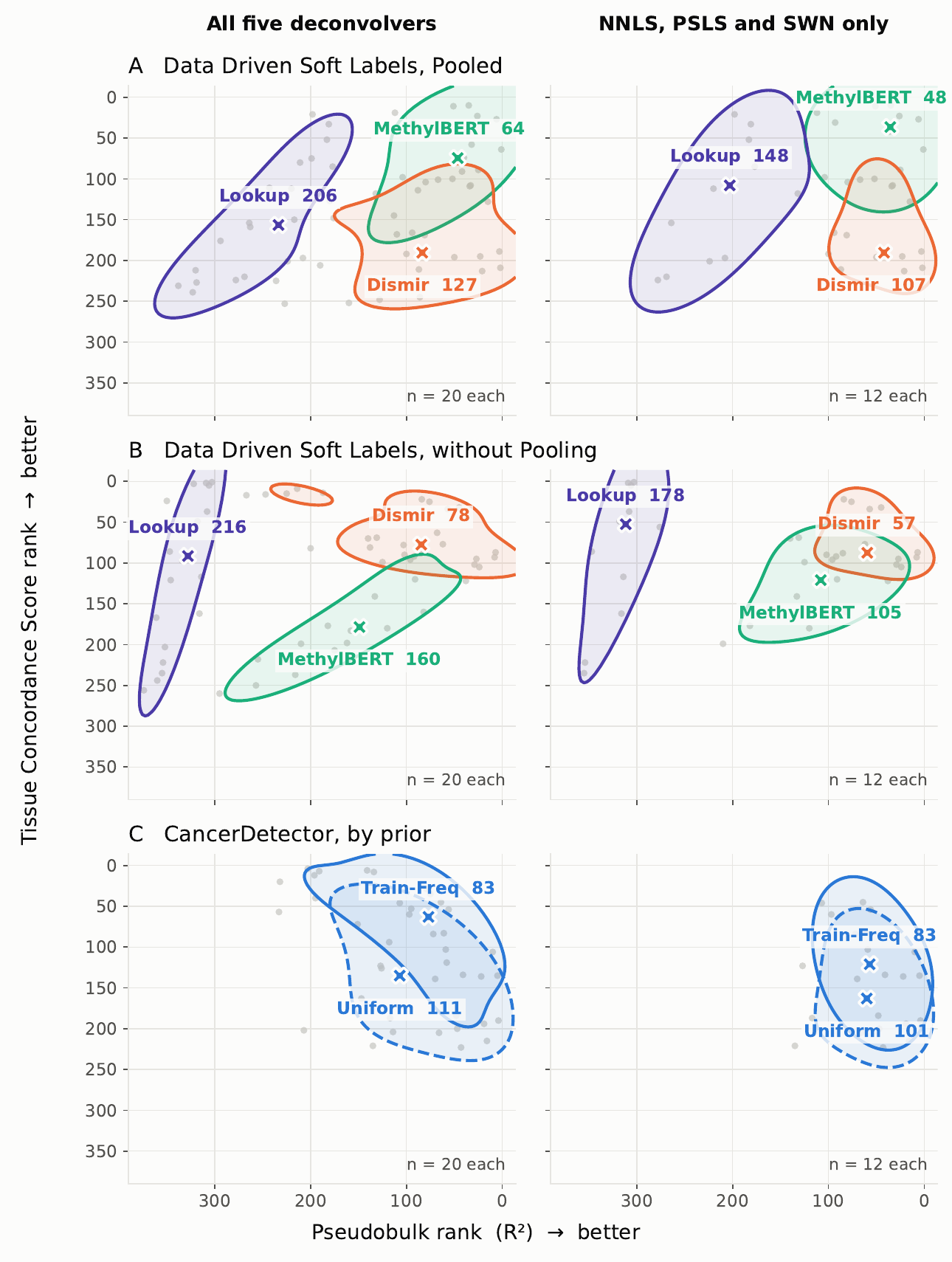}
  \caption{KDE contours of joint metrics plane for a subset of \textit{Syto} configurations.
    Contours outline the highest-density region containing 50\% of the fitted KDE probability mass represented on the grid, drawn from a two-dimensional Gaussian kernel density estimate.
    (\texttt{scipy.stats.gaussian\_kde}, bandwidth factor $0.70$ for $n=20$ and $0.95$ for $n=12$, evaluated on
    a $140\times140$ grid). Panels \textbf{(A)} and \textbf{(B)} represent two best labeling schemes. Panel \textbf{(C)} is focused on \textit{CancerDetector} with different priors. The colored $\times$ markers represent each group's median combined rank.
    \textit{Syto} configurations grounded in \textit{MethylBERT} fitted on \textit{Data Driven Labels with Pooling} outperform options built on any other combination of labeling scheme and classifier.}
  \label{fig:soft_labels_contours}
\end{figure}

\subsection{Pseudobulk experiment detailed results}
\label{sect:supp_pseudobulk_exp_detailed_results}

\begin{table}[H]
    \centering
    \caption{Full results of the pseudobulk experiment for all baselines, with and without our calibration methods. For $R^2$, MAE, MSE and KL, we indicated first the value of the metric computed on the whole test set, and then the 95\% confidence interval computed with BCa bootstrap~\cite{efron1987bca} with 10,000 resamples.}
    \label{tab:supp_uxm}
    {\footnotesize
        \resizebox{\textwidth}{!}{%
\begin{tabular}{ll cc cccc}
                \toprule
                \multirow{2}{*}{\textbf{Baseline}}                                & \multirow{2}{*}{\textbf{Dec.\ Calib.}}
                                                                                  & $\mathbf{R}^\mathbf{2}$                & \textbf{LoA}                      & \textbf{LoA worst-class} & \textbf{MAE}
                                                                                  & \textbf{MSE}                           & \textbf{KL}                                                                                                                                                                                \\
                                                                                  &                                        & ($\times 10^{-2}$)                & ($\times 10^{-2}$)       & ($\times 10^{-2}$) & ($\times 10^{-3}$)
                                                                                  & ($\times 10^{-4}$)                     & ($\times 10^{-2}$)                                                                                                                                                                         \\
                \midrule
                \multirow{4}{*}{\textit{CelFiE}~\cite{caggiano2021comprehensive}} & None                                   & 96.05 {\scriptsize[96.00, 96.10]} & [-3.55, 3.55]            & [-9.96, 9.15]      & 6.42 {\scriptsize[6.40, 6.44]}    & 3.28 {\scriptsize[3.24, 3.32]} & 14.29 {\scriptsize[14.23, 14.34]} \\
                                                                                  & Lin.\ clip+norm                        & 97.41 {\scriptsize[97.39, 97.44]} & [-2.87, 2.87]            & [-6.30, 7.98]      & 4.33 {\scriptsize[4.31, 4.34]}    & 2.15 {\scriptsize[2.12, 2.17]} & 10.44 {\scriptsize[10.36, 10.53]} \\
\bottomrule
            \end{tabular}%
        }}
\end{table}

\begin{table}[H]
    \centering
    \caption{Full results for our extension of \textbf{CancerDetector}~\cite{li2018cancerdetector}. For $R^2$, MAE, MSE and KL, we indicated first the value of the metric computed on the whole test set, and then the 95\% confidence interval computed with BCa bootstrap~\cite{efron1987bca} with 10,000 resamples.}
    \label{tab:supp_cancer_detector}
    {\footnotesize
        \resizebox{\textwidth}{!}{%
            \begin{tabular}{l cc cccc cc}
                \toprule
                \multirow{2}{*}{\textbf{Prior}} & \multirow{2}{*}{\textbf{Deconvolver}} & \multirow{2}{*}{\textbf{Dec.\ Calib.}}
                                                & $\mathbf{R}^\mathbf{2}$               & \textbf{LoA}                           & \textbf{LoA worst-class}          & \textbf{MAE}
                                                & \textbf{MSE}                          & \textbf{KL}                                                                                                                                                                                                        \\
                                                &                                       &
                                                & ($\times 10^{-2}$)                    & ($\times 10^{-2}$)                     & ($\times 10^{-2}$)                & ($\times 10^{-3}$)
                                                & ($\times 10^{-4}$)                    & ($\times 10^{-2}$)                                                                                                                                                                                                 \\
                \midrule
                \multirow{20}{*}{Training Counts}
                                                & \multirow{4}{*}{XGB}                  & None                                   & 97.32 {\scriptsize[97.27, 97.37]} & [-2.92, 2.92]      & [-8.14, 8.45] & 3.68 {\scriptsize[3.66, 3.70]} & 2.22 {\scriptsize[2.18, 2.26]} & 6.47 {\scriptsize[6.43, 6.52]} \\
                                                &                                       & Lin.\ clip+norm                        & 97.66 {\scriptsize[97.62, 97.71]} & [-2.73, 2.73]      & [-8.07, 8.14] & 3.37 {\scriptsize[3.35, 3.38]} & 1.94 {\scriptsize[1.90, 1.98]} & 6.47 {\scriptsize[6.41, 6.52]} \\
\bottomrule
            \end{tabular}%
        }}
\end{table}

\begin{table}[H]
    \centering
    \caption{Full results for \textbf{hard labels with background class and Top 156 features for deconvolution}. For $R^2$, MAE, MSE and KL, we indicated first the value of the metric computed on the whole test set, and then the 95\% confidence interval computed with BCa bootstrap~\cite{efron1987bca} with 10,000 resamples.}
    \label{tab:supp_hard_labels_top10}
    {\footnotesize
        \resizebox{\textwidth}{!}{%
            \begin{tabular}{l cc cccc cc}
                \toprule
                \multirow{2}{*}{\textbf{Classifier}} & \multirow{2}{*}{\textbf{Deconvolver}} & \multirow{2}{*}{\textbf{Dec.\ Calib.}} & $\mathbf{R}^\mathbf{2}$           & \textbf{LoA}       & \textbf{LoA worst-class} & \textbf{MAE}                      & \textbf{MSE}                      & \textbf{KL}                       \\
                                                     &                                       &                                        & ($\times 10^{-2}$)                & ($\times 10^{-2}$) & ($\times 10^{-2}$)       & ($\times 10^{-3}$)                & ($\times 10^{-4}$)                & ($\times 10^{-2}$)                \\
                \midrule
                \multirow{20}{*}{\shortstack[l]{\textit{Dismir}~\cite{li2021dismir}                                                                                                                                                                                                                                                   \\(CNN+LSTM)}}
                                                     & \multirow{4}{*}{XGB}                  & None                                   & 96.99 {\scriptsize[96.93, 97.05]} & [-3.10, 3.10]      & [-8.21, 8.72]            & 3.79 {\scriptsize[3.77, 3.81]}    & 2.50 {\scriptsize[2.45, 2.55]}    & 6.69 {\scriptsize[6.64, 6.74]}    \\
                                                     &                                       & Lin.\ clip+norm                        & 97.36 {\scriptsize[97.31, 97.41]} & [-2.90, 2.90]      & [-7.86, 8.57]            & 3.45 {\scriptsize[3.43, 3.47]}    & 2.19 {\scriptsize[2.14, 2.24]}    & 6.47 {\scriptsize[6.42, 6.53]}    \\
\bottomrule
            \end{tabular}%
        }}
\end{table}

\begin{table}[H]
    \centering
    \caption{Full results for \textbf{hard labels with background class and Diag. + Bckg.\ features for deconvolution}. For $R^2$, MAE, MSE and KL, we indicated first the value of the metric computed on the whole test set, and then the 95\% confidence interval computed with BCa bootstrap~\cite{efron1987bca} with 10,000 resamples.}
    \label{tab:supp_hard_labels_diag_bckg}
    {\footnotesize
        \resizebox{\textwidth}{!}{%
            \begin{tabular}{l cc cccc cc}
                \toprule
                \multirow{2}{*}{\textbf{Classifier}} & \multirow{2}{*}{\textbf{Deconvolver}} & \multirow{2}{*}{\textbf{Dec.\ Calib.}} & $\mathbf{R}^\mathbf{2}$           & \textbf{LoA}       & \textbf{LoA worst-class} & \textbf{MAE}                      & \textbf{MSE}                      & \textbf{KL}                       \\
                                                     &                                       &                                        & ($\times 10^{-2}$)                & ($\times 10^{-2}$) & ($\times 10^{-2}$)       & ($\times 10^{-3}$)                & ($\times 10^{-4}$)                & ($\times 10^{-2}$)                \\
                \midrule
                \multirow{20}{*}{\shortstack[l]{\textit{Dismir}~\cite{li2021dismir}                                                                                                                                                                                                                                                   \\(CNN+LSTM)}}
                                                     & \multirow{4}{*}{XGB}                  & None                                   & 96.95 {\scriptsize[96.90, 96.99]} & [-3.12, 3.12]      & [-7.48, 8.33]            & 4.00 {\scriptsize[3.98, 4.02]}    & 2.54 {\scriptsize[2.49, 2.58]}    & 6.90 {\scriptsize[6.85, 6.94]}    \\
                                                     &                                       & Lin.\ clip+norm                        & 97.37 {\scriptsize[97.32, 97.41]} & [-2.90, 2.90]      & [-7.21, 8.18]            & 3.61 {\scriptsize[3.59, 3.63]}    & 2.18 {\scriptsize[2.14, 2.22]}    & 6.50 {\scriptsize[6.45, 6.55]}    \\
\bottomrule
            \end{tabular}%
        }}
\end{table}

\begin{table}[H]
    \centering
    \caption{Full results for \textbf{Canonical Soft Labels ($\epsilon=0.1$) and Top 156 features for deconvolution}. For $R^2$, MAE, MSE and KL, we indicated first the value of the metric computed on the whole test set, and then the 95\% confidence interval computed with BCa bootstrap~\cite{efron1987bca} with 10,000 resamples.}
    \label{tab:supp_soft_canonical_top10percent}
    {\footnotesize
        \resizebox{\textwidth}{!}{%
            \begin{tabular}{l cc cccc cc}
                \toprule
                \multirow{2}{*}{\textbf{Classifier}} & \multirow{2}{*}{\textbf{Deconvolver}} & \multirow{2}{*}{\textbf{Dec.\ Calib.}} & $\mathbf{R}^\mathbf{2}$           & \textbf{LoA}       & \textbf{LoA worst-class} & \textbf{MAE}                   & \textbf{MSE}                   & \textbf{KL}                       \\
                                                     &                                       &                                        & ($\times 10^{-2}$)                & ($\times 10^{-2}$) & ($\times 10^{-2}$)       & ($\times 10^{-3}$)             & ($\times 10^{-4}$)             & ($\times 10^{-2}$)                \\
                \midrule
                \multirow{20}{*}{\shortstack[l]{\textit{Dismir}~\cite{li2021dismir}                                                                                                                                                                                                                                             \\(CNN+LSTM)}}
                                                     & \multirow{4}{*}{XGB}                  & None                                   & 97.15 {\scriptsize[97.10, 97.19]} & [-3.02, 3.02]      & [-7.43, 8.27]            & 3.89 {\scriptsize[3.87, 3.91]} & 2.37 {\scriptsize[2.33, 2.41]} & 6.68 {\scriptsize[6.64, 6.73]}    \\
                                                     &                                       & Lin.\ clip+norm                        & 97.57 {\scriptsize[97.53, 97.61]} & [-2.78, 2.78]      & [-7.10, 8.07]            & 3.48 {\scriptsize[3.47, 3.50]} & 2.01 {\scriptsize[1.98, 2.05]} & 6.25 {\scriptsize[6.20, 6.30]}    \\
\bottomrule
            \end{tabular}%
        }}
\end{table}

\begin{table}[H]
    \centering
    \caption{Full results for \textbf{Canonical Soft Labels ($\epsilon=0.1$) and Diag.\ + Bckg.\ features for deconvolution}. For $R^2$, MAE, MSE and KL, we indicated first the value of the metric computed on the whole test set, and then the 95\% confidence interval computed with BCa bootstrap~\cite{efron1987bca} with 10,000 resamples.}
    \label{tab:supp_soft_canonical_diag_bckg}
    {\footnotesize
        \resizebox{\textwidth}{!}{%
            \begin{tabular}{l cc cccc cc}
                \toprule
                \multirow{2}{*}{\textbf{Classifier}} & \multirow{2}{*}{\textbf{Deconvolver}} & \multirow{2}{*}{\textbf{Dec.\ Calib.}} & $\mathbf{R}^\mathbf{2}$           & \textbf{LoA}       & \textbf{LoA worst-class} & \textbf{MAE}                   & \textbf{MSE}                   & \textbf{KL}                       \\
                                                     &                                       &                                        & ($\times 10^{-2}$)                & ($\times 10^{-2}$) & ($\times 10^{-2}$)       & ($\times 10^{-3}$)             & ($\times 10^{-4}$)             & ($\times 10^{-2}$)                \\
                \midrule
                \multirow{20}{*}{\shortstack[l]{\textit{Dismir}~\cite{li2021dismir}                                                                                                                                                                                                                                             \\(CNN+LSTM)}}
                                                     & \multirow{4}{*}{\textit{XGB}}         & None                                   & 95.45 {\scriptsize[95.38, 95.52]} & [-3.81, 3.81]      & [-10.15, 10.75]          & 4.84 {\scriptsize[4.82, 4.86]} & 3.78 {\scriptsize[3.72, 3.85]} & 9.45 {\scriptsize[9.39, 9.51]}    \\
                                                     &                                       & Lin.\ clip+norm                        & 96.32 {\scriptsize[96.26, 96.38]} & [-3.43, 3.43]      & [-9.96, 10.10]           & 4.07 {\scriptsize[4.05, 4.10]} & 3.06 {\scriptsize[3.01, 3.11]} & 8.82 {\scriptsize[8.76, 8.89]}    \\
\bottomrule
            \end{tabular}%
        }}
\end{table}

\begin{table}[H]
    \centering
    \caption{Full results for \textbf{data-driven soft-labels (with pooling)}. For $R^2$, MAE, MSE and KL, we indicated first the value of the metric computed on the whole test set, and then the 95\% confidence interval computed with BCa bootstrap~\cite{efron1987bca} with 10,000 resamples.}
    \label{tab:supp_dd_soft_with_pool}
    {\footnotesize
        \resizebox{\textwidth}{!}{%
            \begin{tabular}{l cc cccc cc}
                \toprule
                \multirow{2}{*}{\textbf{Classifier}} & \multirow{2}{*}{\textbf{Deconvolver}} & \multirow{2}{*}{\textbf{Dec.\ Calib.}} & $\mathbf{R}^\mathbf{2}$           & \textbf{LoA}       & \textbf{LoA worst-class} & \textbf{MAE}                   & \textbf{MSE}                   & \textbf{KL}                       \\
                                                     &                                       &                                        & ($\times 10^{-2}$)                & ($\times 10^{-2}$) & ($\times 10^{-2}$)       & ($\times 10^{-3}$)             & ($\times 10^{-4}$)             & ($\times 10^{-2}$)                \\
                \midrule
                \multirow{20}{*}{\shortstack[l]{\textit{Dismir}~\cite{li2021dismir}                                                                                                                                                                                                                                             \\(CNN+LSTM)}}
                                                     & \multirow{4}{*}{XGB}                  & None                                   & 97.45 {\scriptsize[97.40, 97.49]} & [-2.85, 2.85]      & [-7.32, 8.31]            & 3.67 {\scriptsize[3.65, 3.69]} & 2.12 {\scriptsize[2.08, 2.16]} & 6.35 {\scriptsize[6.31, 6.39]}    \\
                                                     &                                       & Lin.\ clip+norm                        & 97.74 {\scriptsize[97.70, 97.78]} & [-2.68, 2.68]      & [-7.16, 7.97]            & 3.37 {\scriptsize[3.36, 3.39]} & 1.88 {\scriptsize[1.84, 1.91]} & 6.16 {\scriptsize[6.11, 6.21]}    \\
\bottomrule
            \end{tabular}%
        }}
\end{table}

\begin{table}[H]
    \centering
    \caption{Full results for \textbf{data-driven soft-labels (without pooling)}. For $R^2$, MAE, MSE and KL, we indicated first the value of the metric computed on the whole test set, and then the 95\% confidence interval computed with BCa bootstrap~\cite{efron1987bca} with 10,000 resamples.}
    \label{tab:supp_dd_soft_without_pool}
    {\footnotesize
        \resizebox{\textwidth}{!}{%
            \begin{tabular}{l cc cccc cc}
                \toprule
                \multirow{2}{*}{\textbf{Classifier}} & \multirow{2}{*}{\textbf{Deconvolver}} & \multirow{2}{*}{\textbf{Dec.\ Calib.}} & $\mathbf{R}^\mathbf{2}$           & \textbf{LoA}       & \textbf{LoA worst-class} & \textbf{MAE}                      & \textbf{MSE}                      & \textbf{KL}                       \\
                                                     &                                       &                                        & ($\times 10^{-2}$)                & ($\times 10^{-2}$) & ($\times 10^{-2}$)       & ($\times 10^{-3}$)                & ($\times 10^{-4}$)                & ($\times 10^{-2}$)                \\
                \midrule
                \multirow{20}{*}{\shortstack[l]{\textit{Dismir}~\cite{li2021dismir}                                                                                                                                                                                                                                                   \\(CNN+LSTM)}}
                                                     & \multirow{4}{*}{XGB}                  & None                                   & 97.08 {\scriptsize[97.03, 97.13]} & [-3.05, 3.05]      & [-10.44, 10.21]          & 3.77 {\scriptsize[3.75, 3.79]}    & 2.42 {\scriptsize[2.38, 2.46]}    & 6.72 {\scriptsize[6.67, 6.77]}    \\
                                                     &                                       & Lin.\ clip+norm                        & 97.40 {\scriptsize[97.35, 97.44]} & [-2.88, 2.88]      & [-9.94, 9.94]            & 3.45 {\scriptsize[3.43, 3.47]}    & 2.16 {\scriptsize[2.12, 2.20]}    & 6.41 {\scriptsize[6.35, 6.46]}    \\
\bottomrule
            \end{tabular}%
        }}
\end{table}

\newpage
\subsection{OOD experiment detailed results}
\label{sect:supp_rrbs_exp_detailed_results}
\suppressfloats[t]

\begin{table}[H]
  \centering
  \caption{Number of (deconvolver, calibrator) configurations achieving higher TCS than \textit{Celfie U25} per classifier, labeling scheme and feature selection strategy.
    Top section shows classifiers trained on labeled data; bottom section shows labeling-independent classifiers.
    Each cell of the table reports the count as \textit{total\,{\scriptsize(strict)}}, where \textit{total} is the number of configurations with higher average TCS (maximum is 20) and \textit{strict} is the number of configurations with mean significantly different based on Wilcoxon signed-rank test ($pvalue \leq 0.05$).
    The TCS is averaged per tissue and averaged again to ensure equal weight of each tissue in the final score.
    Feature selection strategies: \textbf{DiagBckg} uses the combination of diagonal + background column of the feature matrices; \textbf{Top156} uses the top-156 features (cf.~\Cref{supp:dec_features_selection}).}
  \label{tab:rrbs_config_counts}
  {\footnotesize
    \resizebox{\textwidth}{!}{%
      \begin{tabular}{l c c c c c c}
        \toprule
        \multirow{3}{*}{\textbf{Classifier}}
                                                      & \multicolumn{2}{c}{\textbf{Hard Labels w/ bckg.}}
                                                      & \multicolumn{2}{c}{\textbf{Canonical Soft Labels}}
                                                      & \textbf{DD Soft Labels}                            & \textbf{DD Soft Labels}                                                                                            \\
                                                      &                                                    &                         &                      &                     & \textbf{w/o pool.}   & \textbf{w/ pool.}    \\
        \cmidrule(lr){2-3}\cmidrule(lr){4-5}\cmidrule(lr){6-6}\cmidrule(lr){7-7}
                                                      & DiagBckg                                           & Top156                  & DiagBckg             & Top156              & Top156               & Top156               \\
        \midrule
        \multicolumn{7}{l}{\textit{(a) Classifiers trained on labeled data}}                                                                                                                                                    \\
        \midrule
        \textit{Dismir}\cite{li2021dismir} (CNN+LSTM) & 20 {\scriptsize(0)}                                & 20 {\scriptsize(1)}     & 12 {\scriptsize(0)}  & 16 {\scriptsize(0)} & 20 {\scriptsize(11)} & 13 {\scriptsize(4)}  \\
        \textit{MethylBERT}\cite{jeong2025methylbert} & 0 {\scriptsize(0)}                                 & 0 {\scriptsize(0)}      & 0 {\scriptsize(0)}   & 0  {\scriptsize(0)} & 15 {\scriptsize(0)}  & 19 {\scriptsize(10)} \\
        \textit{Lookup Classifier} (1-NN)             & 0 {\scriptsize(0)}                                 & 0 {\scriptsize(0)}      & N/A                  & N/A                 & 16 {\scriptsize(10)} & 14 {\scriptsize(0)}  \\
        \midrule
        \multicolumn{7}{l}{\textit{(b) Labeling-independent classifiers}}                                                                                                                                                       \\
        \midrule
                                                      & Uniform prior                                      & Training counts prior   & \multicolumn{4}{c}{}                                                                     \\
        \midrule
        \textit{CancerDetector}\cite{li2018cancerdetector}
                                                      & 17 {\scriptsize(4)}                                & 20 {\scriptsize(11)}    & \multicolumn{4}{c}{}                                                                     \\
        \bottomrule
      \end{tabular}%
    }}
\end{table}

\begin{table}[H]
  \centering
  \caption{Comparison of TCS (mean and standard deviation in parentheses) against \textit{Celfie U25} for Hard Labeling and Canonical Soft Labeling schemes using the \textbf{top 156 features}.
    Each row corresponds to a deconvolver-calibrator combination, evaluated with three downstream classifiers under hard labels (\textit{Lookup Classifier}, \textit{MethylBERT}, \textit{Dismir}) and two under canonical soft labels (\textit{MethylBERT}, \textit{Dismir}).
    Cells shaded in \textcolor{magenta}{magenta} indicate combinations whose mean TCS is significantly \emph{lower} than that of \textit{Celfie U25}, while cells shaded in \textcolor{teal}{teal} indicate combinations whose mean TCS is significantly \emph{higher} (Wilcoxon signed-rank test, $p \leq 0.05$).
    Color intensity scales with the magnitude of the difference.}
  \label{tab:tcs_vs_uxm_1}
  {\footnotesize
    \resizebox{\textwidth}{!}{%
      \begin{tabular}{llccccc}
        \toprule
                      &                 & \multicolumn{3}{c}{Hard Labels} & \multicolumn{2}{c}{Canonical Soft Labels}                                                                                                  \\
        \cmidrule(lr){3-5} \cmidrule(lr){6-7}
        Deconvolver   & Calibrator      & \textit{Lookup Classifier}      & \textit{MethylBERT}                       & \textit{Dismir}                     & \textit{MethylBERT}                    & \textit{Dismir} \\
        \midrule
        \textit{XGB}  & ---             & \cellcolor{magenta!60}  40.16   & \cellcolor{magenta!60}  25.65             & 70.65  (17.07)                      & \cellcolor{magenta!45}  53.20  (12.61) & 66.95  (16.16)  \\
        \textit{XGB}  & Lin.\ clip+norm & \cellcolor{magenta!60}  40.25   & \cellcolor{magenta!60}  25.95             & 70.68  (17.08)                      & \cellcolor{magenta!37}  55.16  (13.61) & 66.99  (16.16)  \\
        \textit{XGB}  & Lin.\ simplex   & \cellcolor{magenta!60}  40.56   & \cellcolor{magenta!60}  25.98             & 70.73  (17.09)                      & \cellcolor{magenta!32}  56.43  (14.08) & 67.02  (16.18)  \\
        \textit{XGB}  & Vec.\ scaling   & \cellcolor{magenta!60}  44.70   & \cellcolor{magenta!60}  25.13             & 71.20  (17.43)                      & 59.46  (16.01)                         & 67.49  (16.52)  \\
        \textit{MLP}  & ---             & \cellcolor{magenta!60}  14.12   & \cellcolor{magenta!45}  53.09             & 67.09  (16.11)                      & \cellcolor{magenta!29}  57.06  (16.27) & 66.28  (19.42)  \\
        \textit{MLP}  & Lin.\ clip+norm & \cellcolor{magenta!60}  13.89   & \cellcolor{magenta!39}  54.61             & 67.13  (16.14)                      & \cellcolor{magenta!26}  57.86  (17.12) & 66.29  (19.45)  \\
        \textit{MLP}  & Lin.\ simplex   & \cellcolor{magenta!60}  15.11   & \cellcolor{magenta!36}  55.44             & 67.14  (16.14)                      & 58.76  (17.82)                         & 66.35  (19.46)  \\
        \textit{MLP}  & Vec.\ scaling   & \cellcolor{magenta!60}  20.07   & \cellcolor{magenta!31}  56.52             & 67.68  (16.22)                      & \cellcolor{magenta!24}  58.31  (18.33) & 66.73  (19.58)  \\
        \textit{SWN}  & ---             & \cellcolor{magenta!60}  34.26   & \cellcolor{magenta!60}  48.15             & 71.23  (19.01)                      & 62.31   (17.7)                         & 60.81  (25.59)  \\
        \textit{SWN}  & Lin.\ clip+norm & \cellcolor{magenta!60}  30.91   & \cellcolor{magenta!57}  50.18             & 70.97  (18.87)                      & 63.14  (18.19)                         & 60.67  (25.35)  \\
        \textit{SWN}  & Lin.\ simplex   & \cellcolor{magenta!60}  30.96   & \cellcolor{magenta!57}  50.12             & 70.96  (18.88)                      & 64.04  (18.85)                         & 60.54  (25.29)  \\
        \textit{SWN}  & Vec.\ scaling   & \cellcolor{magenta!60}  32.11   & \cellcolor{magenta!49}  52.03             & 71.26  (19.15)                      & 64.41  (19.58)                         & 60.74  (25.86)  \\
        \textit{NNLS} & ---             & \cellcolor{magenta!60}  34.25   & \cellcolor{magenta!57}  49.98             & 69.76  (17.02)                      & \cellcolor{magenta!22}  58.75  (16.69) & 67.95  (17.96)  \\
        \textit{NNLS} & Lin.\ clip+norm & \cellcolor{magenta!60}  33.02   & \cellcolor{magenta!50}  51.86             & 69.68  (17.07)                      & 59.35  (17.28)                         & 67.60  (17.96)  \\
        \textit{NNLS} & Lin.\ simplex   & \cellcolor{magenta!60}  33.94   & \cellcolor{magenta!48}  52.34             & 69.87   (17.1)                      & 61.11  (18.19)                         & 67.75  (17.99)  \\
        \textit{NNLS} & Vec.\ scaling   & \cellcolor{magenta!60}  37.02   & \cellcolor{magenta!42}  53.96             & \cellcolor{teal!29}  71.73  (18.42) & 62.64  (20.65)                         & 70.26   (19.6)  \\
        \textit{PSLS} & ---             & \cellcolor{magenta!60}  33.95   & \cellcolor{magenta!55}  50.53             & 66.74  (18.69)                      & 60.01  (17.82)                         & 67.51  (17.45)  \\
        \textit{PSLS} & Lin.\ clip+norm & \cellcolor{magenta!60}  32.45   & \cellcolor{magenta!49}  52.17             & 66.21  (18.76)                      & 60.49  (18.41)                         & 66.85  (17.51)  \\
        \textit{PSLS} & Lin.\ simplex   & \cellcolor{magenta!60}  32.97   & \cellcolor{magenta!47}  52.58             & 66.50  (18.67)                      & 61.76  (19.14)                         & 67.14  (17.34)  \\
        \textit{PSLS} & Vec.\ scaling   & \cellcolor{magenta!60}  42.12   & \cellcolor{magenta!40}  54.42             & 65.69  (19.88)                      & 62.93   (20.9)                         & 66.90  (19.09)  \\
        \bottomrule
      \end{tabular}
    }}
\end{table}

\begin{table}[H]
  \centering
  \caption{Comparison of TCS (mean and standard deviation in parentheses) against \textit{Celfie U25} for Hard Labeling and Canonical Soft Labeling schemes using the \textbf{Diag.\ + Bckg.\ features}.
    Each row corresponds to a deconvolver-calibrator combination, evaluated with two downstream classifiers under hard labels (\textit{MethylBERT}, \textit{Dismir}) and two under canonical soft labels (\textit{MethylBERT}, \textit{Dismir}).
    Cells shaded in \textcolor{magenta}{magenta} indicate combinations whose mean TCS is significantly \emph{lower} than that of \textit{Celfie U25}, while cells shaded in \textcolor{teal}{teal} indicate combinations whose mean TCS is significantly \emph{higher} (Wilcoxon signed-rank test, $p \leq 0.05$).
    Color intensity scales with the magnitude of the difference.
  }
  \label{tab:tcs_vs_uxm_2}
  {\footnotesize
    \resizebox{\textwidth}{!}{%
      \begin{tabular}{llccccc}
        \toprule
                      &                 & \multicolumn{3}{c}{Hard Labels}        & \multicolumn{2}{c}{Canonical Soft Labels}                                                                                                     \\
        \cmidrule(lr){3-5} \cmidrule(lr){6-7}
        Deconvolver   & Calibrator      & \textit{Lookup Classifier}             & \textit{MethylBERT}                       & \textit{Dismir} & \textit{MethylBERT}                    & \textit{Dismir}                        \\
        \midrule
        \textit{XGB}  & ---             & \cellcolor{magenta!60}  40.16  (16.75) & \cellcolor{magenta!60}  36.94   (8.69)    & 70.30  (16.84)  & \cellcolor{magenta!60}  37.20   (9.18) & 69.52  (17.73)                         \\
        \textit{XGB}  & Lin.\ clip+norm & \cellcolor{magenta!60}  40.25  (17.93) & \cellcolor{magenta!60}  38.77   (9.66)    & 70.33  (16.85)  & \cellcolor{magenta!60}  38.65   (9.97) & 69.55  (17.73)                         \\
        \textit{XGB}  & Lin.\ simplex   & \cellcolor{magenta!60}  40.56  (18.41) & \cellcolor{magenta!60}  38.91   (9.75)    & 70.38  (16.87)  & \cellcolor{magenta!60}  39.10  (10.22) & 69.59  (17.75)                         \\
        \textit{XGB}  & Vec.\ scaling   & \cellcolor{magenta!60}  44.70  (20.77) & \cellcolor{magenta!60}  41.02  (12.07)    & 70.85  (17.18)  & \cellcolor{magenta!60}  39.99   (12.1) & 70.05  (18.09)                         \\
        \textit{MLP}  & ---             & \cellcolor{magenta!60}  13.44   (7.98) & \cellcolor{magenta!60}  42.78   (13.8)    & 66.94  (16.37)  & 57.00   (15.9)                         & 61.40  (21.62)                         \\
        \textit{MLP}  & Lin.\ clip+norm & \cellcolor{magenta!60}  15.09   (8.66) & \cellcolor{magenta!60}  45.04  (14.88)    & 66.89  (16.37)  & 57.65  (16.33)                         & 61.46   (21.6)                         \\
        \textit{MLP}  & Lin.\ simplex   & \cellcolor{magenta!60}  16.48    (7.8) & \cellcolor{magenta!60}  44.89  (15.03)    & 66.86  (16.35)  & 58.21  (16.85)                         & 61.52  (21.62)                         \\
        \textit{MLP}  & Vec.\ scaling   & \cellcolor{magenta!60}  27.49  (13.76) & \cellcolor{magenta!60}  46.52  (15.64)    & 66.87  (16.28)  & 57.96  (16.86)                         & 61.45  (21.43)                         \\
        \textit{SWN}  & ---             & \cellcolor{magenta!60}  36.87   (9.62) & \cellcolor{magenta!60}  44.50  (12.74)    & 69.51  (18.72)  & \cellcolor{magenta!25}  58.11  (15.71) & \cellcolor{magenta!48}  52.29  (28.02) \\
        \textit{SWN}  & Lin.\ clip+norm & \cellcolor{magenta!60}  32.72   (9.97) & \cellcolor{magenta!60}  46.21  (13.48)    & 69.28  (18.61)  & 58.92  (16.36)                         & \cellcolor{magenta!49}  52.17  (27.81) \\
        \textit{SWN}  & Lin.\ simplex   & \cellcolor{magenta!60}  32.98  (10.53) & \cellcolor{magenta!60}  46.14  (13.53)    & 69.21  (18.58)  & 59.90  (17.11)                         & \cellcolor{magenta!49}  52.12  (27.75) \\
        \textit{SWN}  & Vec.\ scaling   & \cellcolor{magenta!60}  32.92  (11.77) & \cellcolor{magenta!60}  47.87  (14.91)    & 69.43  (18.82)  & 60.13  (17.82)                         & \cellcolor{magenta!50}  51.81  (28.17) \\
        \textit{NNLS} & ---             & \cellcolor{magenta!60}  34.25  (14.54) & \cellcolor{magenta!60}  44.74   (14.7)    & 66.07  (19.42)  & \cellcolor{magenta!40}  54.35  (16.26) & 66.03  (18.99)                         \\
        \textit{NNLS} & Lin.\ clip+norm & \cellcolor{magenta!60}  33.02  (15.38) & \cellcolor{magenta!60}  45.76  (15.65)    & 65.61  (19.46)  & \cellcolor{magenta!40}  54.34  (17.03) & 65.57   (19.0)                         \\
        \textit{NNLS} & Lin.\ simplex   & \cellcolor{magenta!60}  33.94  (16.54) & \cellcolor{magenta!60}  45.92  (15.97)    & 65.84   (19.4)  & \cellcolor{magenta!37}  55.06  (17.74) & 65.76  (18.89)                         \\
        \textit{NNLS} & Vec.\ scaling   & \cellcolor{magenta!60}  37.02  (20.33) & \cellcolor{magenta!60}  47.43   (18.4)    & 65.65  (20.53)  & 56.51  (19.81)                         & 66.21  (20.29)                         \\
        \textit{PSLS} & ---             & \cellcolor{magenta!60}  33.95  (14.01) & \cellcolor{magenta!60}  41.64  (14.44)    & 66.75  (18.68)  & \cellcolor{magenta!56}  50.37  (16.08) & 67.28  (17.77)                         \\
        \textit{PSLS} & Lin.\ clip+norm & \cellcolor{magenta!60}  32.45  (15.11) & \cellcolor{magenta!60}  42.52   (15.3)    & 66.21  (18.75)  & \cellcolor{magenta!56}  50.24  (16.78) & 66.77  (17.81)                         \\
        \textit{PSLS} & Lin.\ simplex   & \cellcolor{magenta!60}  32.97  (16.26) & \cellcolor{magenta!60}  42.79  (15.71)    & 66.50  (18.67)  & \cellcolor{magenta!53}  51.11  (17.67) & 67.02  (17.67)                         \\
        \textit{PSLS} & Vec.\ scaling   & \cellcolor{magenta!60}  42.09  (20.08) & \cellcolor{magenta!60}  44.97  (17.88)    & 65.71  (19.88)  & \cellcolor{magenta!46}  52.87  (19.46) & 66.80  (19.18)                         \\
        \bottomrule
      \end{tabular}
    }}
\end{table}

\begin{table}[H]
  \centering
  \caption{
    Comparison of TCS (mean and standard deviation in parentheses) against \textit{Celfie U25} for Data-Driven Soft-Labels with pooling and Data-Driven Soft-Labels without pooling using the \textbf{top 156 features}.
    Each row corresponds to a deconvolver-calibrator combination, evaluated with three classifiers under Data-Driven Soft-Label (\textit{Lookup Classifier},\textit{MethylBERT}, \textit{Dismir}), with and without pooling.
    Cells shaded in \textcolor{magenta}{magenta} indicate combinations whose mean TCS is significantly \emph{lower} than that of \textit{Celfie U25}, while cells shaded in \textcolor{teal}{teal} indicate combinations whose mean TCS is significantly \emph{higher} (Wilcoxon signed-rank test, $p \leq 0.05$).
    Color intensity scales with the magnitude of the difference.}
  \label{tab:tcs_vs_uxm_3}
  {\footnotesize
    \resizebox{\textwidth}{!}{%
      \begin{tabular}{llcccccc}
        \toprule
                      &                 & \multicolumn{3}{c}{Data-Driven Soft-Labels (with pooling)} & \multicolumn{3}{c}{Data-Driven Soft-Labels (without pooling)}                                                                                                                                                               \\
        \cmidrule(lr){3-5} \cmidrule(lr){6-8}
        Deconvolver   & Calibrator      & \textit{Lookup Classifier}                                 & \textit{MethylBERT}                                           & \textit{Dismir}                        & \textit{Lookup Classifier}          & \textit{MethylBERT}                    & \textit{Dismir}                     \\
        \midrule
        \textit{XGB}  & ---             & 62.71  (16.86)                                             & 63.26   (12.3)                                                & 66.87  (19.83)                         & 60.58  (13.41)                      & \cellcolor{magenta!17}  60.18  (15.53) & \cellcolor{teal!30}  72.10  (17.02) \\
        \textit{XGB}  & Lin.\ clip+norm & 61.58   (16.1)                                             & 64.65  (13.01)                                                & 66.89  (19.84)                         & 61.32  (14.83)                      & \cellcolor{magenta!14}  60.77  (16.38) & \cellcolor{teal!30}  72.13  (17.02) \\
        \textit{XGB}  & Lin.\ simplex   & 63.14  (17.14)                                             & 66.40  (13.73)                                                & 66.90  (19.84)                         & 64.85  (15.84)                      & 62.25  (17.15)                         & \cellcolor{teal!30}  72.15  (17.03) \\
        \textit{XGB}  & Vec.\ scaling   & 64.33  (18.08)                                             & 69.34  (14.88)                                                & 67.36  (20.17)                         & \cellcolor{teal!27}  71.29  (18.91) & 63.95  (19.25)                         & \cellcolor{teal!32}  72.54  (17.21) \\
        \textit{MLP}  & ---             & 66.20  (16.67)                                             & 67.70  (16.16)                                                & \cellcolor{magenta!15}  60.71  (18.14) & 66.42  (16.89)                      & 63.94  (16.32)                         & 69.01   (16.1)                      \\
        \textit{MLP}  & Lin.\ clip+norm & 66.72  (16.72)                                             & 68.63  (16.34)                                                & \cellcolor{magenta!14}  60.73  (18.12) & 67.55   (16.7)                      & 64.63  (16.75)                         & 69.09  (16.07)                      \\
        \textit{MLP}  & Lin.\ simplex   & 67.86  (17.25)                                             & \cellcolor{teal!21}  69.90  (16.72)                           & \cellcolor{magenta!14}  60.83  (18.18) & 68.43  (15.43)                      & 65.73  (17.48)                         & 69.25  (16.11)                      \\
        \textit{MLP}  & Vec.\ scaling   & 69.16  (18.07)                                             & \cellcolor{teal!19}  69.36  (16.68)                           & \cellcolor{magenta!12}  61.25  (18.22) & \cellcolor{teal!43}  75.44  (17.14) & 65.35  (17.52)                         & \cellcolor{teal!20}  69.61  (16.04) \\
        \textit{SWN}  & ---             & 66.78  (16.07)                                             & \cellcolor{teal!24}  70.71  (15.51)                           & 68.18  (20.44)                         & \cellcolor{teal!17}  68.80  (16.82) & 66.96   (15.7)                         & \cellcolor{teal!27}  71.33  (17.66) \\
        \textit{SWN}  & Lin.\ clip+norm & 67.76  (15.89)                                             & \cellcolor{teal!27}  71.30  (15.64)                           & 67.94  (20.32)                         & \cellcolor{teal!23}  70.31  (16.38) & 67.56  (16.06)                         & \cellcolor{teal!26}  71.07  (17.57) \\
        \textit{SWN}  & Lin.\ simplex   & 68.82  (16.34)                                             & \cellcolor{teal!32}  72.50  (16.11)                           & 67.90  (20.34)                         & \cellcolor{teal!26}  70.97  (14.62) & 68.57  (16.76)                         & \cellcolor{teal!26}  71.04  (17.58) \\
        \textit{SWN}  & Vec.\ scaling   & 70.22  (17.05)                                             & \cellcolor{teal!32}  72.47  (16.15)                           & 68.26  (20.56)                         & \cellcolor{teal!41}  74.96  (17.16) & 68.44   (16.8)                         & \cellcolor{teal!27}  71.27  (17.78) \\
        \textit{NNLS} & ---             & 63.54  (15.63)                                             & 66.49  (17.21)                                                & 64.38  (19.03)                         & 63.72  (14.81)                      & 65.33  (16.82)                         & 68.65  (17.26)                      \\
        \textit{NNLS} & Lin.\ clip+norm & 65.23  (15.88)                                             & 67.32  (17.09)                                                & 64.23  (18.98)                         & 67.60   (14.8)                      & 65.87  (17.04)                         & 68.44  (17.24)                      \\
        \textit{NNLS} & Lin.\ simplex   & 68.43  (17.29)                                             & \cellcolor{teal!20}  69.60  (17.49)                           & 64.42  (18.97)                         & \cellcolor{teal!29}  71.83  (14.85) & 67.54  (17.68)                         & \cellcolor{teal!17}  68.78  (17.23) \\
        \textit{NNLS} & Vec.\ scaling   & 71.34  (18.99)                                             & \cellcolor{teal!26}  71.19  (18.71)                           & 65.43  (19.79)                         & \cellcolor{teal!51}  77.28  (18.97) & 69.48  (19.03)                         & \cellcolor{teal!19}  69.26  (17.85) \\
        \textit{PSLS} & ---             & 63.81  (15.87)                                             & 67.58  (16.34)                                                & 65.57   (18.2)                         & 62.50  (15.49)                      & 66.17  (17.45)                         & 68.34  (17.04)                      \\
        \textit{PSLS} & Lin.\ clip+norm & 65.39   (15.9)                                             & 68.71   (16.0)                                                & 65.44   (18.1)                         & 66.58  (15.07)                      & 66.61  (17.49)                         & 68.15  (16.99)                      \\
        \textit{PSLS} & Lin.\ simplex   & 68.11  (16.88)                                             & \cellcolor{teal!27}  71.24  (16.19)                           & 65.64  (18.08)                         & \cellcolor{teal!21}  69.91  (14.65) & 68.09  (18.05)                         & 68.49  (16.97)                      \\
        \textit{PSLS} & Vec.\ scaling   & 71.04  (18.35)                                             & \cellcolor{teal!28}  71.63  (17.28)                           & 66.37  (18.62)                         & \cellcolor{teal!47}  76.39  (18.63) & 69.51  (19.05)                         & 68.71  (17.43)                      \\
        \bottomrule
      \end{tabular}
    }}
\end{table}

\begin{table}[H]
  \centering
  \caption{Comparison of TCS (mean and standard deviation in parentheses) against \textit{Celfie U25} for \textit{CancerDetector} with different priors.
    Cells shaded in \textcolor{magenta}{magenta} indicate combinations whose mean TCS is significantly \emph{lower} than that of \textit{Celfie U25},
    while cells shaded in \textcolor{teal}{teal} indicate combinations whose mean TCS is significantly \emph{higher} (Wilcoxon signed-rank test, $p \leq 0.05$).
    Color intensity scales with the magnitude of the difference.}
  \label{tab:tcs_vs_uxm_4}
  {\footnotesize
    \begin{tabular}{llcc}
      \toprule
      Deconvolver   & Calibrator      & Training Counts                     & Uniform                             \\
      \midrule
      \textit{XGB}  & ---             & \cellcolor{teal!27}  71.44  (17.56) & 69.90  (18.55)                      \\
      \textit{XGB}  & Lin.\ clip+norm & \cellcolor{teal!31}  72.31  (17.92) & \cellcolor{teal!25}  70.79  (18.99) \\
      \textit{XGB}  & Lin.\ simplex   & \cellcolor{teal!39}  74.34  (18.77) & \cellcolor{teal!32}  72.66  (19.94) \\
      \textit{XGB}  & Vec.\ scaling   & \cellcolor{teal!42}  75.13  (19.68) & \cellcolor{teal!35}  73.43  (21.25) \\
      \textit{MLP}  & ---             & \cellcolor{teal!19}  69.36  (15.79) & 65.16   (15.3)                      \\
      \textit{MLP}  & Lin.\ clip+norm & \cellcolor{teal!22}  70.18  (15.96) & 66.52  (15.94)                      \\
      \textit{MLP}  & Lin.\ simplex   & \cellcolor{teal!27}  71.22  (16.19) & 67.43  (16.56)                      \\
      \textit{MLP}  & Vec.\ scaling   & \cellcolor{teal!26}  71.03  (16.68) & 68.47  (17.02)                      \\
      \textit{SWN}  & ---             & 67.53  (16.02)                      & 65.67  (15.61)                      \\
      \textit{SWN}  & Lin.\ clip+norm & 68.92  (16.28)                      & 67.01  (15.98)                      \\
      \textit{SWN}  & Lin.\ simplex   & \cellcolor{teal!22}  70.11  (16.87) & 68.12  (16.56)                      \\
      \textit{SWN}  & Vec.\ scaling   & \cellcolor{teal!24}  70.56  (16.85) & 68.94  (16.87)                      \\
      \textit{NNLS} & ---             & 64.83  (16.98)                      & 63.73  (16.89)                      \\
      \textit{NNLS} & Lin.\ clip+norm & 65.72  (17.24)                      & 64.76  (17.13)                      \\
      \textit{NNLS} & Lin.\ simplex   & 68.04   (18.1)                      & 67.04   (18.0)                      \\
      \textit{NNLS} & Vec.\ scaling   & \cellcolor{teal!23}  70.46  (18.98) & \cellcolor{teal!21}  69.77  (19.35) \\
      \textit{PSLS} & ---             & 65.31  (18.63)                      & 63.63  (17.94)                      \\
      \textit{PSLS} & Lin.\ clip+norm & 65.51  (18.66)                      & 64.13   (17.9)                      \\
      \textit{PSLS} & Lin.\ simplex   & 67.04  (19.21)                      & 65.64  (18.53)                      \\
      \textit{PSLS} & Vec.\ scaling   & 67.56   (19.8)                      & 67.07  (19.13)                      \\
      \bottomrule
    \end{tabular}
  }
\end{table}

\subsection{Calibration as a standalone add-on to the baseline methods}
\label{sect:supp_calibration_standalone}
\suppressfloats[t]

We propose that calibration can be applied on the existing deconvolution
algorithm as a standalone method to influence derived cell-type fractions. To
asses effects of different calibration schemes, we have computed TCS for each
option evaluated in this paper. Additionaly, we looked at distribution of
allocated mass outside of the set of expected cell-types in accordance with
reference and inclusion treshold of 1\%. Finally, we meassured sparsity of each
solution by counting fraction of cell-types recieveing zero proportion as well
as tendency of each calibration method to avard a component with highest mass
in the baseline results with even higher mass. The corresponding pseudobulk
results for all four baselines, with and without our calibrators, are reported
in \Cref{tab:supp_uxm}.

\begin{figure}[H]
  \centering
  \includegraphics[width=\textwidth]{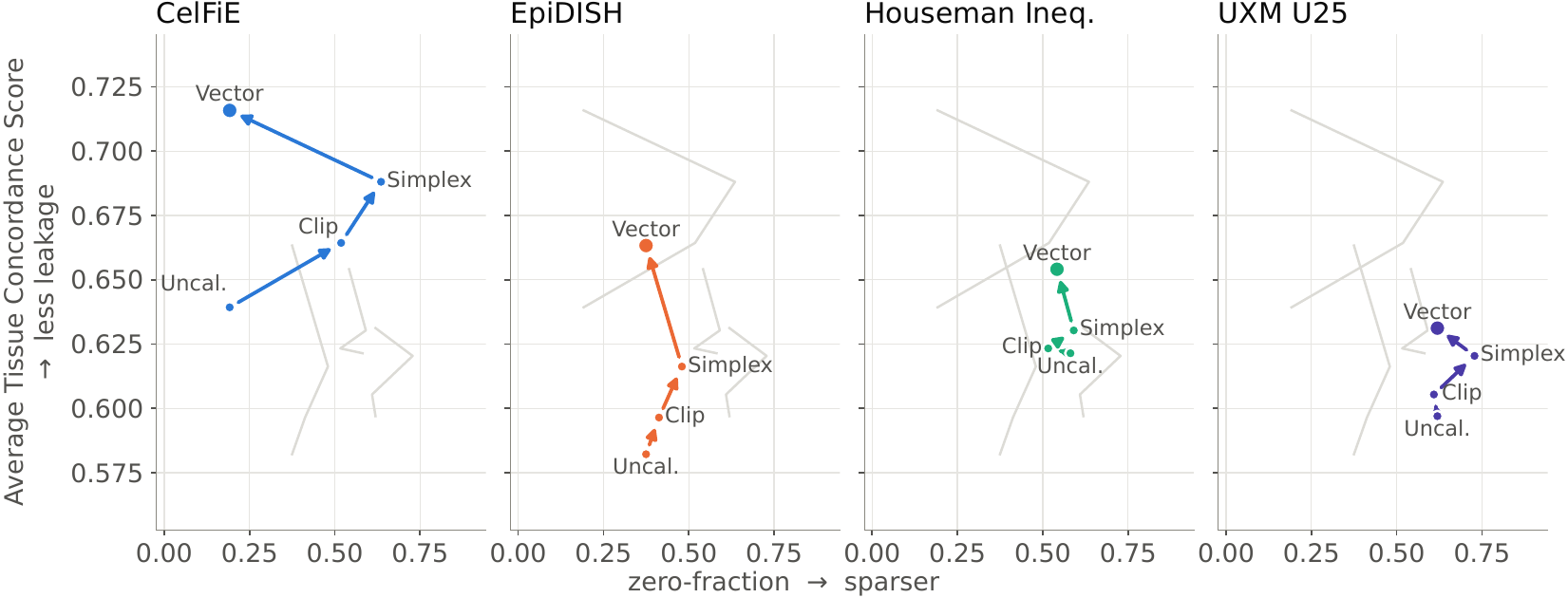}
  \caption{The sample-averaged TCS vs Sparsity trajectories for baseline deconvolvers and calibration methods.
    \emph{Leakage} used as a term to describe allocation of deconvoluted proportion to cell types that are not present in the constructed reference set of expected cell types per tissue.
    \emph{Sparsity} is computed as fraction of cell types with zero proportions.
  }
  \label{fig:calibration_bias_trajectory_sparsity}
\end{figure}

\begin{figure}[H]
  \centering
  \includegraphics[width=\textwidth]{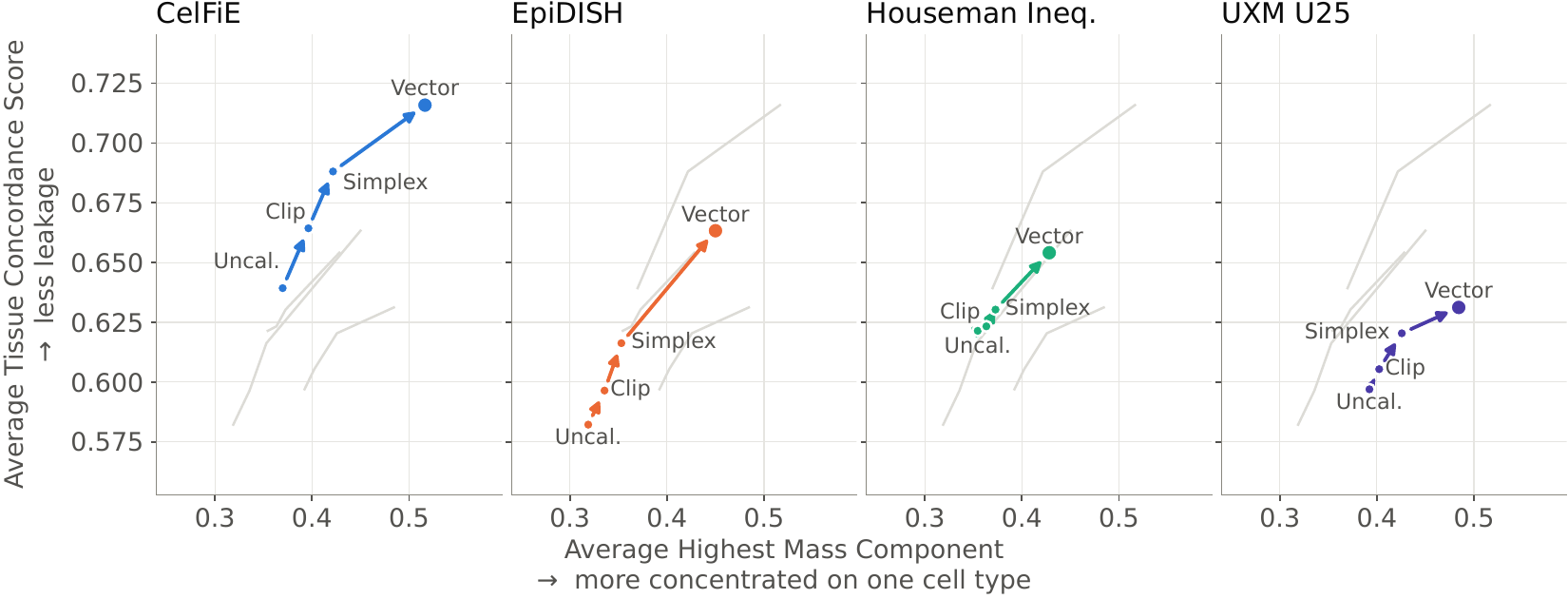}
  \caption{The sample-averaged TCS vs Proportion allocated to the cell type with highest mass trajectories for baseline deconvolvers and calibration methods.
    \emph{Leakage} used as a term to describe allocation of deconvoluted proportion to cell types that are not present in the constructed reference set of expected cell types per tissue.
    \emph{Average Highest Mass Component} is computed as average of maximal proportions across all deconvoluted samples.
  }
  \label{fig:calibration_bias_trajectory_high_mass}
\end{figure}

\begin{figure}[H]
  \centering
  \includegraphics[width=\textwidth]{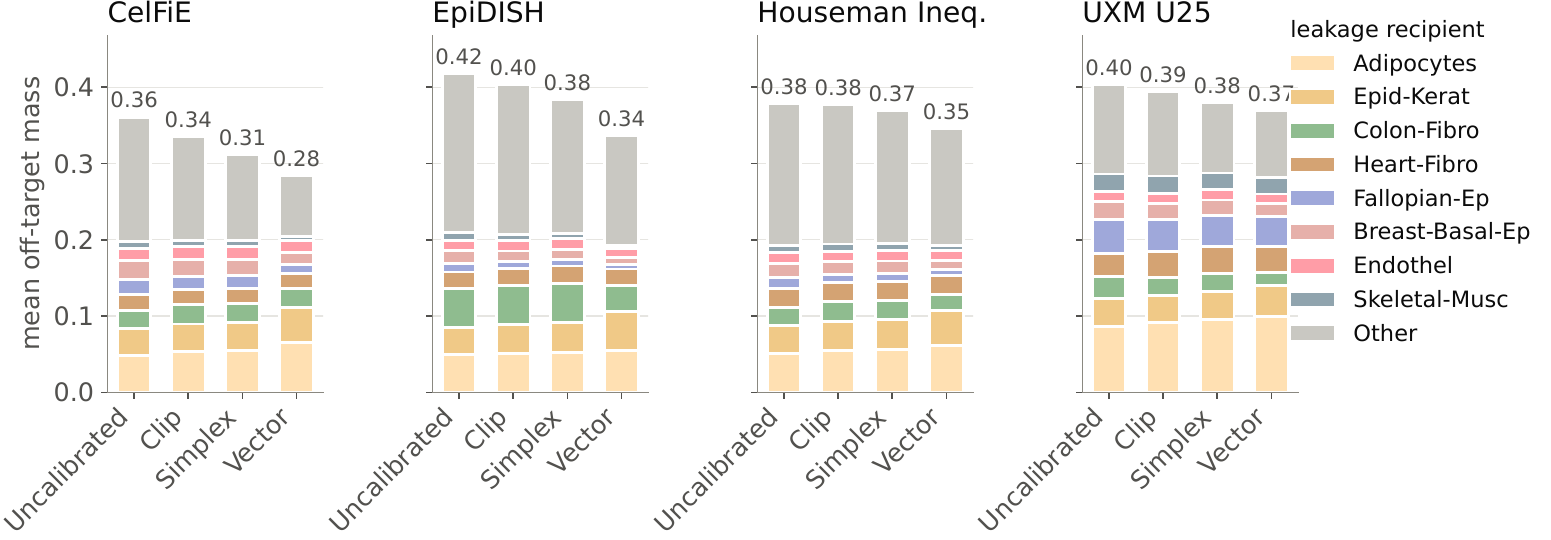}
  \caption{Distributions of the predicted cell types proportions that are not in the expected set of cell types (also refered here as \emph{leaked} proportions).
    The values in stacked bar plots are averaged per 16 tissues. The value on top of each bar is a total \emph{leakage}, also equivalent to $1-TCS$.
  }
  \label{fig:leakage_destinations}
\end{figure}

Based on conducted analysis, we make four observations:

\begin{enumerate}[leftmargin=1.5em,topsep=2pt,itemsep=2pt,parsep=0pt]
  \item Applying calibration consistently improves TCS with progressively bigger gains
        in the next order: \textit{Uncalibrated} -> \textit{Clip} -> \textit{Simplex}
        -> \textit{Vector}.
  \item All calibration methods are tend to allocate even higher mass to the most
        dominant component of uncalibrated input, with \textit{Vector} method resulting
        in the most dramatic effect
        (see~\Cref{fig:calibration_bias_trajectory_high_mass}). Furthermore, based on
        the gentler slope, \textit{Vector} calibration tend to benefit less for each
        point increase in the most dominant component than other methods, suggesting
        that more mass is redistributed within expected set of cell types for this
        method comparing to others.
  \item Applying \textit{Simplex} method consistently increases sparsity of the
        resulting deconvolved proportions, while \textit{Vector} method has almost no
        effect (see~\Cref{fig:calibration_bias_trajectory_sparsity}).
  \item Calibration tend to increase \emph{leaked} proprtions for some of the cell
        types and decrease for other across all baselines
        (see~\Cref{fig:leakage_destinations}). Consistenly growing fraction of
        \textit{adipocytes}may indicate reinforcing a bias present in the reference
        atlas; it may also suggest that \textit{adipocytes} are underdetected in the
        Tabula Sapines reference, hence are not making into expected cell types for the
        relevant tissues despite being truly present as all baseline methods detect.
\end{enumerate}

Overall, calibration seems to be usefull with a choice of calibration methods
depending on the application context. We see \textit{Simplex} method as a
suitable choice for scenarious where one is willing to sacrifice detection
limit for reducing number of falsely predicted cell types. \textit{Vector}
method may be used where expected cell type profile does not feature a single
dominant component which this method is likely to exploit and overestimate.

\subsection{Effect of the reference atlas choice on the \textit{UXM} baseline} \label{sect:supp_uxm_atlas_variants}
\suppressfloats[t]

\begin{figure}[H]
  \centering
  \begin{minipage}{0.48\textwidth}
    \centering
    \includegraphics[width=\textwidth]{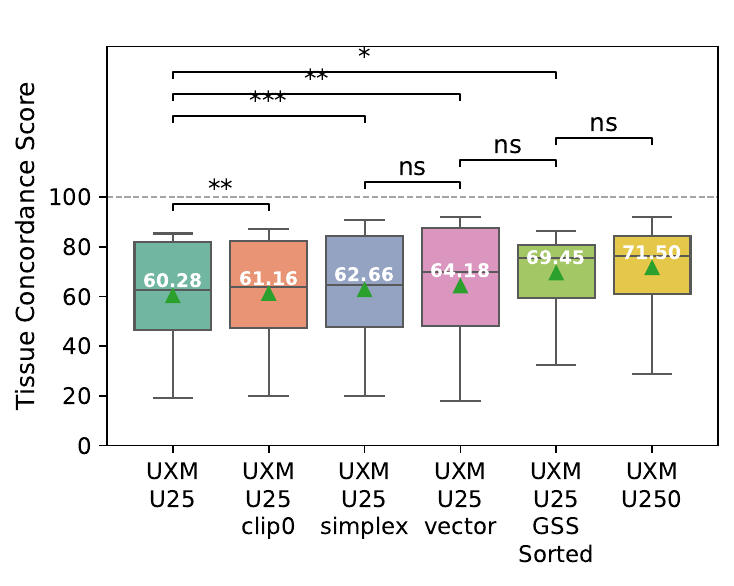}
  \end{minipage}
  \hfill
  \begin{minipage}{0.48\textwidth}
    \centering
    \includegraphics[width=\textwidth]{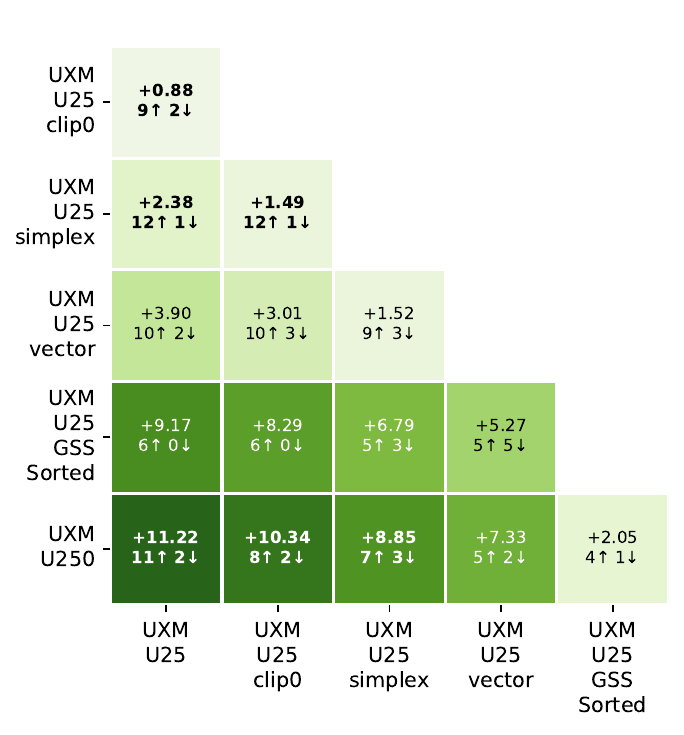}
  \end{minipage}

  \caption{Tissue-averaged TCS for the baseline \textit{UXM} setting (\textit{UXM U25}), various calibration schemes, modified atals optimized for proxy GSS (\textit{UXM U25 GSS sorted}, cf.~\Cref{sect:supp_atlas_analysis}) and extended $10\times$ atlas (\textit{UXM} U250).
    Left: Box-plot of sample-level TCS\@. The mean is reported in white text and the significance is derived from the pvalue of the Wilcoxon signed-rank test (\textit{scipy.stats.wilcoxon}).
    ``*'', ``**'' and ``***'' mean p-value $\leq$0.05, $\leq$0.01 and $\leq$0.001, respectively.
    Right: Heatmap of differences of tissue-averaged TCS\@. The values in parentheses indicate the number of tissues where significantly higher (↑) or lower (↓) TCS is observed for the row method relative to column method, using wilcoxon two sided test and significance threshold of 0.05.}
  \label{fig:uxm_rrbs_results_all}
\end{figure}

We analyzed how the \textit{UXM} baseline method fares against three variants
of itself: calibrated versions, a version using a same-sized reference but with
markers selected for higher specificity and a version using 250 marker regions
per cell type instead of the original 25. As \Cref{fig:uxm_rrbs_results_all}
reveals, linear calibration with clip-normalize, linear calibration with
simplex projection and vector scaling slightly improve upon the original
baseline scores (+0.84, p$\leq$0.001; +2.34, p$\leq$0.001 and +3.42,
p$\leq$0.001 respectively). The best results are achieved with the 10$x$ larger
atlas (+10.69, p$\leq$0.001); comparable gains can be obtained with a
same-sized subset of the U250 atlas optimized for marker specificity (+8.36,
p$\leq$0.001).

\newpage

\subsection{Analysis of \textit{Syto} sensitivity to the choice of reference atlas}

\label{sect:gr_sensitivity_analysis}

The major part of the results in this paper is built based on two versions of
the same reference atlas of cell-type-specific GR groups, one for each of hg19
and hg38 genomic assemblies (\Cref{sect:supp_region_sets}). By design, we refer
to GR groups in the broader sense of informative genomic regions where
underlying signal of interest has discriminative properties. When it comes to
methylome-based deconvolution, more specific terms like Differentially
Methylated Region (DMR) or Differentially Methylated Loci (DML) and their
synonyms are frequently used. Another frequent choice is to refer to
\textit{markers} as elements of the reference atlases, where each
\textit{marker} represents a characteristic localized methylation signature
that is expressed in the form of a metric, like, for example, a fraction of
unmethylated reads in a certain genomic region that is expected in different
cell types, as in the atlas by \citet{loyfer2023dna}.

As it has been highlighted in the literature, the \textit{marker} selection
method and the number of \textit{markers} employed in the reference affect
deconvolution performance, with the magnitude and direction of effect depending
on the particular combination of \textit{markers} and deconvolution method
\cite{sun2024systematic, deridder2024benchmarking, giuili2025benchmark}. In
\textit{Syto}, such \textit{markers} are not explicitly given at any of the
three stages (classification, deconvolution, calibration); however, we use the
genomic regions where the \textit{markers} are found as a criterion to filter
data at training and inference and as an additional context for classification.
Therefore, \textit{Syto} is expected to have its own sensitivity profile with
respect to reference atlas composition.

In \Cref{sect:supp_atlas_analysis} we demonstrated heterogeneity in the
specificity of the GR groups in \textit{U25.l4.hg38}~\cite{loyfer2023dna} and
proposed an alternative atlas of the same size, composed from regions
maximizing per-region specificity. We then showed in
\Cref{sect:supp_uxm_atlas_variants} that using the proposed alternative atlas
(here and after \textit{U25.l4.hg38.GSS}) elevates \textit{UXM} by +9.17 on the
TCS scale (look for \textit{UXM U25 GSS sorted} in
\Cref{fig:uxm_rrbs_results_all}). We hypothesised that \textit{Syto} will also
benefit from using a more specific atlas, as the pseudobulk experiment offered
some evidence that it will.

First, we have observed that per-individual cell-type deconvolution performance
is associated with a measure of the \textit{markers} specificity. Indeed,
Spearman correlation between the proxy GSS and per-cell-type deconvolution
$R^2$ was 0.61 for \textit{Syto} with hard labels, 0.62 for \textit{Syto} with
data-driven soft labels \footnote{The reported numbers for \textit{Syto} are
  for the \textit{Dismir} classifier with the \textit{SWN} deconvolver. Similar
  numbers were observed across other \textit{Syto} combinations.}, and 0.58 for
\textit{UXM}. Moreover, on the three cell types with the least specific
regions, per-cell-type $R^2$ was 0.49 / 0.76 / 0.84 for \textit{UXM} versus
0.83 / 0.88 / 0.95 for Syto w/ hard labels and 0.84 / 0.92 / 0.95 for Syto w/
DD soft labels, showing that \textit{Syto} performance degrades much less on
genomic regions with low specificity, and that soft labeling improves
performance on low-specificity regions.

To put our assumptions to the test, we have conducted a focused experiment with
an alternative atlas, limiting our choice of classifier to \textit{Dismir} and
evaluation to \textit{OOD RRBS} context. We opted out of using a
\textit{pseudobulk} experiment as the attained performance was already
saturated. We gave priority to \textit{Dismir} owing to its robustness to the
labeling scheme choice (see \Cref{tab:rrbs_config_counts}).

Our main findings are:
\begin{itemize}
  \item Switching to \textit{U25.l4.hg38.GSS} atlas improves results across different
        labeling schemes. The improvement magnitude differs, but direction is
        concordant with \textit{UXM} (\Cref{tab:regions_sensitivity_tcs_gains}).
  \item The TCS gains are reinforced by calibration
        (\Cref{tab:regions_sensitivity_calibration_gains}).
  \item The \textit{UXM} improvement partially stems from eliminating Adipocytes and
        Fallopian Epithelial bias. As can be seen from
        \Cref{tab:regions_sensitivity_tcs_gains}, the change in TCS score is not as
        dramatic for \textit{Syto} as for \textit{UXM}.
        \Cref{tab:regions_sensitivity_recipient_delta} reveals that this is not only
        due to the higher baseline for \textit{Syto}, but also because \textit{Syto}
        doesn't have the same bias to begin with, which takes us back to the argument
        that \textit{Syto} is more robust to the choice of less specific regions.
\end{itemize}

Additionally, we observed higher Pearson correlation between \textit{UXM} and
\textit{Syto} predicted proportions under the \textit{U25.l4.hg38.GSS} atlas
($0.894$--$0.907$ vs $0.775$--$0.809$ under \textit{U25.l4.hg38} for the
matched SWN deconvolver; $0.837-0.910$ vs $0.778-0.829$ across tested 15
Labeling-Deconvolver combinations). We see two possible explanations: either
both solutions are closer to the ground truth, or GSS-atlas increased
specificity prompts both methods towards a common solution more aggressively.

Our interpretation is subject to the limitations of TCS, discussed in
\Cref{sect:tcs_analysis}. One notable example arises under the
\textit{U25.l4.hg38.GSS} atlas: \textit{UXM} dramatically increases the
fraction of Smooth Muscle allocated to uterus and ovary tissues, and
\textit{Syto} does the same for uterus and shows a lesser increase for ovary,
albeit from an already elevated base ( see \Cref{tab:smooth-musc-atlas}). Since
Smooth Muscle is moderately present in both tissues, overall allocation
unfairly boosts TCS\@. Among \textit{Syto} varinats, \textit{Syto} under hard
labels "gains" the most from switching to \textit{U25.l4.hg38.GSS} atlas, while
\textit{Syto} under soft labels with pooling seems to be more robust than
\textit{UXM}. If one were to remove uterus and ovary tissues from analysis, the
TCS gain from switching to \textit{U25.l4.hg38.GSS} calculated on the remaining
14 tissues will be considerably lower for \textit{UXM} (+5.2 vs +9.2 on the
full tissue set). As for \textit{Syto}, recomputing TCS gain on 14 remaining
tissues effectively nullifies gains for uncalibrated variants, reducing maximal
gain for calibrated ones (e.g +3.3 vs +6.0 for vector calibrated \textit{SWN}
under soft labeling scheme).

To conclude, switching to \textit{U25.l4.hg38.GSS} atlas seems to help
\textit{UXM} disproportionately more than \textit{Syto}. While it helps
\textit{UXM} to reduce the performance gap, \textit{Syto} maintains an
advantage. The results reinforce the idea that \textit{Syto} is more forgiving
of the selection of less specific \textit{markers} than \textit{UXM}.

\begin{table}[tbp]

  \centering
  \caption{Comparison of TCS gains for \textit{UXM} and \textbf{uncalibrated} result for each labeling scheme using \textit{Dismir} classifier and \textit{SWN} deconvolver.
    TCS values are averaged within each of the 16 mapped tissues first, then over the 16 tissues.
    Confidence intervals are computed by drawing 16 tissues with replacement and averaging $\Delta$ 4000 times.
    \textit{UXM} improves in 11 of the 16 tissues;
    \textit{Syto} trades bigger gains in roughly half of the tissues over smaller losses in the rest.
    \textit{U25.l4.hg38} atlas is denoted as U25. \textit{U25.l4.hg38.GSS} atlas is denoted as GSS.}
  \label{tab:regions_sensitivity_tcs_gains}

  {\footnotesize
    \begin{tabular}{lccccc}
      \toprule
      Method                        & TCS U25 & TCS GSS & $\Delta$       & 95\% CI           & Tissues improved \\
      \midrule
      \textit{UXM}                  & 60.28   & 69.45   & \textbf{+9.17} & [+2.98, +16.01]   & 11/16            \\
      \textit{Syto} --- Hard        & 71.23   & 74.48   & +3.25          & [$-$2.63, +10.76] & 8/16             \\
      \textit{Syto} --- Soft        & 68.18   & 71.83   & +3.66          & [$-$1.62, +9.96]  & 9/16             \\
      \textit{Syto} --- Soft-NoPool & 71.33   & 72.80   & +1.47          & [$-$3.80, +7.25]  & 7/16             \\
      \bottomrule
    \end{tabular}
  }
\end{table}

\begin{table}[tbp]
  \centering
  \caption{Relative TCS gains per calibration scheme for \textit{Dismir} classifier + \textit{SWN} deconvolver.
    TCS values are averaged within each of the 16 mapped tissues first, then over the 16 tissues.
    The reported number is computed as $\overline{TCS}_{U25.l4.hg38.GSS}-\overline{TCS}_{U25.l4.hg38}$.}
  \label{tab:regions_sensitivity_calibration_gains}

  {\footnotesize
    \begin{tabular}{lcccc}
      \toprule
      Labeling Scheme         & Uncalibrated & Clip  & Simplex & Vector         \\
      \midrule
      Hard Labels             & +3.25        & +4.00 & +2.33   & \textbf{+4.52} \\
      Soft Labels w/ pooling  & +3.66        & +4.83 & +4.94   & \textbf{+5.96} \\
      Soft Labels w/o pooling & +1.47        & +2.50 & +2.15   & \textbf{+4.00} \\
      \bottomrule
    \end{tabular}
  }
\end{table}

\begin{table}[tbp]

  \centering

  \caption{Per cell type $\Delta$ of the off-target mass fraction, ranked by pooled $|\Delta|$. Negative sign means that the method using the \textit{U25.l4.hg38.GSS} atlas leaks less mass to that cell type than the method operating on \textit{U25.l4.hg38}.}

  \label{tab:regions_sensitivity_recipient_delta}
  {\footnotesize
    \begin{tabular}{lccccl}
      \toprule
      Recipient         & UXM               & Hard     & Soft     & Soft-NoPool & Direction        \\
      \midrule
      Adipocytes        & \textbf{$-$0.049} & $-$0.014 & $-$0.026 & +0.012      & mixed            \\
      Colon-Fibro       & \textbf{+0.030}   & $-$0.014 & $-$0.009 & $-$0.025    & mixed            \\
      Breast-Basal-Ep   & $-$0.021          & $-$0.013 & $-$0.013 & $-$0.013    & \textbf{all $-$} \\
      Fallopian-Ep      & \textbf{$-$0.050} & +0.002   & 0.000    & +0.003      & mixed            \\
      Pancreas-Delta    & +0.015            & +0.011   & +0.012   & +0.014      & \textbf{all $+$} \\
      Smooth-Musc       & $-$0.002          & +0.011   & +0.008   & +0.015      & mixed            \\
      Pancreas-Duct     & $-$0.016          & $-$0.004 & $-$0.004 & $-$0.003    & \textbf{all $-$} \\
      Lung-Ep-Bron      & $-$0.001          & $-$0.006 & $-$0.013 & $-$0.005    & \textbf{all $-$} \\
      Liver-Hep         & +0.003            & +0.004   & +0.007   & +0.008      & \textbf{all $+$} \\
      Kidney-Ep         & +0.004            & +0.004   & +0.005   & +0.006      & \textbf{all $+$} \\
      Oligodend         & $-$0.004          & $-$0.004 & $-$0.005 & $-$0.003    & \textbf{all $-$} \\
      Thyroid-Ep        & $-$0.006          & $-$0.001 & +0.007   & $-$0.002    & mixed            \\
      Breast-Luminal-Ep & +0.005            & $-$0.002 & $-$0.003 & $-$0.003    & mixed            \\
      Endothel          & +0.001            & +0.003   & +0.001   & $-$0.006    & mixed            \\
      \bottomrule
    \end{tabular}
  }
\end{table}

\begin{table}[t]
  \centering
  \caption{Predicted Smooth Muscle mass fraction in the two tissues where switching the GSS atlas
    produces the biggest gain. \textit{Syto} results are for \textit{Dismir} classifier + \textit{SWN} deconvolver. Values are averaged over the samples of each
    tissue; the reference fraction in the header is the Tabula Sapiens
    composition mapped onto the Loyfer vocabulary. Smooth Muscle is on the
    expected list of both tissues.
    \textit{U25.l4.hg38} atlas is denoted as U25. \textit{U25.l4.hg38.GSS} atlas is denoted as GSS.}
  \label{tab:smooth-musc-atlas}
  \begin{tabular}{lcccccc}
    \toprule
                         & \multicolumn{3}{c}{Ovary (ref.\ $0.162$, $n{=}16$)}
                         & \multicolumn{3}{c}{Uterus (ref.\ $0.205$, $n{=}12$)}                                               \\
    \cmidrule(lr){2-4} \cmidrule(lr){5-7}
    Method               & U25                                                  & GSS   & $\Delta$ & U25   & GSS   & $\Delta$ \\
    \midrule
    UXM                  & 0.103                                                & 0.565 & $+0.462$ & 0.266 & 0.683 & $+0.417$ \\
    Syto --- Hard        & 0.658                                                & 0.785 & $+0.127$ & 0.212 & 0.800 & $+0.587$ \\
    Syto --- Soft        & 0.508                                                & 0.587 & $+0.078$ & 0.129 & 0.555 & $+0.426$ \\
    Syto --- Soft-NoPool & 0.398                                                & 0.627 & $+0.230$ & 0.157 & 0.562 & $+0.404$ \\
    \bottomrule

  \end{tabular}

\end{table}
\clearpage

\newpage
\subsection{Detailed analysis of TCS failure modes and recovered proportions in the OOD RRBS data}
\label{sect:tcs_analysis}

\providecommand{\tcslegend}[1]{{\setlength{\fboxsep}{0pt}%
      \fcolorbox{black!25}{#1}{\rule{0pt}{1.3ex}\rule{2.2ex}{0pt}}}}

TCS is a conceptually simple metric with several limitations. As mentioned in
the main section of this paper, TCS cannot distinguish within expected
cell-type errors and thus represents an imperfect upper bound on deconvolution
accuracy. It cannot be used to quantify the method's performance absolutely.
Nonetheless, we claim that, combined with specific evidence from additional
analysis, it can serve as a measure of relative performance. In this
supplementary section, we list TCS failure modes and highlight them in
experimental data. We then show that, despite several prominent examples of
failure modes being observed in our data, their effects are mostly uniform
across all compared methods, including \emph{Syto} and baseline variants. For a
few cases where failures specific to concrete variants are suspected but cannot
be confidently called, we lay out our arguments on why their relative effects
are still marginal.

\paragraph{TCS failure modes.} TCS can both overestimate and underestimate algorithms' performance. Among
reasons imaginable, three failure modes stand out as likely to account for most
of the occurring errors:

\begin{enumerate}[leftmargin=1.5em,topsep=2pt,itemsep=2pt,parsep=0pt]
  \item \textbf{Spurious expected cell type / sensible overpredicted cell type} — a cell type with negligible proportion in the tissue is listed as expected, rewarding a method that overpredicts it. A variation of the same failure mode is when the initial proportion is moderate, so the cell is expected but its proportion is significantly overestimated.
  \item \textbf{Missing expected cell type} — an abundant cell type may be under-detected by the scRNA-seq behind Tabula Sapiens, so a method that correctly recovers it is penalised.
  \item \textbf{Ubiquitous cell type} — a cell type is truly present in a majority of tissues, rewarding any method biased toward overpredicting it.
\end{enumerate}

One can control for (1) by setting a threshold for a minimum fraction that cell
type must contribute to the tissue composition in order to be added to the set
of expected tissue cell types. A higher threshold reduces the risk of (1) but
simultaneously increases the risk of penalty for (2). Notably, \textbf{Missing
  expected cell type} can occur even when the threshold for a minimum fraction is
set to zero, purely because the reference method completely fails to detect
truly present cell types. One cannot control for (3) by adjusting the metric;
if the underlying method is biased towards ubiquitous cell types, it will gain
more points on TCS scale. Endothelial cells are a good example; being present
in most of the tissues (14 out of 16 in our OOD experiment), any method that
allocates 100\% to these cell types will have a very high average TCS (exactly
87.5 for our data). Luckily, one can easily check if that is the case by
examining actual proportions.

\begin{figure}[tbp]
  \centering
  \includegraphics[width=0.92\textwidth]{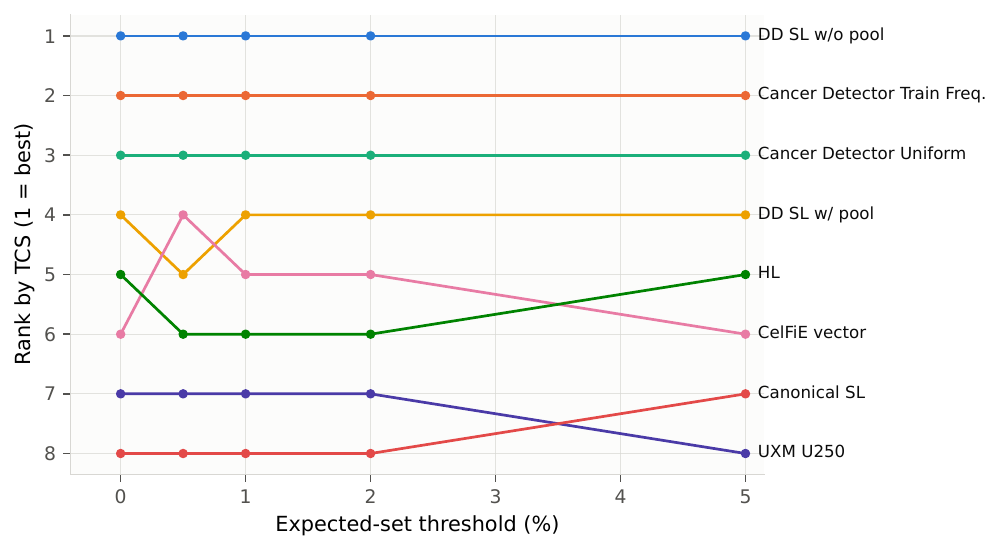}
  \caption{Dependency between TCS rankings of eight top-performing methods and cell type proportion inclusion threshold. Tested values of threshold are: 0\%, 0.5\%, 1\%, 2\%, and 5\%}
  \label{fig:cell_inclusion_threshold_stability}
\end{figure}

In our work, we set a threshold for a minimum fraction that a cell type must
contribute to the tissue composition at 1\%. We also checked how setting
different thresholds influences ranking across best methods and found that
varying the threshold between 0.5\% and 5\% does not affect the main claims of
the paper (see \Cref{fig:cell_inclusion_threshold_stability}).

\begin{figure}[tbp]
  \centering
  \includegraphics[width=0.92\textwidth]{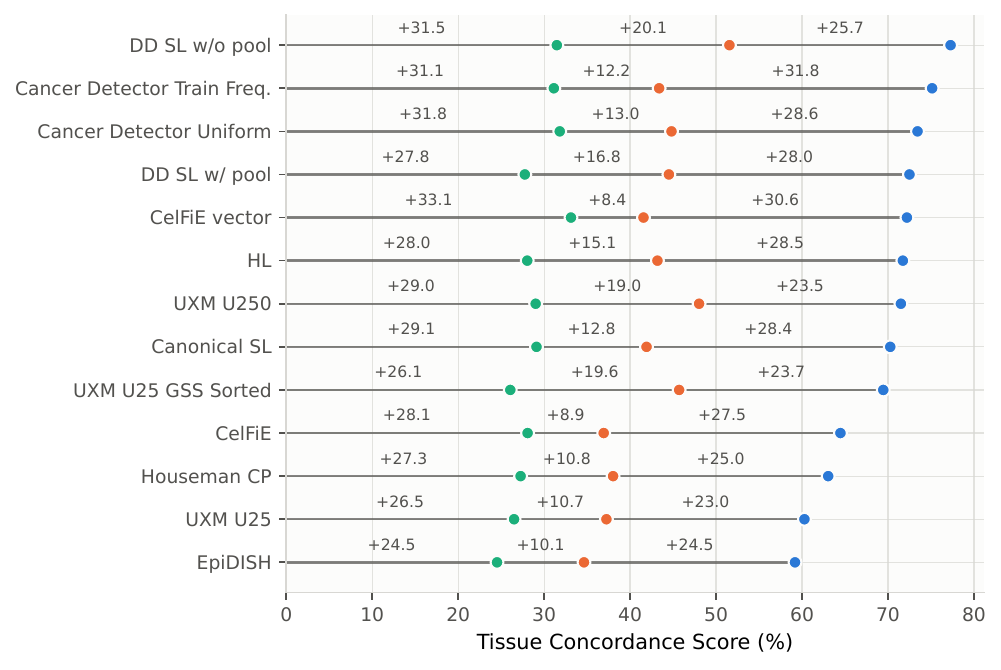}
  \caption{TCS for all baselines, best \emph{Syto} variants,  \emph{UXM} using alternative atlases, and best calibrated variant of the baseline as a function of cell types included in the set of expected cell types for computing TCS\@.
    Threshold for a minimum fraction is set to 1\%. blue = all expected cell types; orange = excluding the 3 types expected in $\geq14/16$ tissues (Blood-Mono+Macro, Blood-T, Endothel); green = excluding the 7 expected in $\geq8/16$ (those three plus Blood-B, Blood-Granul, Dermal-Fibro, Smooth-Musc)
  }
  \label{fig:ubiquity_ladder}
\end{figure}

\begin{figure}[tbp]
  \centering
  \includegraphics[width=0.92\textwidth]{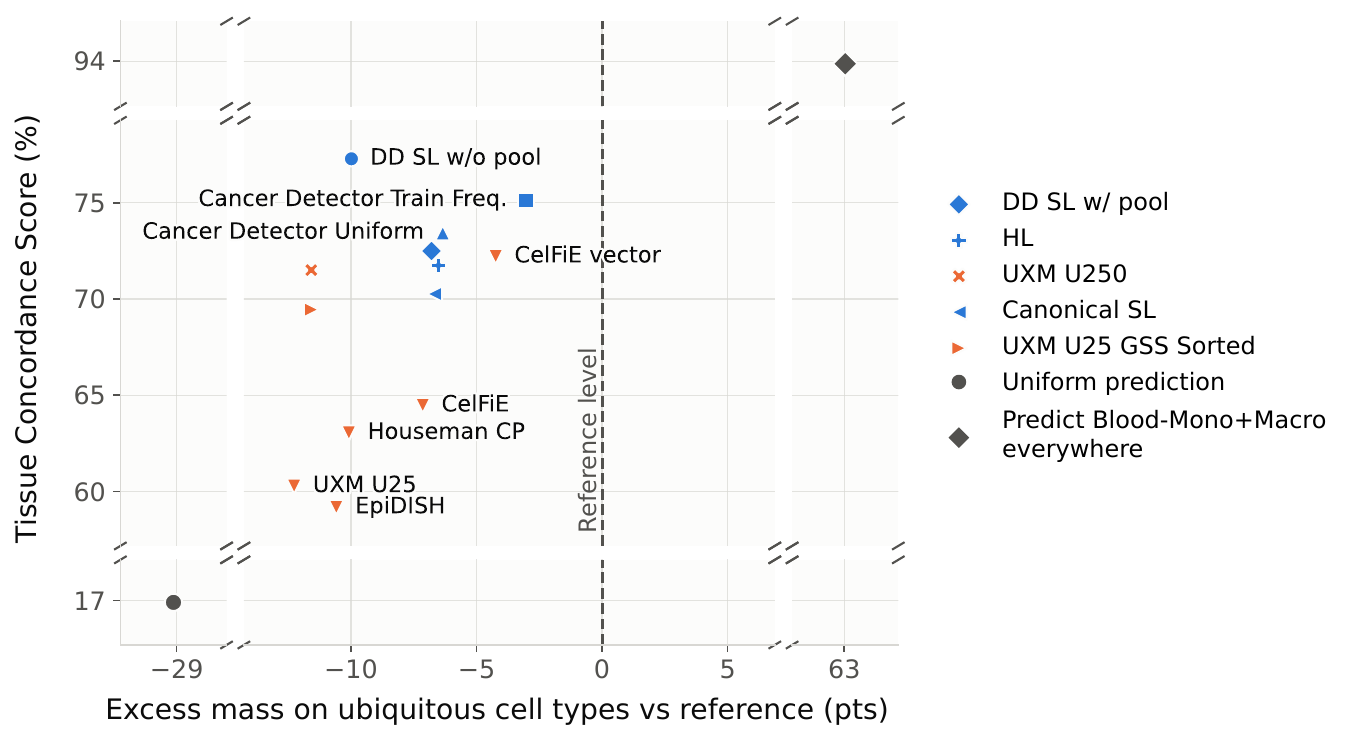}
  \caption{Coordinate plot projecting selected attained TCS vs mass allocated to the cell types present in at least 14/16 tissues. "uniform over all cell types" - TCS for a method that predicts cell type proportions uniformly.
    Reference levele line corresponds to the average total fraction of Top-3 ubiquitous cell types in Tabula Sapiens reference. Baselines and their variants are colored in orange. \emph{Syto} methods are colored in blue.
    Uniform prediction results and results steming from predicting Blood-Mono+Macro everywhere are in grey and added for perspective.}
  \label{fig:ubiquity_scatter}
\end{figure}

Next, we examined how much mass methods allocated to ubiquitous cell types, in
absolute terms and in relation to the Tabula Sapiens reference (see
\Cref{fig:ubiquity_scatter}). Additionally, we modeled the ranking of the
methods based on the full set of expected cell types for a given threshold of
1\% and two reduced sets: one including only cell types that present no more
than in 13 out of 16 tissues and a more conservative set of cell types that are
present in at most 7 tissues (see \Cref{fig:ubiquity_ladder}).

Baselines and \emph{Syto} methods consistently allocate less mass to the three
most ubiquitous cell types than specified in the reference. Three baselines,
\emph{UXM U25}, \emph{EpiDISH} and \emph{Houseman CP} as well as \emph{UXM
  U250} and \emph{UXM U250 GSS sorted} and best \emph{Syto} variant all allocate
roughly between $-10$ to $-12$ points to ubiquitous cell types comparing to the
reference. The best baseline method, \emph{CelFiE}, and the majority of
remaining best \emph{Syto} variants across labeling schemes allocate roughly
between $-7$ and $-6$ points to ubiquitous cell types compared to the
reference. The closest to the reference method, \emph{Cancer Detector} with
train frequencies prior, allocates $\approx -3$ pp. relative to Tabula Sapiens.

Examining TCS of reduced cell type sets allows us to see how much of the TCS
gain is spread across the most unique, moderately and ubiquitously present cell
types. Excluding the three most ubiquitous cell types from TCS computation
repositions \emph{UXM U250} and \emph{UXM U250 GSS sorted} as second and third
best solutions after \emph{DD SL w/o pool·Lookup·NNLS·vector}. It also makes
\emph{Uniform} prior a better choice for \emph{CancerDetector} and
\emph{EpiDISH} and \emph{Houseman CP} better baselines than \emph{CelFiE}.
Focusing on cell types present in no more than seven tissues perturbs ranking,
positioning \textit{vector}-calibrated \textit{CelFiE} as first and ranking
uncalibrated \textit{CelFiE} above all other baselines again and slightly above
\textit{DD SL w/ pool·MethylBERT·SWN·simplex} and
\textit{HL·Dismir·NNLS·vector}, still below other \textit{Syto} variants.

Without access to the ground truth, the failure mode (3) cannot be fully
excluded. Nonetheless, based on collected evidence, it is possibly influencing
the ranking within \emph{Syto} methods and is much less likely to affect
conclusions for \emph{Syto} advantage over examined baselines.

\begin{figure}[tbp]
  \centering
  \includegraphics[width=\textwidth]{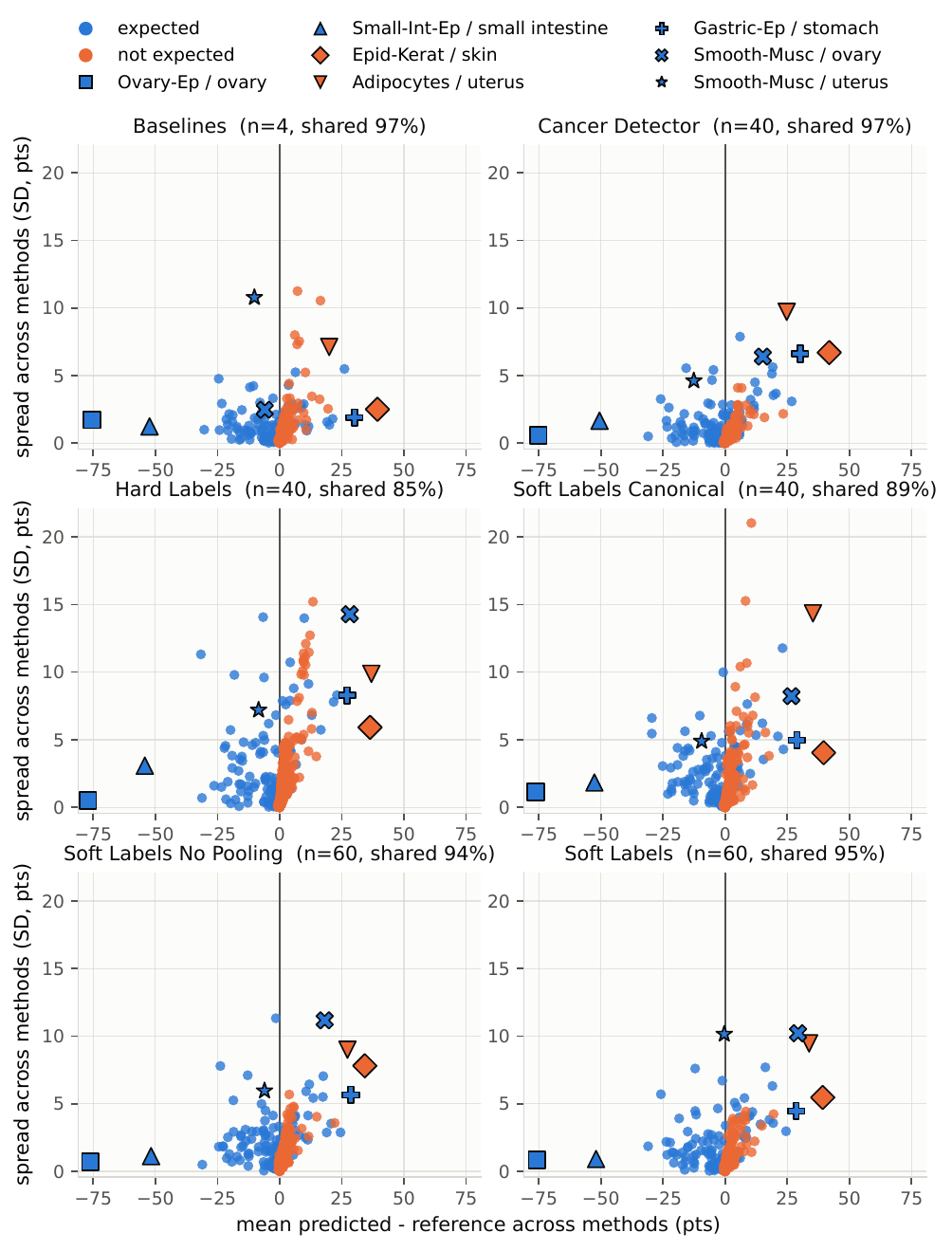}
  \caption{Re-distribution of mass compared to Tabula Sapiens reference per method group. Each dot represented the aggregated prediction per each combination of cell type and tissue.
    The values in parentheses indicate: n - number of methods in the group; shared - percentage of total variation relative to the reference that is shared across methods within one group.
    For Hard Labels Lookup classifier-based results were excluded. The feature selection scheme is \textit{Top 156} across \emph{Syto} methods.}
  \label{fig:shared_vs_specific_grid}
\end{figure}

To screen for failure modes (1) and (2), for each method we selected the cell
type with the highest predicted proportion per tissue and compared it with the
expected dominant cell type in the reference. We then computed within-group
agreement for each group of methods by counting how many methods allocated the
highest proportion to the same cell type. The results are summarized in
\Cref{tab:dominance}. Finally, we looked at each individual predicted cell type
proportion in comparison to the Tabula Sapiens reference and computed method
group-level means and standard deviations for each tissue cell-type
combination. We then projected them onto 2D scatter to see what biases are
shared and differ between different method groups (see
\Cref{fig:shared_vs_specific_grid}).

\begin{table}[tbp]
  \centering
  \caption{Dominant cell type called per tissue by each labelling scheme, against the Tabula Sapiens reference.
    Each entry is the cell type that most configurations in the group call dominant, with the number that agree.
    Shading: \tcslegend{teal!15}~matches Tabula Sapiens;
    \tcslegend{blue!12}~$\ast$~differs, but the called type is in the expected set, so TCS is unaffected;
    \tcslegend{magenta!15}~$\dagger$~differs and lies outside the expected set.}
  \label{tab:dominance}
  \footnotesize
  \setlength{\tabcolsep}{4pt}
  \resizebox{\textwidth}{!}{%
    \begin{tabular}{llllllll}
      \toprule
      Tissue          & Tabula Sapiens   & Baseline                                           & Cancer Detector                                     & HL                                                    & Canonical SL                                         & DD SL w/o pool                                      & DD SL w/ pool                                       \\
      \midrule
      Adipose Tissue  & Dermal-Fibro     & \cellcolor{blue!12}$\ast$\,Adipocytes (4/4)        & \cellcolor{blue!12}$\ast$\,Endothel (37/40)         & \cellcolor{blue!12}$\ast$\,Adipocytes (24/40)         & \cellcolor{blue!12}$\ast$\,Adipocytes (28/40)        & \cellcolor{blue!12}$\ast$\,Endothel (47/60)         & \cellcolor{blue!12}$\ast$\,Adipocytes (40/60)       \\
      Bladder         & Bladder-Ep       & \cellcolor{teal!15}Bladder-Ep (3/4)                & \cellcolor{teal!15}Bladder-Ep (40/40)               & \cellcolor{teal!15}Bladder-Ep (24/40)                 & \cellcolor{teal!15}Bladder-Ep (20/40)                & \cellcolor{teal!15}Bladder-Ep (51/60)               & \cellcolor{teal!15}Bladder-Ep (46/60)               \\
      Blood Vessel    & Dermal-Fibro     & \cellcolor{magenta!15}$\dagger$\,Heart-Fibro (2/4) & \cellcolor{blue!12}$\ast$\,Endothel (28/40)         & \cellcolor{blue!12}$\ast$\,Blood-Mono+Macro (19/40)   & \cellcolor{magenta!15}$\dagger$\,Heart-Fibro (20/40) & \cellcolor{blue!12}$\ast$\,Endothel (30/60)         & \cellcolor{blue!12}$\ast$\,Blood-Mono+Macro (24/60) \\
      Heart           & Endothel         & \cellcolor{teal!15}Endothel (4/4)                  & \cellcolor{teal!15}Endothel (40/40)                 & \cellcolor{teal!15}Endothel (21/40)                   & \cellcolor{teal!15}Endothel (34/40)                  & \cellcolor{teal!15}Endothel (60/60)                 & \cellcolor{teal!15}Endothel (59/60)                 \\
      Kidney          & Kidney-Ep        & \cellcolor{teal!15}Kidney-Ep (4/4)                 & \cellcolor{teal!15}Kidney-Ep (40/40)                & \cellcolor{teal!15}Kidney-Ep (36/40)                  & \cellcolor{teal!15}Kidney-Ep (40/40)                 & \cellcolor{teal!15}Kidney-Ep (60/60)                & \cellcolor{teal!15}Kidney-Ep (60/60)                \\
      Liver           & Liver-Hep        & \cellcolor{teal!15}Liver-Hep (4/4)                 & \cellcolor{teal!15}Liver-Hep (39/40)                & \cellcolor{teal!15}Liver-Hep (40/40)                  & \cellcolor{teal!15}Liver-Hep (40/40)                 & \cellcolor{teal!15}Liver-Hep (60/60)                & \cellcolor{teal!15}Liver-Hep (57/60)                \\
      Lung            & Blood-Mono+Macro & \cellcolor{blue!12}$\ast$\,Endothel (4/4)          & \cellcolor{blue!12}$\ast$\,Endothel (39/40)         & \cellcolor{magenta!15}$\dagger$\,Fallopian-Ep (13/40) & \cellcolor{blue!12}$\ast$\,Endothel (24/40)          & \cellcolor{blue!12}$\ast$\,Endothel (56/60)         & \cellcolor{blue!12}$\ast$\,Endothel (42/60)         \\
      Ovary           & Ovary-Ep         & \cellcolor{blue!12}$\ast$\,Smooth-Musc (2/4)       & \cellcolor{blue!12}$\ast$\,Smooth-Musc (33/40)      & \cellcolor{blue!12}$\ast$\,Smooth-Musc (39/40)        & \cellcolor{blue!12}$\ast$\,Smooth-Musc (40/40)       & \cellcolor{blue!12}$\ast$\,Smooth-Musc (49/60)      & \cellcolor{blue!12}$\ast$\,Smooth-Musc (60/60)      \\
      Pancreas        & Pancreas-Acinar  & \cellcolor{teal!15}Pancreas-Acinar (4/4)           & \cellcolor{teal!15}Pancreas-Acinar (40/40)          & \cellcolor{teal!15}Pancreas-Acinar (38/40)            & \cellcolor{teal!15}Pancreas-Acinar (40/40)           & \cellcolor{teal!15}Pancreas-Acinar (60/60)          & \cellcolor{teal!15}Pancreas-Acinar (60/60)          \\
      Prostate        & Prostate-Ep      & \cellcolor{teal!15}Prostate-Ep (4/4)               & \cellcolor{teal!15}Prostate-Ep (40/40)              & \cellcolor{teal!15}Prostate-Ep (40/40)                & \cellcolor{teal!15}Prostate-Ep (40/40)               & \cellcolor{teal!15}Prostate-Ep (60/60)              & \cellcolor{teal!15}Prostate-Ep (60/60)              \\
      Salivary Gland  & Head-Neck-Ep     & \cellcolor{teal!15}Head-Neck-Ep (4/4)              & \cellcolor{blue!12}$\ast$\,Endothel (16/40)         & \cellcolor{magenta!15}$\dagger$\,Fallopian-Ep (13/40) & \cellcolor{magenta!15}$\dagger$\,Adipocytes (16/40)  & \cellcolor{blue!12}$\ast$\,Endothel (39/60)         & \cellcolor{teal!15}Head-Neck-Ep (35/60)             \\
      Skin            & Blood-T          & \cellcolor{magenta!15}$\dagger$\,Epid-Kerat (4/4)  & \cellcolor{magenta!15}$\dagger$\,Epid-Kerat (40/40) & \cellcolor{magenta!15}$\dagger$\,Epid-Kerat (39/40)   & \cellcolor{magenta!15}$\dagger$\,Epid-Kerat (40/40)  & \cellcolor{magenta!15}$\dagger$\,Epid-Kerat (60/60) & \cellcolor{magenta!15}$\dagger$\,Epid-Kerat (60/60) \\
      Small Intestine & Small-Int-Ep     & \cellcolor{blue!12}$\ast$\,Blood-B (4/4)           & \cellcolor{blue!12}$\ast$\,Blood-B (40/40)          & \cellcolor{blue!12}$\ast$\,Blood-B (33/40)            & \cellcolor{blue!12}$\ast$\,Blood-B (36/40)           & \cellcolor{blue!12}$\ast$\,Blood-B (60/60)          & \cellcolor{blue!12}$\ast$\,Blood-B (60/60)          \\
      Spleen          & Blood-B          & \cellcolor{blue!12}$\ast$\,Endothel (3/4)          & \cellcolor{teal!15}Blood-B (25/40)                  & \cellcolor{blue!12}$\ast$\,Endothel (16/40)           & \cellcolor{teal!15}Blood-B (24/40)                   & \cellcolor{teal!15}Blood-B (31/60)                  & \cellcolor{teal!15}Blood-B (41/60)                  \\
      Stomach         & Blood-Mono+Macro & \cellcolor{blue!12}$\ast$\,Gastric-Ep (4/4)        & \cellcolor{blue!12}$\ast$\,Gastric-Ep (40/40)       & \cellcolor{blue!12}$\ast$\,Gastric-Ep (36/40)         & \cellcolor{blue!12}$\ast$\,Gastric-Ep (36/40)        & \cellcolor{blue!12}$\ast$\,Gastric-Ep (60/60)       & \cellcolor{blue!12}$\ast$\,Gastric-Ep (60/60)       \\
      Uterus          & Dermal-Fibro     & \cellcolor{blue!12}$\ast$\,Endothel (2/4)          & \cellcolor{magenta!15}$\dagger$\,Adipocytes (31/40) & \cellcolor{magenta!15}$\dagger$\,Adipocytes (39/40)   & \cellcolor{magenta!15}$\dagger$\,Adipocytes (32/40)  & \cellcolor{magenta!15}$\dagger$\,Adipocytes (47/60) & \cellcolor{magenta!15}$\dagger$\,Adipocytes (46/60) \\
      \bottomrule
    \end{tabular}
  }
\end{table}

First and foremost, a clear case of failure mode (2) emerged in \textit{skin}
tissue recovered proportions. \textit{Epidermis-Keratinocytes} constitute less
than 1\% of skin, slightly above 0.5\% according to the Tabula Sapiens
reference, hence never making it into the set of expected cell types for this
tissue. All examined methods, baselines and \emph{Syto} variants alike, predict
an average fraction of \textit{Epidermis-Keratinocytes} in skin tissue samples
$\approx 40\%$ with variance depending on the method group. A double-digit
fraction of \textit{Epidermis-Keratinocytes} in skin is expected. Therefore,
across our benchmark, all methods have approximately the same penalty for this
tissue, lowering the global TCS score but not affecting final ranking. Another
interesting case is a fraction of \textit{Adipocytes} in the uterus. The spread
between method groups is higher than in the previous example; however,
\textit{Adipocytes} receives roughly 20\% to 40\% on average in each method
group, highlighting another possible case of failure mode (2).

Moving on to the examples where TCS correctly penalized all methods in the
benchmark, we can highlight failure to detect \textit{Ovary Epithelial} in
ovarian tissue and \textit{Small Intestinal Epithelial} in small intestine
samples. All methods recover near-zero fractions for these cases, well below
the reference, with minimal variance within method groups. However, \emph{Syto}
allocated to \textit{Smooth Muscle} in ovarian tissue roughly from 18\% to 30\%
on average per method group, with proportions as high as 75.9\% for \emph{DD SL
  w/ pool·Lookup·NNLS·vector} and reference value of 16.9\%. In contrast,
baseline methods put slightly less than the reference, between 8.6\% and 14.4\%
depending on the method. Therefore, \emph{Syto} gains a higher TCS score for
overpredicting moderately present cell types, which can be considered as
evidence for the failure mode (1). Notably, \emph{UXM U250} and \emph{UXM U25
  GSS Sorted} allocate 46.5\% and 56.5\% to \textit{Smooth Muscle} in this
example, partially explaining their TCS superiority over \emph{U25}-based
baselines and also suggesting that the true fraction of \textit{Smooth Muscle}
might be higher than in the reference. Another potential example of
overpredicting moderately present cell type is in recovered proportions of
\textit{Gastric Epithelial} in stomach samples, which is roughly 30\% higher on
average across all baselines and \emph{Syto} variants. The exerted influence of
this potential overprediction is, however, uniform.

Given that ovarian tissue favours \emph{Syto} highly, we also recomputed TCS
ranks using a set of 15 tissues, ovary excluded (\Cref{tab:tcs-no-ovary}).
While it tightens the gap between the baselines and \emph{Syto}, it does not
change the relative positioning of \emph{Syto} labeling schemes, nor does it
change the fact that \emph{Syto} achieves higher \textit{TCS}. Overall, Data
Driven Soft Labels with pooling agree with the reference most of the time with
respect to the most dominant cell type (for 8 tissues out of 16). The Hard
Labeling scheme produces the most discordant results (agrees on 6 tissues out
of 16), and the rest of the methods, including baselines, are in between (agree
on 7 tissues out of 16).

\begin{table}[tbp]
  \centering
  \caption{Best \emph{Syto} configuration per labelling scheme and the uncalibrated baselines, with and without ovary.
    Tissue Concordance Score (\%), averaged within each tissue and then across tissues.
    Subscript 16 is the full set; subscript 15 excludes ovary.
    Configurations are those selected as best per labelling scheme on all 16 tissues, so the rows match the figures.
    Ranks run over all 378 (360 variants of \emph{Syto} + 4 uncalibrated baselines, 12 calibrated baselines variants, \emph{UXM U25 GSS Sorted} and \emph{UXM U250}) methods; $\Delta$Rank $>0$ means the method ranks higher once ovary is removed.}
  \label{tab:tcs-no-ovary}
  \footnotesize
  \setlength{\tabcolsep}{5pt}
  \begin{tabular}{lrrrrr}
    \toprule
    Method                                                                                           & TCS$_{16}$ & TCS$_{15}$ & Rank$_{16}$ & Rank$_{15}$ & $\Delta$Rank \\
    \midrule
    \multicolumn{6}{l}{\itshape Best configuration per labelling scheme}                                                                                                  \\
    \midrule
    DD SL w/o pool\textperiodcentered{}Lookup\textperiodcentered{}NNLS\textperiodcentered{}vector    & 77.28      & 76.59      & 1           & 1           & 0            \\
    Cancer Detector Train Freq.\textperiodcentered{}XGB\textperiodcentered{}vector                   & 75.13      & 75.43      & 4           & 4           & 0            \\
    Cancer Detector Uniform\textperiodcentered{}XGB\textperiodcentered{}vector                       & 73.43      & 73.70      & 7           & 8           & $-1$         \\
    DD SL w/ pool\textperiodcentered{}MethylBERT\textperiodcentered{}SWN\textperiodcentered{}simplex & 72.50      & 72.42      & 10          & 16          & $-6$         \\
    HL\textperiodcentered{}Dismir\textperiodcentered{}NNLS\textperiodcentered{}vector                & 71.73      & 71.21      & 18          & 35          & $-17$        \\
    Canonical SL\textperiodcentered{}Dismir\textperiodcentered{}NNLS\textperiodcentered{}vector      & 70.26      & 70.12      & 52          & 58          & $-6$         \\
    \midrule
    \multicolumn{6}{l}{\itshape Uncalibrated baselines, CelFiE vector, and UXM variants using alternative atlases}                                                        \\
    \midrule
    CelFiE vector                                                                                    & 72.20      & 74.51      & 13          & 7           & $+6$         \\
    UXM U250                                                                                         & 71.50      & 72.83      & 20          & 11          & $+9$         \\
    UXM U25 GSS Sorted                                                                               & 69.45      & 70.19      & 72          & 54          & $+18$        \\
    CelFiE                                                                                           & 64.47      & 67.02      & 210         & 150         & $+60$        \\
    Houseman CP                                                                                      & 63.04      & 65.33      & 231         & 210         & $+21$        \\
    UXM U25                                                                                          & 60.28      & 63.01      & 261         & 237         & $+24$        \\
    EpiDISH                                                                                          & 59.18      & 61.86      & 268         & 247         & $+21$        \\
    \bottomrule
  \end{tabular}
\end{table}

\clearpage

\flushbottom
\makeatletter
\renewcommand{\topfraction}{0.85}
\renewcommand{\bottomfraction}{0.4}
\renewcommand{\textfraction}{0.1}
\renewcommand{\floatpagefraction}{0.7}
\setlength{\@fptop}{0\p@ \@plus 1fil}
\setlength{\@fpsep}{8\p@ \@plus 2fil}
\setlength{\@fpbot}{0\p@ \@plus 1fil}
\makeatother

\newpage
\section{Experiments compute resources} \label{sec:supp_compute}

The majority of experiments were conducted on a single workstation and are
fully reproducible on equivalent hardware. To accelerate batch jobs, pseudobulk
generation and deconvolver fitting were additionally offloaded to a
high-performance computing cluster; however, the primary workstation is
sufficient to run the entire pipeline end-to-end.

\paragraph{Primary workstation.}
An HP Z2 Tower G9 equipped with an Intel Core i9-12900K (16 cores, 24 threads,
up to 5.2\,GHz), 64\,GB DDR5 RAM, an NVIDIA RTX A4500 GPU (20\,GB VRAM) and a
2\,TB Samsung NVMe SSD, running Ubuntu 22.04 LTS. This machine was used for all
stages of the pipeline: data preparation (restoring reads from \texttt{.pat}
files and constructing train/validation/test splits), training and inference of
read-level classifiers (\textit{Dismir}, \textit{MethylBERT}, \textit{Lookup
    Classifier}), pseudobulk mixture generation, deconvolver fitting and inference,
and linear calibrator fitting and inference.

\paragraph{Secondary workstation.}
A Dell Precision 7540 laptop equipped with an Intel Core i7-9850H (6 cores, 12
threads, up to 4.6\,GHz), 32\,GB DDR4 RAM, an NVIDIA Quadro T1000 Mobile GPU
(4\,GB VRAM) and a 512\,GB Western Digital NVMe SSD, running Linux Mint 22.3
(Ubuntu 24.04 base). This machine was used in parallel to fit and evaluate the
\textit{CancerDetector} and \textit{Lookup Classifier} baselines, along with
their corresponding calibrators.

\paragraph{HPC cluster.}
Batch jobs for pseudobulk generation and deconvolver fitting were offloaded to
an HPC cluster.

All details about the compute resources and wall-clock time for each stage of
the pipeline are provided in the supplementary spreadsheet named `Compute
Resources.ods`.

\newpage
\section{Licensing}
\label{sect:licensing}

\paragraph{Data.} This work does not produce new data assets. As stated in the main text, all the
data used is publicly available, and is downloadable from Gene Expression
Omnibus (GEO), a public functional genomics data repository, using accession
numbers: GSE186458 for WGBS data used for training and pseudobulk experiment;
GSE233417 RRBS data for OOD experiment. There is no explicit Licensing
Agreement under which this data is distributed, however \textit{"Unless
      otherwise stated, documents and files on NCBI Web servers may be freely
      downloaded and reproduced."} (see the official disclaimer
\href{https://www.ncbi.nlm.nih.gov/geo/info/disclaimer.html}{https://www.ncbi.nlm.nih.gov/geo/info/disclaimer.html}).
The reference cell type counts from Tabula Sapiens consortium were accessed via
its \href{https://chanzuckerberg.github.io/cellxgene-census/}{public API} and
the access is compliant with data release policy
(\href{https://tabula-sapiens.sf.czbiohub.org/whereisthedata}{https://tabula-sapiens.sf.czbiohub.org/whereisthedata}).

\paragraph{Code.} In this work we have reused or repurposed the next software:
\begin{itemize}
      \item \textit{MethylBERT}: Distibuted under MIT License\\
            (https://github.com/CompEpigen/methylbert/blob/main/LICENSE).
      \item \textit{UXM}: Distributed under its own License\\
            (https://github.com/nloyfer/\textit{UXM}\_deconv/blob/main/LICENSE.m), allows non-commercial, research use.
      \item wgbs\_tools: Distributed under its own License\\
            (https://github.com/nloyfer/wgbs\_tools/blob/master/LICENSE.md).
      \item \textit{CelFiE}: Distibuted under GNU Affero General Public License v3.0 \\
            (https://github.com/christacaggiano/celfie/blob/master/LICENSE.txt).
      \item \textit{EpiDISH}: Distibuted under GNU General Public License v2.0 (GPL-2)\\
            (https://bioconductor.posit.co/packages/3.20/bioc/html/EpiDISH.html).
      \item HuggingFace Transformers: Distributed under Apache License 2.0\\
            (https://github.com/huggingface/transformers/blob/main/LICENSE).
      \item PyTorch: Distributed under its own License\\
            (https://github.com/pytorch/pytorch/blob/main/LICENSE).
\end{itemize}

We implemented \textit{Dismir} and \textit{CancerDetector} from scratch and did
not reuse any existing code for these implementations.

\newpage
\section*{NeurIPS Paper Checklist}

\begin{enumerate}
      \item {\bf Claims}
      \item[] Question: Do the main claims made in the abstract and introduction accurately reflect the paper's contributions and scope?
      \item[] Answer: \answerYes{} %
      \item[] Justification: The paper's main contributions are enumerated at the end of the introduction. %
      \item[] Guidelines:
            \begin{itemize}
                  \item The answer \answerNA{} means that the abstract and introduction do not include
                        the claims made in the paper.
                  \item The abstract and/or introduction should clearly state the claims made,
                        including the contributions made in the paper and important assumptions and
                        limitations. A \answerNo{} or \answerNA{} answer to this question will not be
                        perceived well by the reviewers.
                  \item The claims made should match theoretical and experimental results, and reflect
                        how much the results can be expected to generalize to other settings.
                  \item It is fine to include aspirational goals as motivation as long as it is clear
                        that these goals are not attained by the paper.
            \end{itemize}

      \item {\bf Limitations}
      \item[] Question: Does the paper discuss the limitations of the work performed by the authors?
      \item[] Answer: \answerYes{} %
      \item[] Justification: See the limitation paragraph in the main paper (\Cref{sect:limitations}).
      \item[] Guidelines:
            \begin{itemize}
                  \item The answer \answerNA{} means that the paper has no limitation while the answer
                        \answerNo{} means that the paper has limitations, but those are not discussed
                        in the paper.
                  \item The authors are encouraged to create a separate ``Limitations'' section in
                        their paper.
                  \item The paper should point out any strong assumptions and how robust the results
                        are to violations of these assumptions (e.g., independence assumptions,
                        noiseless settings, model well-specification, asymptotic approximations only
                        holding locally). The authors should reflect on how these assumptions might be
                        violated in practice and what the implications would be.
                  \item The authors should reflect on the scope of the claims made, e.g., if the
                        approach was only tested on a few datasets or with a few runs. In general,
                        empirical results often depend on implicit assumptions, which should be
                        articulated.
                  \item The authors should reflect on the factors that influence the performance of the
                        approach. For example, a facial recognition algorithm may perform poorly when
                        image resolution is low or images are taken in low lighting. Or a
                        speech-to-text system might not be used reliably to provide closed captions for
                        online lectures because it fails to handle technical jargon.
                  \item The authors should discuss the computational efficiency of the proposed
                        algorithms and how they scale with dataset size.
                  \item If applicable, the authors should discuss possible limitations of their
                        approach to address problems of privacy and fairness.
                  \item While the authors might fear that complete honesty about limitations might be
                        used by reviewers as grounds for rejection, a worse outcome might be that
                        reviewers discover limitations that aren't acknowledged in the paper. The
                        authors should use their best judgment and recognize that individual actions in
                        favor of transparency play an important role in developing norms that preserve
                        the integrity of the community. Reviewers will be specifically instructed to
                        not penalize honesty concerning limitations.
            \end{itemize}

      \item {\bf Theory assumptions and proofs}
      \item[] Question: For each theoretical result, does the paper provide the full set of assumptions and a complete (and correct) proof?
      \item[] Answer: \answerYes{} %
      \item[] Justification: The only theoretical result of the paper is the fact that normalized-count soft labels are the unique maximum likelihood estimates of the true conditional categorical distribution (cf.~\Cref{sect:labeling}), as a direct application of a textbook result. The reference is provided alongside the result~\cite{murphy2012machine}.
      \item[] Guidelines:
            \begin{itemize}
                  \item The answer \answerNA{} means that the paper does not include theoretical
                        results.
                  \item All the theorems, formulas, and proofs in the paper should be numbered and
                        cross-referenced.
                  \item All assumptions should be clearly stated or referenced in the statement of any
                        theorems.
                  \item The proofs can either appear in the main paper or the supplemental material,
                        but if they appear in the supplemental material, the authors are encouraged to
                        provide a short proof sketch to provide intuition.
                  \item Inversely, any informal proof provided in the core of the paper should be
                        complemented by formal proofs provided in appendix or supplemental material.
                  \item Theorems and Lemmas that the proof relies upon should be properly referenced.
            \end{itemize}

      \item {\bf Experimental result reproducibility}
      \item[] Question: Does the paper fully disclose all the information needed to reproduce the main experimental results of the paper to the extent that it affects the main claims and/or conclusions of the paper (regardless of whether the code and data are provided or not)?
      \item[] Answer: \answerYes{} %
      \item[] Justification: We list all relevant details of the experimental setups, models architecture and models' hyperparameters in \Cref{sect:supp_details_experimental_setups}, \Cref{sect:supp_data_preparation} and \Cref{sect:supp_methods}.
      \item[] Guidelines:
            \begin{itemize}
                  \item The answer \answerNA{} means that the paper does not include experiments.
                  \item If the paper includes experiments, a \answerNo{} answer to this question will
                        not be perceived well by the reviewers: Making the paper reproducible is
                        important, regardless of whether the code and data are provided or not.
                  \item If the contribution is a dataset and\slash or model, the authors should
                        describe the steps taken to make their results reproducible or verifiable.
                  \item Depending on the contribution, reproducibility can be accomplished in various
                        ways. For example, if the contribution is a novel architecture, describing the
                        architecture fully might suffice, or if the contribution is a specific model
                        and empirical evaluation, it may be necessary to either make it possible for
                        others to replicate the model with the same dataset, or provide access to the
                        model. In general. releasing code and data is often one good way to accomplish
                        this, but reproducibility can also be provided via detailed instructions for
                        how to replicate the results, access to a hosted model (e.g., in the case of a
                        large language model), releasing of a model checkpoint, or other means that are
                        appropriate to the research performed.
                  \item While NeurIPS does not require releasing code, the conference does require all
                        submissions to provide some reasonable avenue for reproducibility, which may
                        depend on the nature of the contribution. For example
                        \begin{enumerate}
                              \item If the contribution is primarily a new algorithm, the paper should make it
                                    clear how to reproduce that algorithm.
                              \item If the contribution is primarily a new model architecture, the paper should
                                    describe the architecture clearly and fully.
                              \item If the contribution is a new model (e.g., a large language model), then there
                                    should either be a way to access this model for reproducing the results or a
                                    way to reproduce the model (e.g., with an open-source dataset or instructions
                                    for how to construct the dataset).
                              \item We recognize that reproducibility may be tricky in some cases, in which case
                                    authors are welcome to describe the particular way they provide for
                                    reproducibility. In the case of closed-source models, it may be that access to
                                    the model is limited in some way (e.g., to registered users), but it should be
                                    possible for other researchers to have some path to reproducing or verifying
                                    the results.
                        \end{enumerate}
            \end{itemize}

      \item {\bf Open access to data and code}
      \item[] Question: Does the paper provide open access to the data and code, with sufficient instructions to faithfully reproduce the main experimental results, as described in supplemental material?
      \item[] Answer: \answerYes{} %
      \item[] Justification:
            The code, models and processed data are released open-source as stated in \Cref{sect:supp_code_and_data}. 
            As mentionned in \Cref{sect:experiments setup}, the source data used in both experiments is publicly available at GEO under accession numbers GSE186458 and GSE233417 respectively.
      \item[] Guidelines:
            \begin{itemize}
                  \item The answer \answerNA{} means that paper does not include experiments requiring
                        code.
                  \item Please see the NeurIPS code and data submission guidelines
                        (\url{https://neurips.cc/public/guides/CodeSubmissionPolicy}) for more details.
                  \item While we encourage the release of code and data, we understand that this might
                        not be possible, so \answerNo{} is an acceptable answer. Papers cannot be
                        rejected simply for not including code, unless this is central to the
                        contribution (e.g., for a new open-source benchmark).
                  \item The instructions should contain the exact command and environment needed to run
                        to reproduce the results. See the NeurIPS code and data submission guidelines
                        (\url{https://neurips.cc/public/guides/CodeSubmissionPolicy}) for more details.
                  \item The authors should provide instructions on data access and preparation,
                        including how to access the raw data, preprocessed data, intermediate data, and
                        generated data, etc.
                  \item The authors should provide scripts to reproduce all experimental results for
                        the new proposed method and baselines. If only a subset of experiments are
                        reproducible, they should state which ones are omitted from the script and why.
                  \item At submission time, to preserve anonymity, the authors should release
                        anonymized versions (if applicable).
                  \item Providing as much information as possible in supplemental material (appended to
                        the paper) is recommended, but including URLs to data and code is permitted.
            \end{itemize}

      \item {\bf Experimental setting/details}
      \item[] Question: Does the paper specify all the training and test details (e.g., data splits, hyperparameters, how they were chosen, type of optimizer) necessary to understand the results?
      \item[] Answer: \answerYes{} %
      \item[] Justification: The construction of the data splits is described in \Cref{sect:supp_split_construction}, the pseudobulk generation is described in \Cref{supp:pseudobulk_sampling}, the predictions aggregation is described in \Cref{supp:predictions_aggregation}, the deconvolution feature selection is described in \Cref{supp:dec_features_selection}, the deconvolution metrics are described in \Cref{sect:supp_deconv_metrics} and the combination of possible elements of \textit{Syto} is described in \Cref{sect:supp_combinations_elements_Syto}.
            The model hyperparameters are listed in \Cref{sec:supp_deconv_params} and \Cref{sec:supp_classifier_training}.
      \item[] Guidelines:
            \begin{itemize}
                  \item The answer \answerNA{} means that the paper does not include experiments.
                  \item The experimental setting should be presented in the core of the paper to a
                        level of detail that is necessary to appreciate the results and make sense of
                        them.
                  \item The full details can be provided either with the code, in appendix, or as
                        supplemental material.
            \end{itemize}

      \item {\bf Experiment statistical significance}
      \item[] Question: Does the paper report error bars suitably and correctly defined or other appropriate information about the statistical significance of the experiments?
      \item[] Answer: \answerYes{} %
      \item[] Justification: For the pseudobulk experiments, we report the 95\% confidence interval over the followed metrics.
            The interval is computed using 10,000 bootstrap samples and the Bias Corrected accelerated bootstrap method~\cite{efron1987bca}.
            For the OOD experiment, we report significance using Wilcoxon signed-rank test using p-value threshold of 0.05.
      \item[] Guidelines:
            \begin{itemize}
                  \item The answer \answerNA{} means that the paper does not include experiments.
                  \item The authors should answer \answerYes{} if the results are accompanied by error
                        bars, confidence intervals, or statistical significance tests, at least for the
                        experiments that support the main claims of the paper.
                  \item The factors of variability that the error bars are capturing should be clearly
                        stated (for example, train/test split, initialization, random drawing of some
                        parameter, or overall run with given experimental conditions).
                  \item The method for calculating the error bars should be explained (closed form
                        formula, call to a library function, bootstrap, etc.)
                  \item The assumptions made should be given (e.g., Normally distributed errors).
                  \item It should be clear whether the error bar is the standard deviation or the
                        standard error of the mean.
                  \item It is OK to report 1-sigma error bars, but one should state it. The authors
                        should preferably report a 2-sigma error bar than state that they have a 96\%
                        CI, if the hypothesis of Normality of errors is not verified.
                  \item For asymmetric distributions, the authors should be careful not to show in
                        tables or figures symmetric error bars that would yield results that are out of
                        range (e.g., negative error rates).
                  \item If error bars are reported in tables or plots, the authors should explain in
                        the text how they were calculated and reference the corresponding figures or
                        tables in the text.
            \end{itemize}

      \item {\bf Experiments compute resources}
      \item[] Question: For each experiment, does the paper provide sufficient information on the computer resources (type of compute workers, memory, time of execution) needed to reproduce the experiments?
      \item[] Answer: \answerYes{} %
      \item[] Justification: We provide the necessary compute resources and wall-clock times for all stages of the experiments in the supplementary spreadsheet named `Compute Resources.ods`, as well as give a brief summary in \Cref{sec:supp_compute}.
      \item[] Guidelines:
            \begin{itemize}
                  \item The answer \answerNA{} means that the paper does not include experiments.
                  \item The paper should indicate the type of compute workers CPU or GPU, internal
                        cluster, or cloud provider, including relevant memory and storage.
                  \item The paper should provide the amount of compute required for each of the
                        individual experimental runs as well as estimate the total compute.
                  \item The paper should disclose whether the full research project required more
                        compute than the experiments reported in the paper (e.g., preliminary or failed
                        experiments that didn't make it into the paper).
            \end{itemize}

      \item {\bf Code of ethics}
      \item[] Question: Does the research conducted in the paper conform, in every respect, with the NeurIPS Code of Ethics \url{https://neurips.cc/public/EthicsGuidelines}?
      \item[] Answer: \answerYes{} %
      \item[] Justification: The authors have reviewed the NeurIPS Code of Ethics and believe that the
            conducted research in the paper conforms, in every respect, with the NeurIPS Code of
            Ethics. %
      \item[] Guidelines:
            \begin{itemize}
                  \item The answer \answerNA{} means that the authors have not reviewed the NeurIPS
                        Code of Ethics.
                  \item If the authors answer \answerNo, they should explain the special circumstances
                        that require a deviation from the Code of Ethics.
                  \item The authors should make sure to preserve anonymity (e.g., if there is a special
                        consideration due to laws or regulations in their jurisdiction).
            \end{itemize}

      \item {\bf Broader impacts}
      \item[] Question: Does the paper discuss both potential positive societal impacts and negative societal impacts of the work performed?
      \item[] Answer: \answerYes{} %
      \item[] Justification: We state potential positive socieatal impact in conclusions. We do not foresee any negative societal impact.
      \item[] Guidelines:
            \begin{itemize}
                  \item The answer \answerNA{} means that there is no societal impact of the work
                        performed.
                  \item If the authors answer \answerNA{} or \answerNo, they should explain why their
                        work has no societal impact or why the paper does not address societal impact.
                  \item Examples of negative societal impacts include potential malicious or unintended
                        uses (e.g., disinformation, generating fake profiles, surveillance), fairness
                        considerations (e.g., deployment of technologies that could make decisions that
                        unfairly impact specific groups), privacy considerations, and security
                        considerations.
                  \item The conference expects that many papers will be foundational research and not
                        tied to particular applications, let alone deployments. However, if there is a
                        direct path to any negative applications, the authors should point it out. For
                        example, it is legitimate to point out that an improvement in the quality of
                        generative models could be used to generate Deepfakes for disinformation. On
                        the other hand, it is not needed to point out that a generic algorithm for
                        optimizing neural networks could enable people to train models that generate
                        Deepfakes faster.
                  \item The authors should consider possible harms that could arise when the technology
                        is being used as intended and functioning correctly, harms that could arise
                        when the technology is being used as intended but gives incorrect results, and
                        harms following from (intentional or unintentional) misuse of the technology.
                  \item If there are negative societal impacts, the authors could also discuss possible
                        mitigation strategies (e.g., gated release of models, providing defenses in
                        addition to attacks, mechanisms for monitoring misuse, mechanisms to monitor
                        how a system learns from feedback over time, improving the efficiency and
                        accessibility of ML).
            \end{itemize}

      \item {\bf Safeguards}
      \item[] Question: Does the paper describe safeguards that have been put in place for responsible release of data or models that have a high risk for misuse (e.g., pre-trained language models, image generators, or scraped datasets)?
      \item[] Answer: \answerNA{} %
      \item[] Justification: The paper poses no such risks.
      \item[] Guidelines:
            \begin{itemize}
                  \item The answer \answerNA{} means that the paper poses no such risks.
                  \item Released models that have a high risk for misuse or dual-use should be released
                        with necessary safeguards to allow for controlled use of the model, for example
                        by requiring that users adhere to usage guidelines or restrictions to access
                        the model or implementing safety filters.
                  \item Datasets that have been scraped from the Internet could pose safety risks. The
                        authors should describe how they avoided releasing unsafe images.
                  \item We recognize that providing effective safeguards is challenging, and many
                        papers do not require this, but we encourage authors to take this into account
                        and make a best faith effort.
            \end{itemize}

      \item {\bf Licenses for existing assets}
      \item[] Question: Are the creators or original owners of assets (e.g., code, data, models), used in the paper, properly credited and are the license and terms of use explicitly mentioned and properly respected?
      \item[] Answer: \answerYes{} %
      \item[] Justification: The licences of the codes and datasets used in the paper are listed in \Cref{sect:licensing}. The original papers the produced the code packages and datasets are always cited.
      \item[] Guidelines:
            \begin{itemize}
                  \item The answer \answerNA{} means that the paper does not use existing assets.
                  \item The authors should cite the original paper that produced the code package or
                        dataset.
                  \item The authors should state which version of the asset is used and, if possible,
                        include a URL.
                  \item The name of the license (e.g., CC-BY 4.0) should be included for each asset.
                  \item For scraped data from a particular source (e.g., website), the copyright and
                        terms of service of that source should be provided.
                  \item If assets are released, the license, copyright information, and terms of use in
                        the package should be provided. For popular datasets,
                        \url{paperswithcode.com/datasets} has curated licenses for some datasets. Their
                        licensing guide can help determine the license of a dataset.
                  \item For existing datasets that are re-packaged, both the original license and the
                        license of the derived asset (if it has changed) should be provided.
                  \item If this information is not available online, the authors are encouraged to
                        reach out to the asset's creators.
            \end{itemize}

      \item {\bf New assets}
      \item[] Question: Are new assets introduced in the paper well documented and is the documentation provided alongside the assets?
      \item[] Answer:\answerYes{} %
      \item[] Justification: We released trained models, processed data, and accompanying documentation in the KU Leuven Research Data Repository, as stated in \Cref{sect:supp_code_and_data}. 
      To facilitate access, we also released the same package as a mirror on the HuggingFace platform under a CC BY 4.0 license.   
      \item[] Guidelines:
            \begin{itemize}
                  \item The answer \answerNA{} means that the paper does not release new assets.
                  \item Researchers should communicate the details of the dataset\slash code\slash
                        model as part of their submissions via structured templates. This includes
                        details about training, license, limitations, etc.
                  \item The paper should discuss whether and how consent was obtained from people whose
                        asset is used.
                  \item At submission time, remember to anonymize your assets (if applicable). You can
                        either create an anonymized URL or include an anonymized zip file.
            \end{itemize}

      \item {\bf Crowdsourcing and research with human subjects}
      \item[] Question: For crowdsourcing experiments and research with human subjects, does the paper include the full text of instructions given to participants and screenshots, if applicable, as well as details about compensation (if any)?
      \item[] Answer: \answerNA{} %
      \item[] Justification: The paper does not involve crowdsourcing nor research with human subjects.
      \item[] Guidelines:
            \begin{itemize}
                  \item The answer \answerNA{} means that the paper does not involve crowdsourcing nor
                        research with human subjects.
                  \item Including this information in the supplemental material is fine, but if the
                        main contribution of the paper involves human subjects, then as much detail as
                        possible should be included in the main paper.
                  \item According to the NeurIPS Code of Ethics, workers involved in data collection,
                        curation, or other labor should be paid at least the minimum wage in the
                        country of the data collector.
            \end{itemize}

      \item {\bf Institutional review board (IRB) approvals or equivalent for research with human subjects}
      \item[] Question: Does the paper describe potential risks incurred by study participants, whether such risks were disclosed to the subjects, and whether Institutional Review Board (IRB) approvals (or an equivalent approval/review based on the requirements of your country or institution) were obtained?
      \item[] Answer: \answerNA{} %
      \item[] Justification: The paper does not involve crowdsourcing nor research with human subjects.
      \item[] Guidelines:
            \begin{itemize}
                  \item The answer \answerNA{} means that the paper does not involve crowdsourcing nor
                        research with human subjects.
                  \item Depending on the country in which research is conducted, IRB approval (or
                        equivalent) may be required for any human subjects research. If you obtained
                        IRB approval, you should clearly state this in the paper.
                  \item We recognize that the procedures for this may vary significantly between
                        institutions and locations, and we expect authors to adhere to the NeurIPS Code
                        of Ethics and the guidelines for their institution.
                  \item For initial submissions, do not include any information that would break
                        anonymity (if applicable), such as the institution conducting the review.
            \end{itemize}

      \item {\bf Declaration of LLM usage}
      \item[] Question: Does the paper describe the usage of LLMs if it is an important, original, or non-standard component of the core methods in this research? Note that if the LLM is used only for writing, editing, or formatting purposes and does \emph{not} impact the core methodology, scientific rigor, or originality of the research, declaration is not required.
      \item[] Answer: \answerNA{} %
      \item[] Justification: The core method development in this research does not involve LLMs as any important, original, or non-standard components.
      \item[] Guidelines:
            \begin{itemize}
                  \item The answer \answerNA{} means that the core method development in this research
                        does not involve LLMs as any important, original, or non-standard components.
                  \item Please refer to our LLM policy in the NeurIPS handbook for what should or
                        should not be described.
            \end{itemize}

\end{enumerate}

\typeout{get arXiv to do 4 passes: Label(s) may have changed. Rerun}

\end{document}